\documentclass[10pt]{article}

\usepackage[final]{graphicx} 
\usepackage[font=normalsize,labelfont=sf,textfont=sf,caption=false]{subfig}

\usepackage{cite}

\usepackage{url}

\usepackage{algorithm}
\usepackage{algorithmic}

\usepackage{multirow}
\usepackage{amsfonts}
\usepackage{amsmath}

\newcommand{\IR}{\mathbb{R}}

\usepackage{tikz}

\begin{document} 
\title{The Data Replication Method for the Classification with Reject Option}
\author{Ricardo Sousa and Jaime S. Cardoso}

\maketitle

\begin{abstract} 
Classification is one of the most important tasks of machine learning. Although the most well studied model is the two-class problem, in many scenarios there is the opportunity to label critical items for manual revision, instead of trying to automatically classify every item.

In this paper we adapt a paradigm initially proposed for the classification of ordinal data to address the classification problem with reject option. The technique reduces the problem of classifying with reject option to the standard two-class problem. The introduced method is then mapped into support vector machines and neural networks. Finally, the framework is extended to multiclass ordinal data with reject option.
An experimental study with synthetic and real data sets, verifies the usefulness of the proposed approach.

\end{abstract} 
\textbf{keywords: } Reject Option, Support Vector Machines, Neural Networks, Supervised Learning, Classification

\section{Introduction}
Decision support systems are becoming ubiquitous in many human activities, most notably in finance and medicine.
Automatic models are being developed to imitate, as closely as possible, the usual human decision.
Within this context, classification is one of the most representative predictive learning tasks.
Classification predicts a categorical value for a specific data item.
The most well studied scenario is when the class to be predicted can assume only two values---binary setting. 
The classifier is developed to partition the feature space in two regions, discriminating between the two classes.

In credit scoring modelling, models are developed to determine how likely applicants are to default with their repayments. Previous repayment history is used to determine whether a customer should be classified into a `good' or a `bad' category~\cite{Thomas2002}. 
Prediction of insurance companies' insolvency has arisen as an important problem in the field of financial research, due to the necessity of protecting the general public whilst minimising the costs associated to this problem~\cite{Thomas2002}. 
In medicine, the last decades have witnessed the development of advanced diagnostic systems as alternative, complementary or a first opinion in many applications~\cite{AIME2007}.
These are just some applications that continue to challenge researchers in the deployment of fully automated decision support systems.

One of the problems with classifying complex items is that many items from distinct classes have similar structures in a feature space, resulting in a setting with overlapping classes. The automation of decisions in this region leads invariably to many wrong predictions. On the other hand, and although items in the historical data are labelled {\em only} as `good' or `bad', the deployment of a decision support system in many environments has the opportunity to label critical items for manual revision, instead of trying to automatically classify every and each item. 
The system automates only those decisions which can be reliably predicted, labelling the critical ones for a human expert to analyse. Therefore, the development of classifiers with a third output class, the reject class, in-between the good and bad classes, is attractive. 

In a preliminary study~\cite{RSousaICMLA2009}, we proposed a new learning methodology, which is extended and explored in various directions in this paper. First, we detail the presentation of the method, introducing the mapping to support vector machines and neural networks. Second, we generalized the framework from binary classification problems to multiclass ordinal data.  Finally, the experimental work reported at the end of the communication is expanded, including a comparison over more datasets and with conventional and state of the art methods.
A principled approach for learning critical regions on complex data is motivated and presented in Section~\ref{sec:two}. The proposed model of this paper is described in Section~\ref{sec:three}. Performance assessment is conducted in Section~\ref{sec:four}. Finally, conclusions are drawn in Section~\ref{sec:five}.

\section{Problem Statement and Standard Solutions}
\label{sec:two}
Predictive modelling tries to find good rules (models) for guessing (predicting) the values of one or more variables (target) in a dataset from the values of other variables. 
Our target can assume only two values, represented by `good' and `bad' classes.
When in possession of a ``complex'' dataset, a simple separator is bound to misclassify some points. 
Two types of errors are possible, `false positives' and `false negatives'. The construction (training) of a model can be conducted to optimise some adopted measure of business performance, be it profit, loss, volume of acquisitions, market share, etc, by giving appropriate weights to the two types of errors. When the weights of the two types of errors are heavily asymmetric, the boundary between the two classes will be pushed near values where the most costly error seldom happens. 

\begin{figure}[!ht] 
\begin{center}
\subfloat[Overlapping regions.]{
  \label{fig:SIMPLESETTING}
  \includegraphics[width=0.45\linewidth]{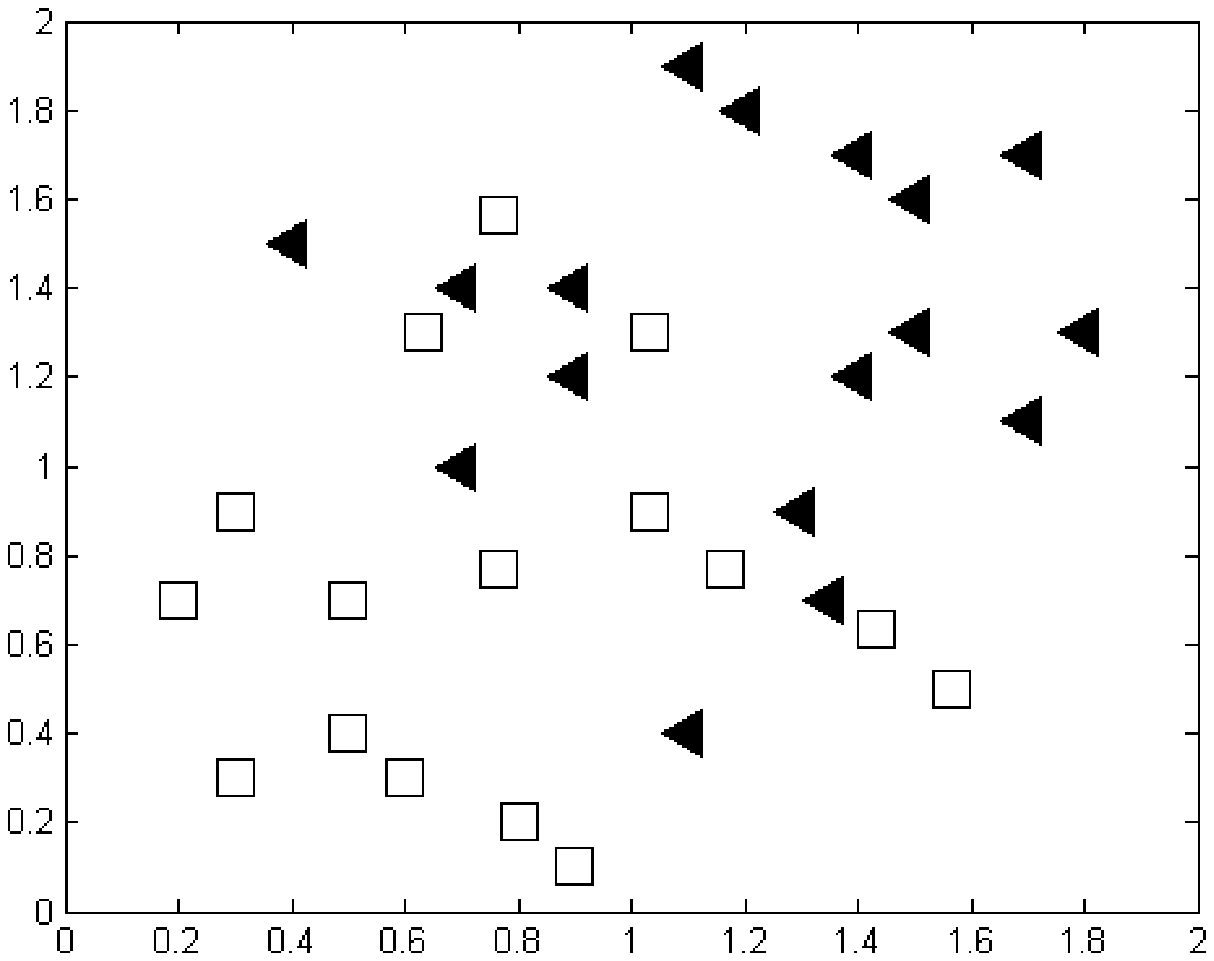}}
\hfill
\subfloat[Typical separator lines, obtained with two independent binary classifiers.]{
  \label{fig:FacingArmsUp}
  \includegraphics[width=0.45\linewidth]{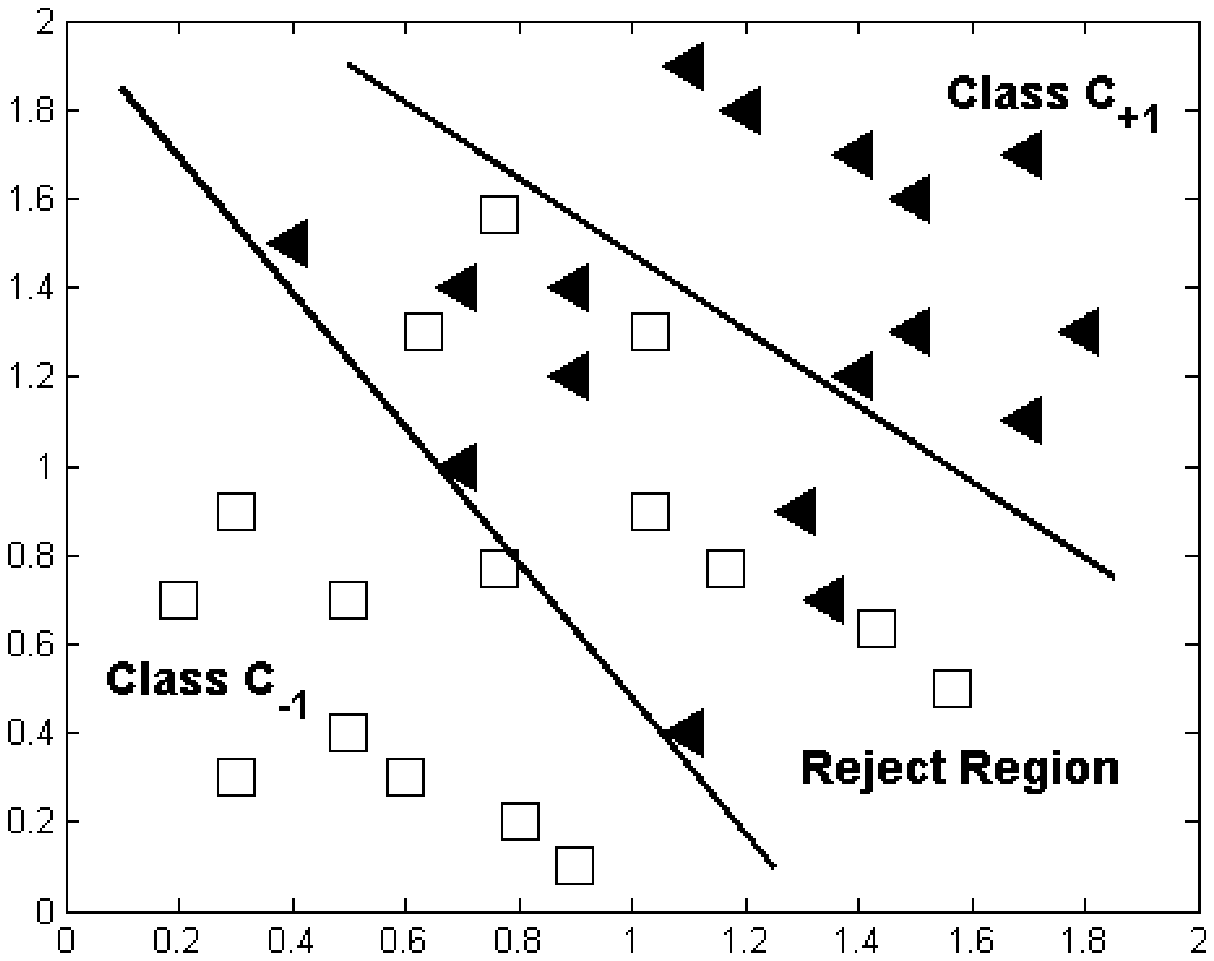}}
\caption{Illustrative setting with overlapping classes.}
\label{fig:SeiLa}
\end{center}
\end{figure}

\begin{figure}[!ht] 
\begin{center}
\subfloat[Intersecting separating lines.]{
        \label{fig:IntersectingSetting}
        \includegraphics[width=0.45\linewidth]{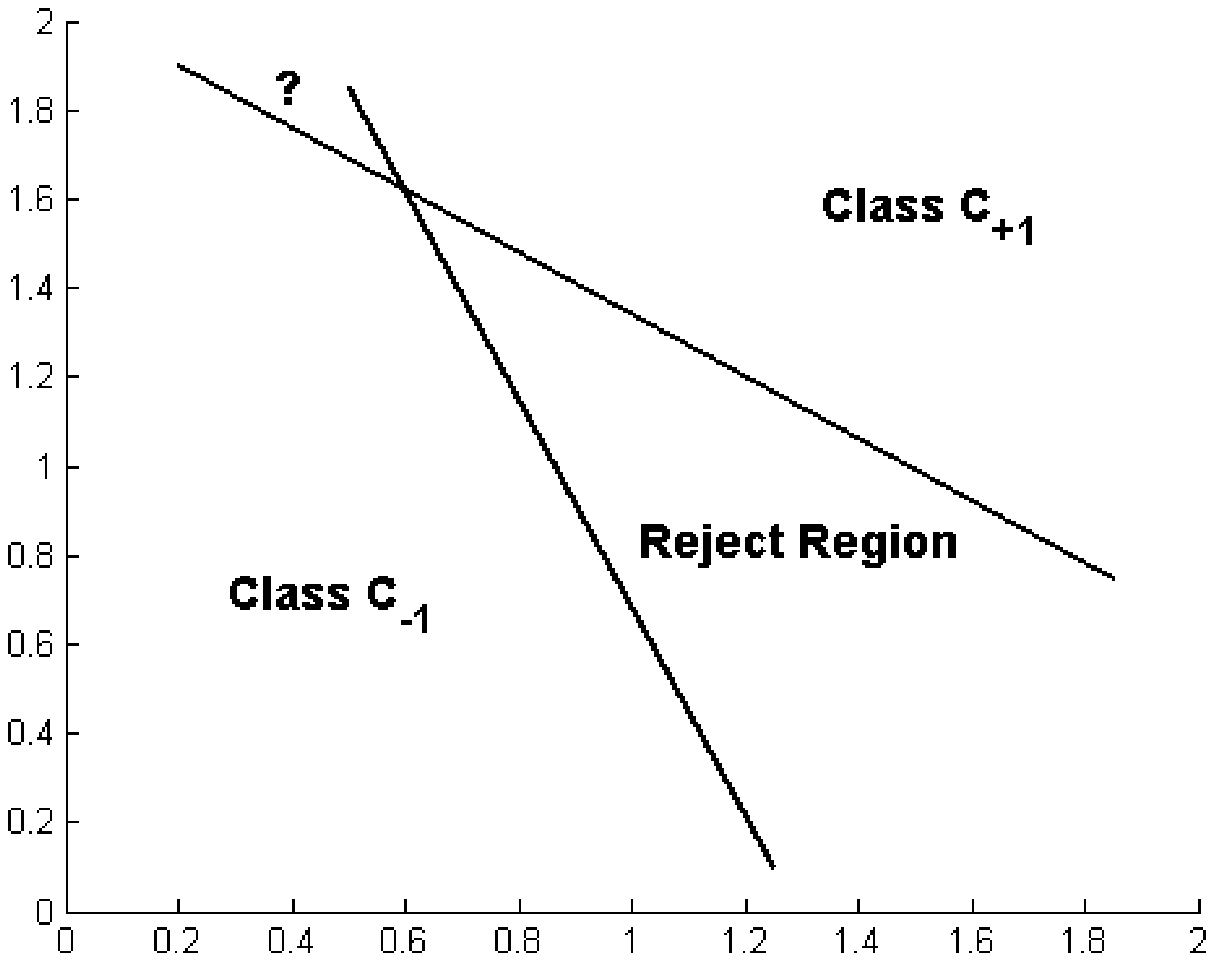}}
        \hfill
\subfloat[Non-intersecting separating lines.]{
        \label{fig:NonIntersectingSetting}
        \includegraphics[width=0.45\linewidth]{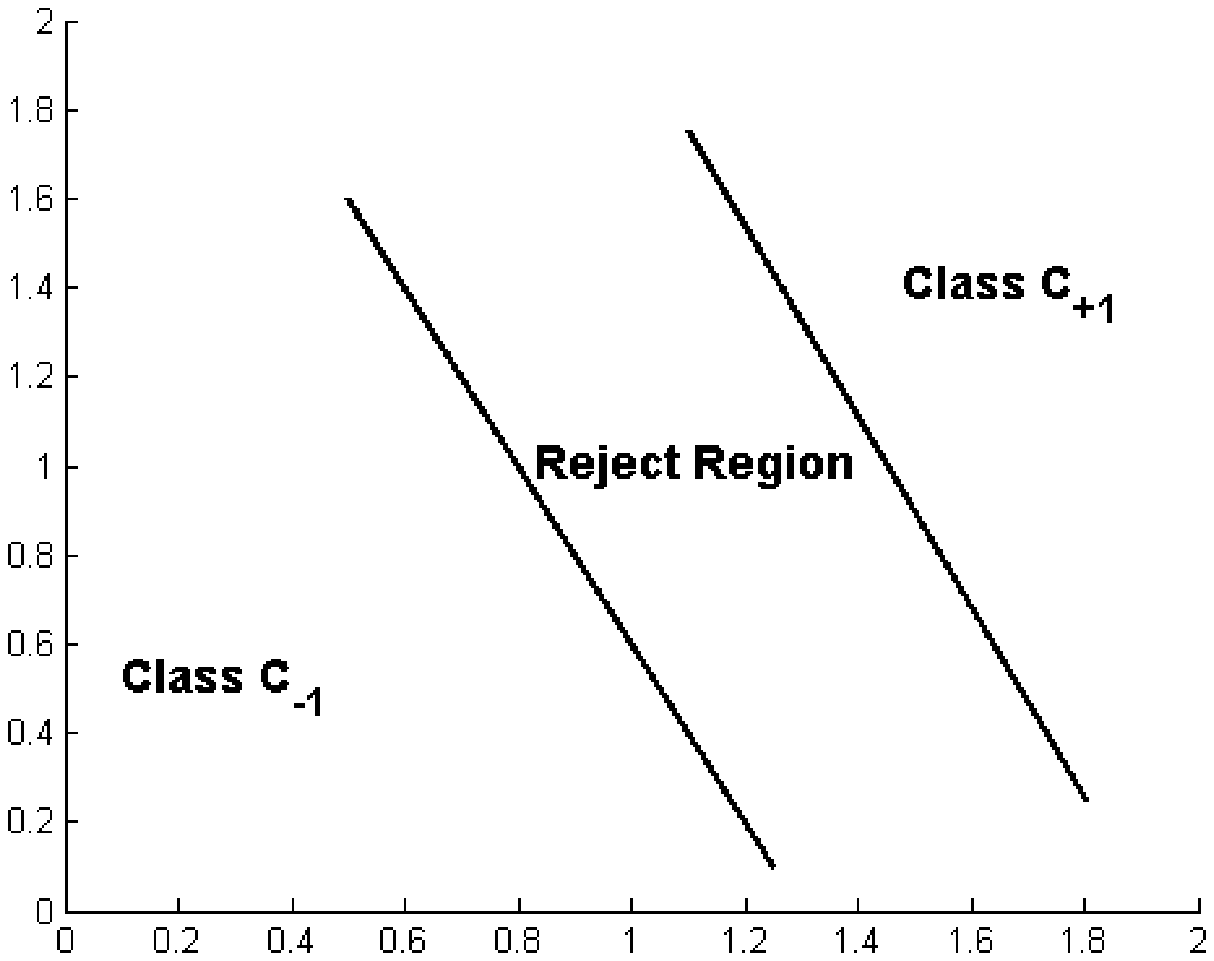}}
\caption{Potential discriminative boundaries.}
\label{fig:SeiLa2}
\end{center}
\end{figure}

This fact suggests a simple procedure to construct a three-class output classifier: training a first binary classifier with a set of weights heavily penalising the false negative errors, we expect that when this classifier predicts an item as negative, it will be truly negative. 
Likewise, training a second binary classifier with a set of weights heavily penalising the false positive errors, we expect that when this classifier predicts an item as positive, it will be truly positive.  
When a item is predicted as positive by the first classifier and negative by the second, it will be labelled for review. This setting is illustrated in Fig.~\ref{fig:SeiLa}.
A problem arises when an item is predicted as positive by the first classifier and negative by the second classifier as in Fig.~\ref{fig:IntersectingSetting}. That can happen because the two separator lines intersect each other. A convenient workaround is then to avoid this problematic state by imposing that the two boundaries of the classifiers do not intersect, Fig.~\ref{fig:NonIntersectingSetting}.

Before delving into the proposed method, it is worth discussing the simple solution of using a single classifier. If more than just discriminating between the two classes, the model to use yields a posterior probability for each target class, then two cutoffs can be defined on this value. All items with predicted probability of belonging to class ${\cal C}_{-1}$ less than a low threshold are labelled as ${\cal C}_{+1}$, items with predicted probability of belonging to class ${\cal C}_{-1}$ higher than a high threshold are labelled as ${\cal C}_{-1}$, items with predicted probability of belonging to class ${\cal C}_{-1}$ in-between the low and high threshold are labelled for review. Two issues were identified with this approach. First, we need to estimate the probability of each class, which is by itself a problem harder than the problem of discriminating classes. Second, the estimation of the two cutoffs is not straightforward nor can be easily fitted into standard frameworks.

The design of classifiers with reject option can be systematised in three different approaches:
\begin{itemize}
\item the design of two, {\em independent}, classifiers. A first classifier is trained to output ${{\cal C}_{-1}}$ only when the probability of ${{\cal C}_{-1}}$  is high and a second classifier trained to output ${{\cal C}_{+1}}$ only when the probability of ${{\cal C}_{+1}}$  is high. The simplicity of this strategy has the weakness of producing intersecting boundaries, leading to regions with a non-logical decision.
\item the design of a single, standard binary classifier. This approach already provides non-intersecting boundaries.  If the classifier provides some approximation to the a posterior class probabilities, then a pattern is rejected if the maximum of the two posterior probabilities is lower than a given threshold. If the classifier does not provide probabilistic outputs, then a rejection threshold targeted to the particular classifier is used. For example, the rejection techniques proposed with support vector machines consist in rejecting patterns those distance from the optimal separating hyperplane is lower than a predefined threshold. The rejection region is determined {\em after} the training of the classifier, by defining appropriate threshold values on the output of the classifier.
\item the design of a single classifier with embedded reject option. This approach has consisted in the design of algorithms specifically adapted for the reject option problem. Although the option has the advantage of determining the reject region during the training phase of the classifier, it requires the implementation of very specific algorithms, usually appropriate for a single class of classifiers, like support vector machines~\cite{Fumera2002a,Bounsiar2008}.
\end{itemize}
In the next subsection we detail some of the previous work in the area related with our proposal.
\subsection{Previous Works}
In one of the first works to analyze the tradeoffs between erring and rejecting, Chow~\cite{Chow1970} derived a general error and reject tradeoff relation for the Bayes optimum recognition system under the assumption of complete knowledge of the a priori probability distribution of the classes and the posterior probabilities which, in real problems, are usually unknown.
Fumera et al.~\cite{Fumera2000a,Fumera2000b} show that Chow's rule does not perform well if a significant error in probability estimation is present, proposing the use of multiple reject thresholds related to the data classes.

In classification with rejection option, the key parameter is the threshold that defines the reject area.
Noticing that the reject region should be determined during the training phase of a classifier, Fumera~\cite{Fumera2002a} proposed a modified training for support vector machines (SVMs) with embedded reject option. A similar approach was applied to Multiple Instance Learning (MIL) for image categorisation~\cite{Zhang2006}. A major difficulty with these approaches is that the resulting formulations are no longer standard optimization procedures and cannot be solved efficiently, lacking some appealing features like convexity and sparsity.
 
In the same line, Bartlett and Wegkamp~\cite{Bartlett2008} consider a convex surrogate of the generalized loss function to efficiently solve the resulting problem under SVMs. Grandvalet  et al.~\cite{Grandvalet2008} extent this work with a double hinge function and a probabilistic viewpoint of the SVM fitting.


In this work we detail a solution that: a) uses standard binary classifiers; b) produces non-intersecting boundaries; c) determines the reject region during the training phase.
The proposed solution is based on the extension of a technique developed for ordinal data. 

\section{An Ordinal Data Approach for Detecting Reject Regions}
\label{sec:three}

The rejection method to be proposed is an extension of a method already proposed in the literature but for the classification of ordinal data. Therefore, and for completeness, we start by reviewing the data replication method; next, we present the novel aspects introduced in this article.
\subsection{The Data Replication Method for Ordinal Data}
The data replication method for ordinal data can be framed under the single binary classifier reduction (SBC), an approach for solving multiclass problems via binary classification relying on a single, standard binary classifier. SBC reductions can be obtained by embedding the original problem in a higher-dimensional space consisting of the original features, as well as one or more other features determined by fixed vectors, designated here as {\em extension features}. This embedding is implemented by replicating the training set points so that a copy of the original point is concatenated with each of the extension features' vectors. The binary labels of the replicated points are set to maintain a particular structure in the extended space. This construction results in an instance of an artificial binary problem, which is fed to a binary learning algorithm that outputs a single binary classifier. To classify a new point, the point is replicated and extended similarly and the resulting replicas are fed to the binary classifier, which generates a number of signals, one for each replica. The class is determined as a function of these signals~\cite{El-Yaniv2008}.

To introduce the data replication method, assume that examples in a classification problem come from one of $K$ ordered classes, labelled from ${\cal C}_1$ to ${\cal C}_K$, 
corresponding to their natural order.
Consider the training set 
$\{\textbf{x}_i^{(k)}\}$, where $k=1, \ldots, K$ denotes the class number, $i=1,\ldots,\ell_k$ is the index within
each class, and $\textbf{x}_i^{(k)} \in \IR^p$, with $p$ the dimension of the feature space. 
Let $\ell = \sum_{k=1}^K \ell_{k}$ be the total number of training examples.	

\begin{figure*}[!ht]
  \centering
  \subfloat[Original dataset in $\IR^2$, $K=3$.]{
    \label{fig:binaryProblemsData}
    \includegraphics[width=0.25\linewidth]{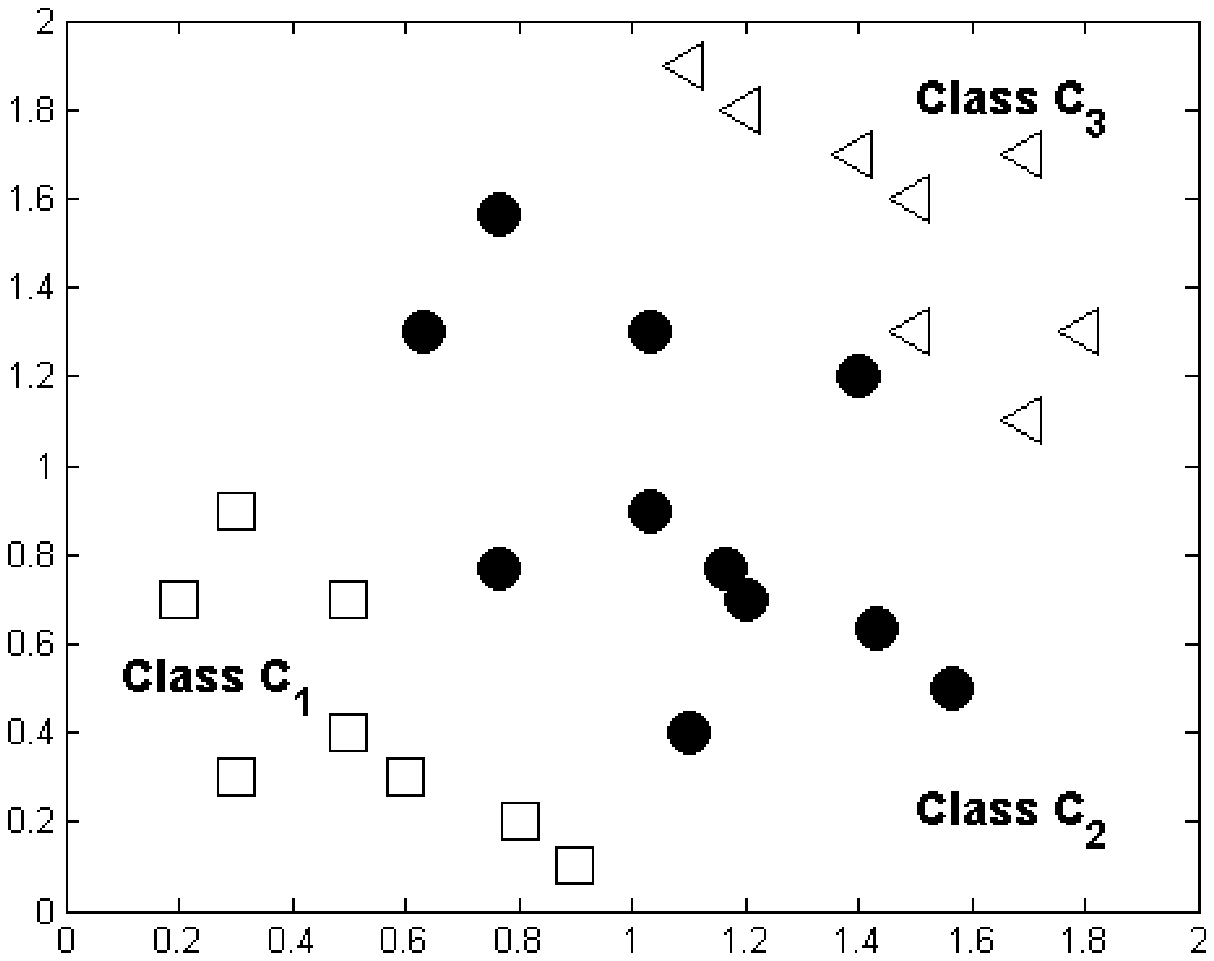}}
  \quad
  \subfloat[Binary problem ${\cal C}_{1}$ against classes $\{{\cal C}_{2}, {\cal C}_{3}\}$.]{
    \label{fig:binaryProblems1}
    \includegraphics[width=0.25\linewidth]{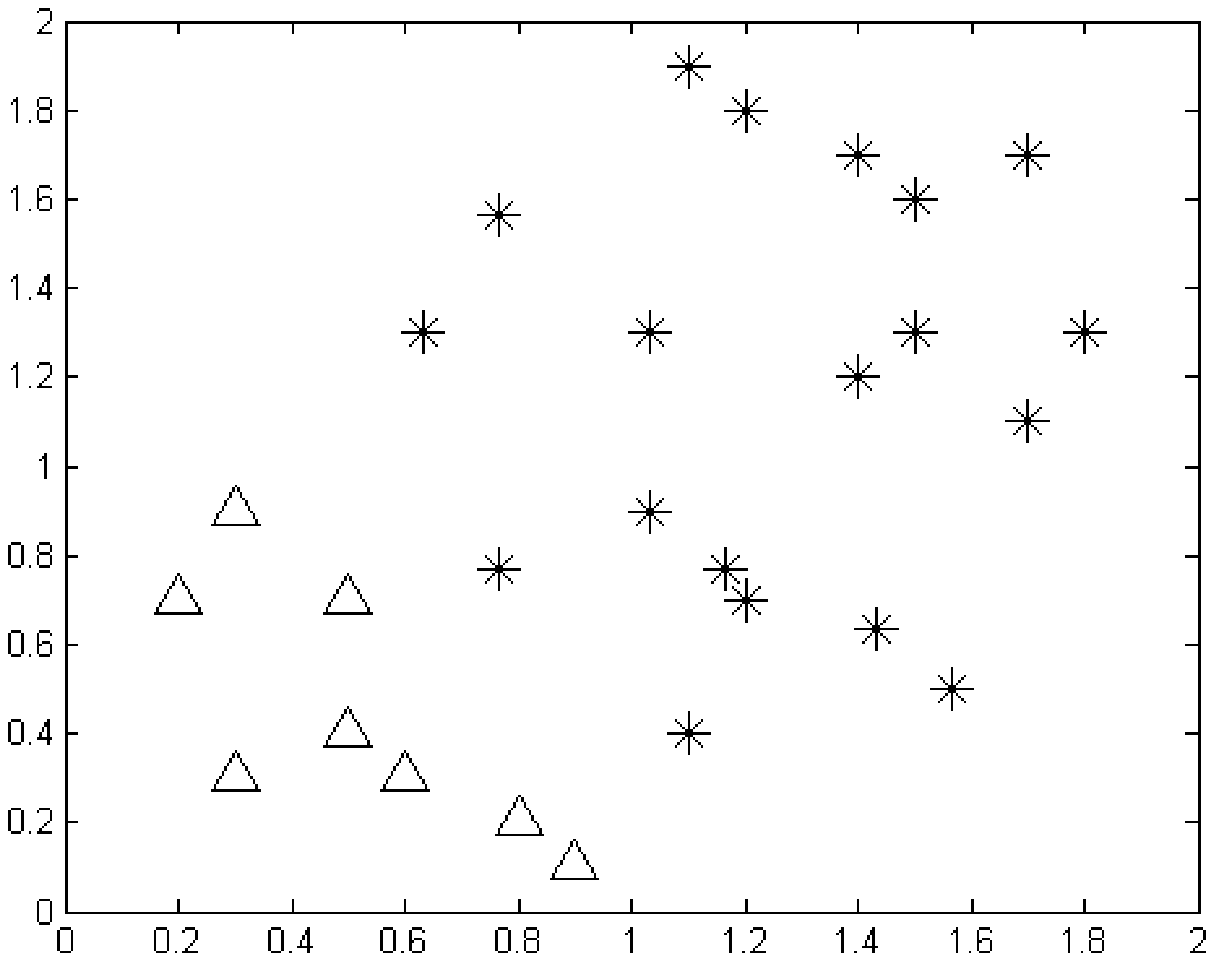}}
  \quad
  \subfloat[Binary problem $\{{\cal C}_{1}, {\cal C}_{2}\}$ against class ${\cal C}_{3}$.]{
    \label{fig:binaryProblems2}
    \includegraphics[width=0.25\linewidth]{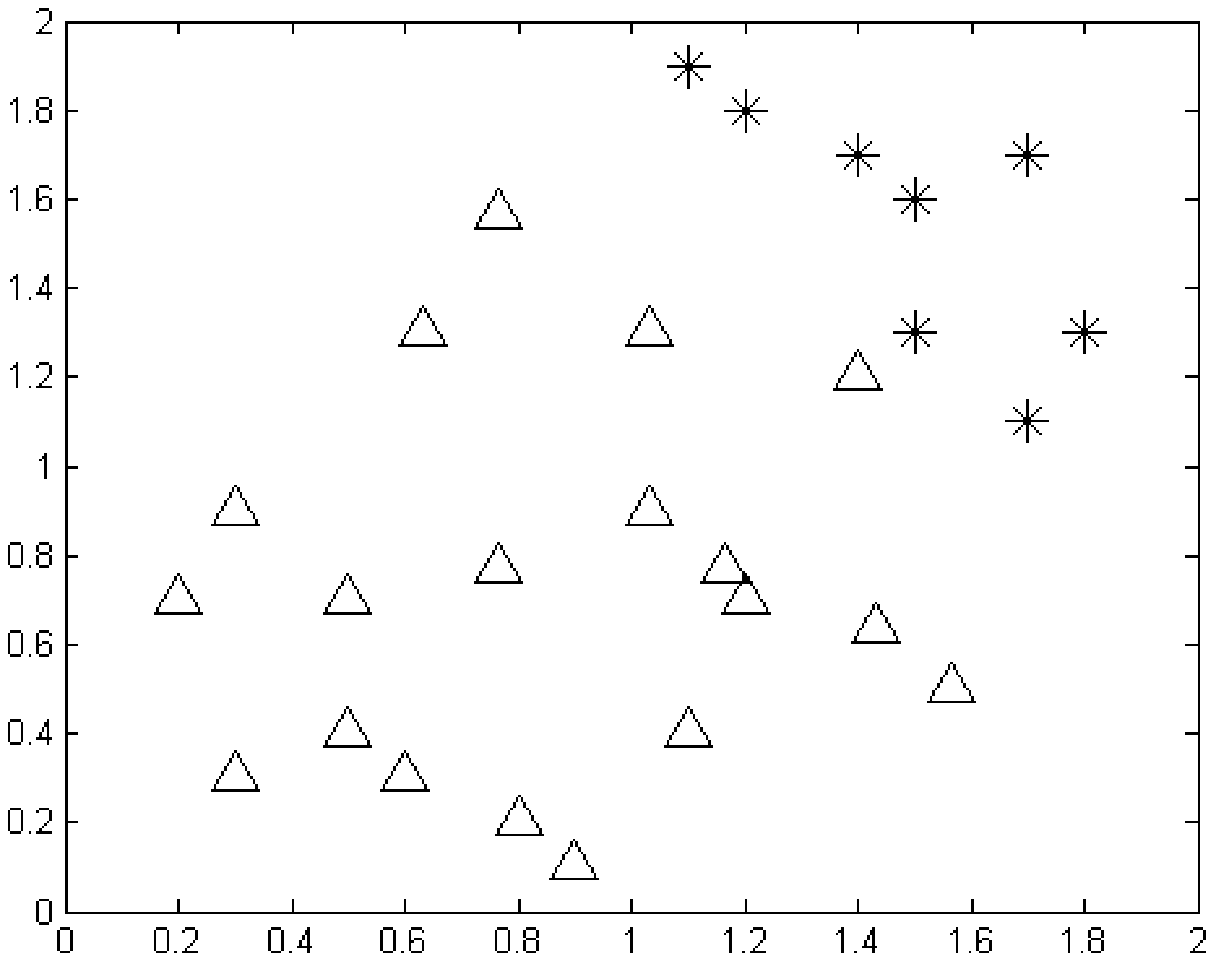}}
  \caption{\label{fig:binaryProblems}Binary problems to be solved simultaneously with the data replication method.}
\end{figure*}

Let us consider a very simplified toy example with just three classes, as depicted in Fig.~\ref{fig:binaryProblemsData}. 
Here, the task is to find two parallel hyperplanes, the first one discriminating class ${\cal C}_{1}$ against classes $\{{\cal C}_{2}, {\cal C}_{3}\}$ and the second hyperplane discriminating classes $\{{\cal C}_{1}, {\cal C}_{2}\}$ against class ${\cal C}_{3}$. These hyperplanes will correspond to the solution of two binary classification problems but with the additional constraint of parallelism---see Fig.~\ref{fig:binaryProblems}. The data replication method suggests solving both problems simultaneously in an augmented feature space~\cite{JaimeJMLR2007}.

\begin{figure}[!ht]
  \centering
  \subfloat[Dataset in $\IR^3$, with samples replicated ($h=1$).]{
    \label{fig:proposedModel2}
    \includegraphics[width=0.45\linewidth]{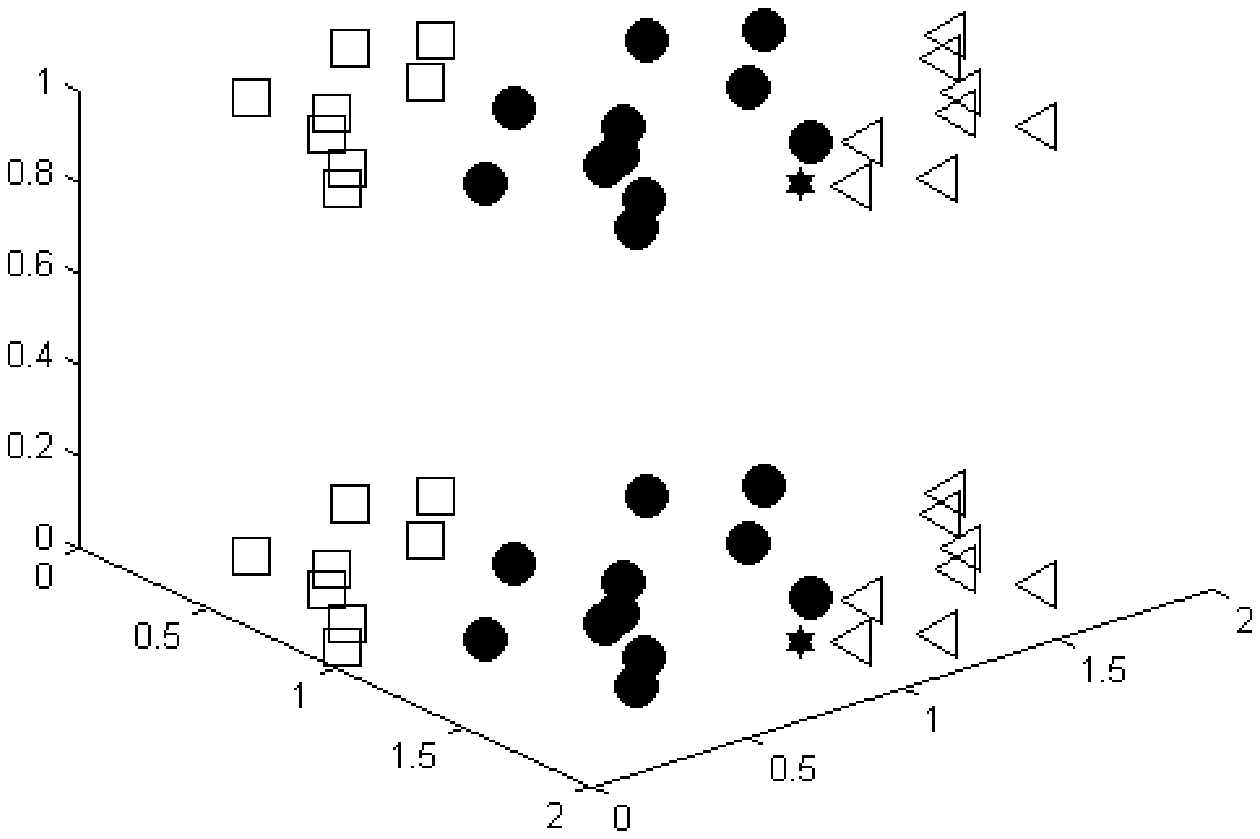}}
  \quad
  \subfloat[Transformation into a binary classification problem.]{
    \label{fig:proposedModel3}
    \includegraphics[width=0.45\linewidth]{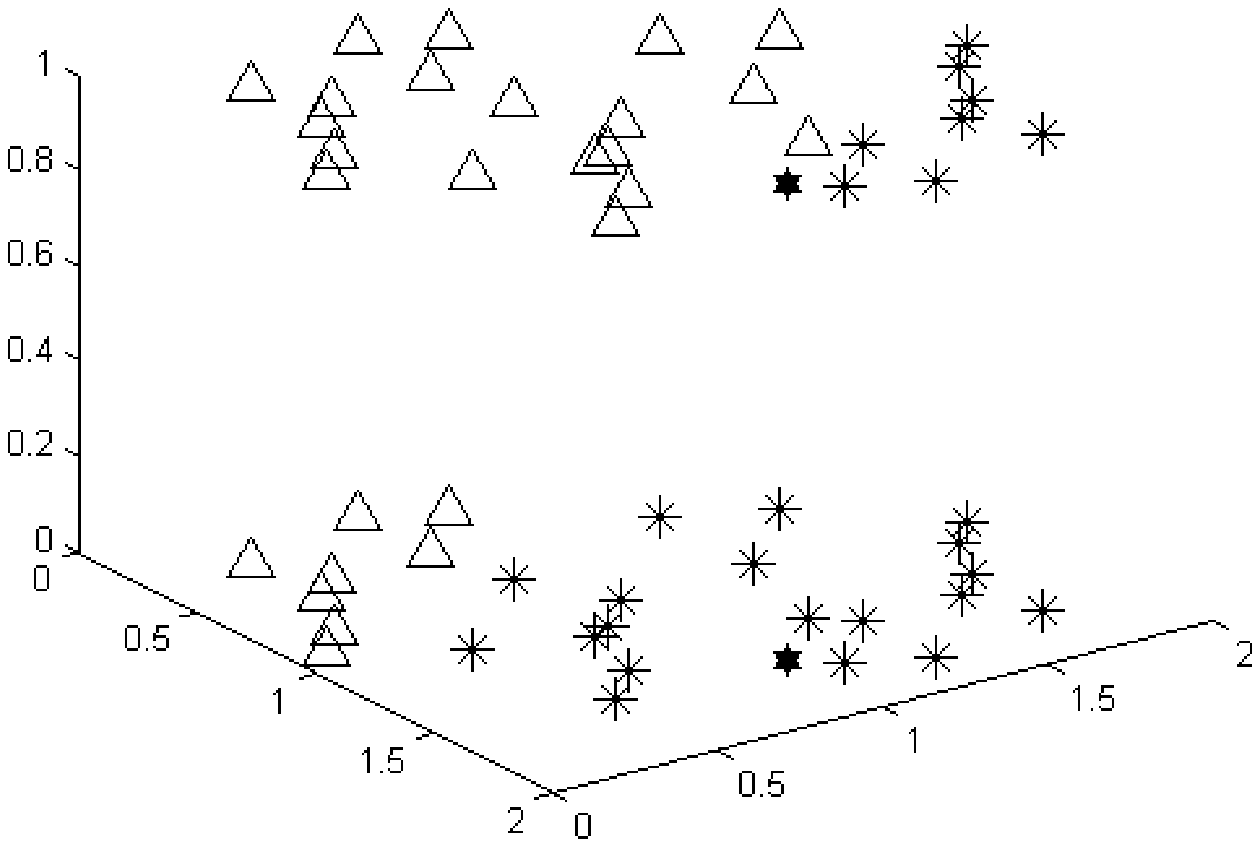}}
  \quad
  \subfloat[Linear solution to the binary problem.]{
    \label{fig:proposedModel4}
    \includegraphics[width=0.45\linewidth]{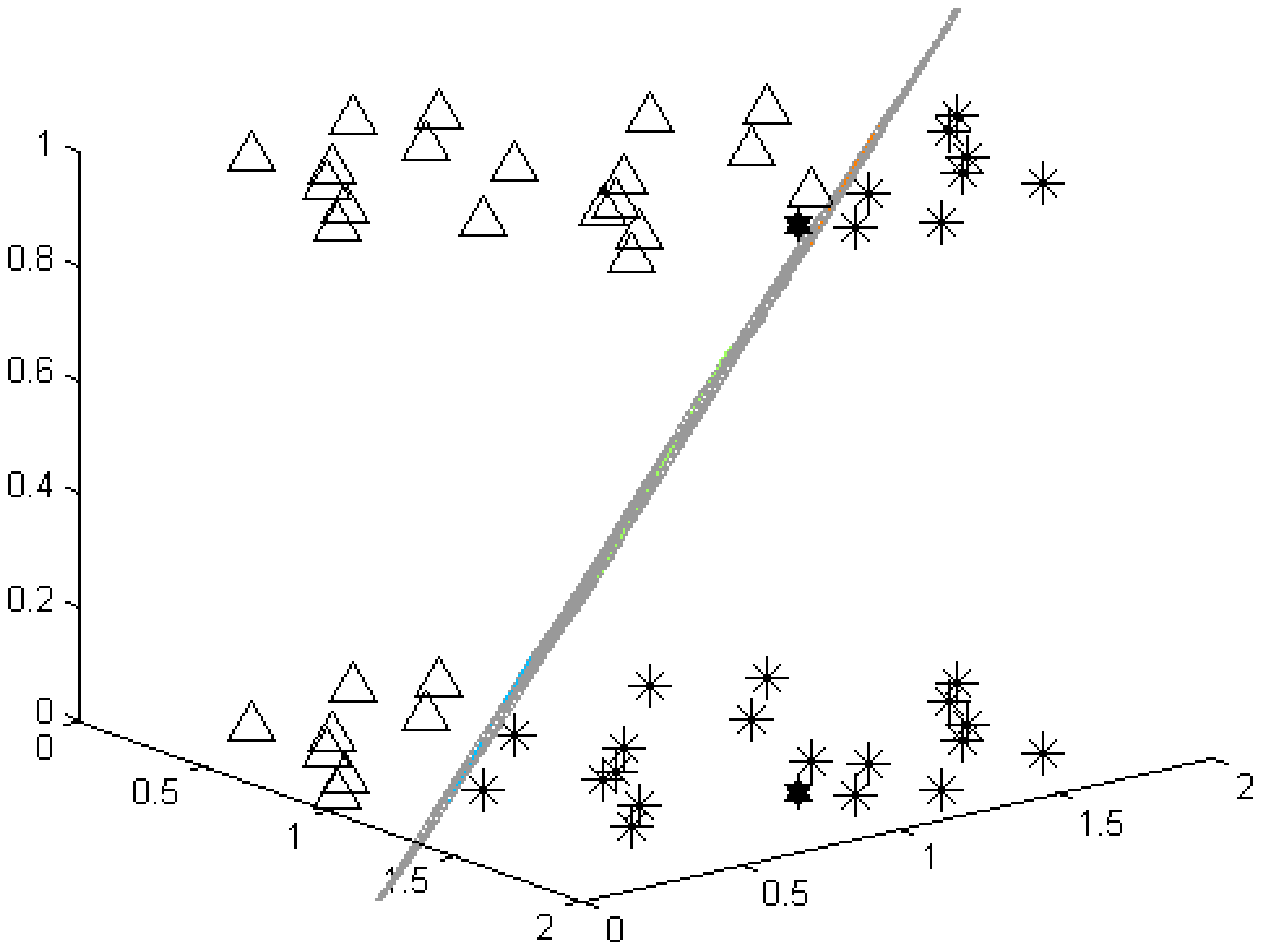}}
  \quad
  \subfloat[Linear solution in the original dataset.]{
    \label{fig:proposedModel5}
    \includegraphics[width=0.45\linewidth]{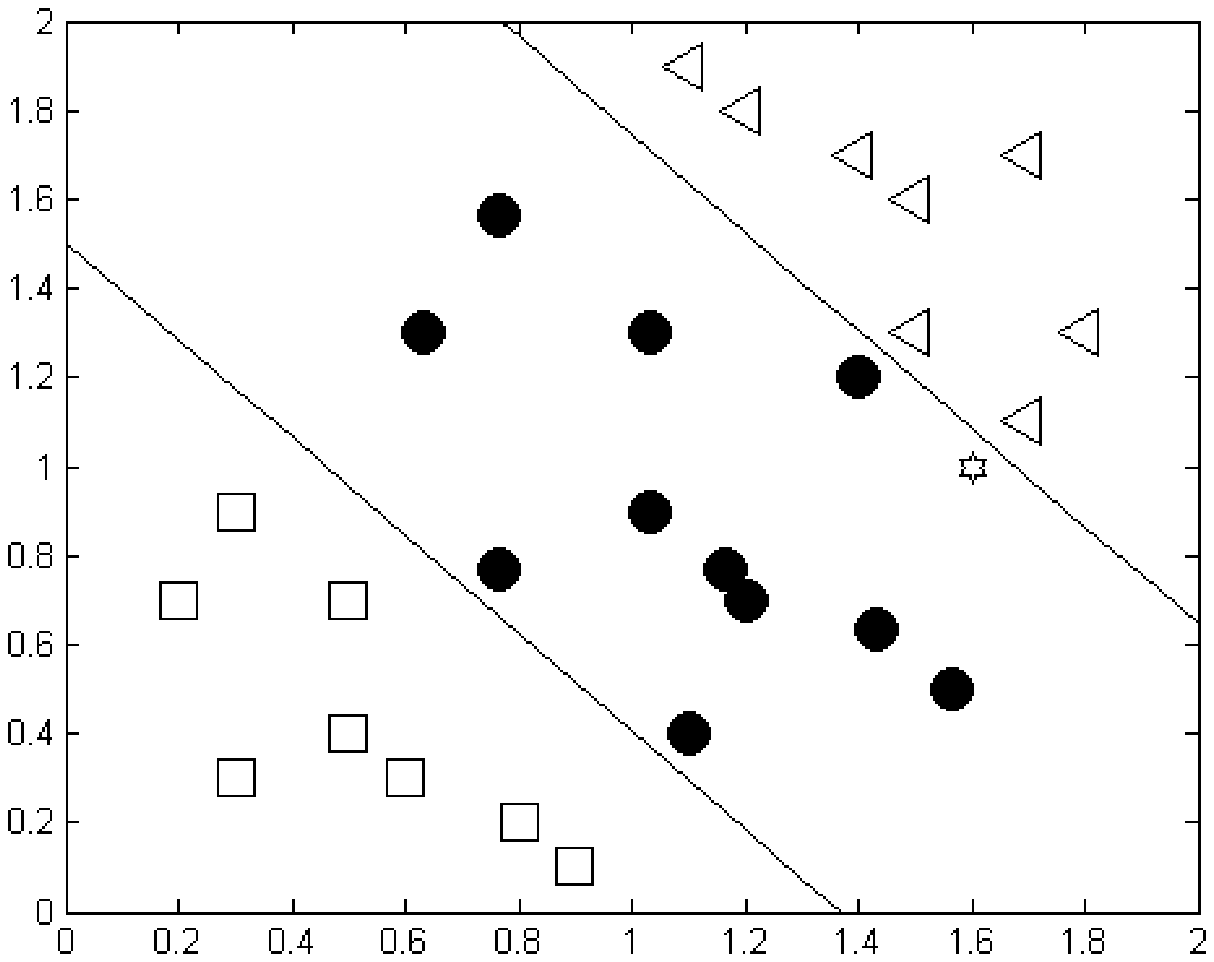}}
  \caption{Data replication model in a toy example (from~\cite{JaimeJMLR2007}).}
  \label{fig:proposedModel}
\end{figure}

In the toy example, using a transformation from the $\IR^2$ initial feature-space to a $\IR^3$ feature space, 
replicate each original point, according to the rule (see Fig.~\ref{fig:proposedModel2}):
\[
\textbf{x} \in \IR^2 
\begin{matrix}
\nearrow \\
\searrow 
\end{matrix}
\begin{matrix}
\left[\begin{smallmatrix}\textbf{x}\\ h\end{smallmatrix}\right] \in \IR^{3}\\
\\
\left[\begin{smallmatrix}\textbf{x}\\ 0\end{smallmatrix}\right] \in \IR^{3}
\end{matrix}, \text{ where $h=$ const} \in \IR^+
\]
Observe that any two points created from the same original point differ only in the extension feature.
Define now a binary training set in the new (higher dimensional) space according to (see Fig.~\ref{fig:proposedModel3}):
\begin{equation}
\begin{split}
\label{toyReplication}
\left[\begin{smallmatrix}\textbf{x}_i^{(1)}\\ 0\end{smallmatrix}\right] \in \overline{{\cal C}}_1, \
\left[\begin{smallmatrix}\textbf{x}_i^{(2)}\\ 0\end{smallmatrix}\right], 
\left[\begin{smallmatrix}\textbf{x}_i^{(3)}\\ 0\end{smallmatrix}\right] \in \overline{{\cal C}}_2 \\ \ 
\left[\begin{smallmatrix}\textbf{x}_i^{(1)}\\ h\end{smallmatrix}\right], 
\left[\begin{smallmatrix}\textbf{x}_i^{(2)}\\ h\end{smallmatrix}\right] \in \overline{{\cal C}}_1, \
\left[\begin{smallmatrix}\textbf{x}_i^{(3)}\\ h\end{smallmatrix}\right] \in \overline{{\cal C}}_2 
\end{split}
\end{equation}
In this step we are defining the two binary problems as a single binary problem in the augmented feature space.
A linear two-class classifier can now be applied on the extended dataset, yielding a hyperplane separating the two classes, see Fig.~\ref{fig:proposedModel4}.
The intersection of this hyperplane with each of the subspace replicas can be used to derive the boundaries in the original dataset, as illustrated in Fig.~\ref{fig:proposedModel5}.

To predict the class of an unseen example, classify both replicas of the example in the extended dataset with the binary classifier. From the sequence of binary labels one can infer the predicted label on the original ordinal classes 
\begin{equation*}
\overline{{\cal C}}_1, \overline{{\cal C}}_1 \Longrightarrow {\cal C}_1\quad 
\overline{{\cal C}}_2, \overline{{\cal C}}_1 \Longrightarrow {\cal C}_2\quad
\overline{{\cal C}}_2, \overline{{\cal C}}_2 \Longrightarrow {\cal C}_3
\end{equation*}
Note that only three sequences are possible~\cite{JaimeJMLR2007}. The generalisation for any problem in $\IR^p$, with $K$ ordinal classes and nonlinear boundaries can be found in~\cite{JaimeJMLR2007}.

Summing up, $(K-1)$ replicas in a $\IR^{p+K-2}$ dimensional space are used to train a binary classifier. 
The target class of an unseen example can be obtained by adding one to the number of ${\cal C}_2$ labels in the sequence of binary labels resulting from the classification of the $(K-1)$ replicas of the example.
\subsection{The Data Replication Method for Detecting Reject Regions}
The scenario of designing a classifier with reject option shares many characteristics with the classification of ordinal data.  It is also reasonable to assume for the reject option scenario that the three output classes are naturally ordered as ${\cal C}_1, {\cal C}_{reject},{\cal C}_2$. As the intersection point of the two boundaries would indicate an example with the three classes equally probable---one would be equally uncertain between assigning ${\cal C}_1$ or ${\cal C}_{reject}$ and between assigning $ {\cal C}_{reject}$ or ${\cal C}_2$---it is plausible to adopt a strategy imposing non-intersecting boundaries. 
In fact, as reviewed in Section~\ref{sec:two}, methods have been proposed with exactly such assumption.
In the scenario of designing a classifier with reject option, we are interested on finding two boundaries: a boundary discriminating ${\cal C}_1$ from $\{{\cal C}_{reject},{\cal C}_2\}$ and a boundary discriminating $\{{\cal C}_1, {\cal C}_{reject}\}$ from ${\cal C}_2$.

\begin{figure}[!ht]
  \centering
  \subfloat[Original binary dataset in $\IR^2$.]{
    \label{fig:rejectModel1}
    \includegraphics[width=0.45\linewidth]{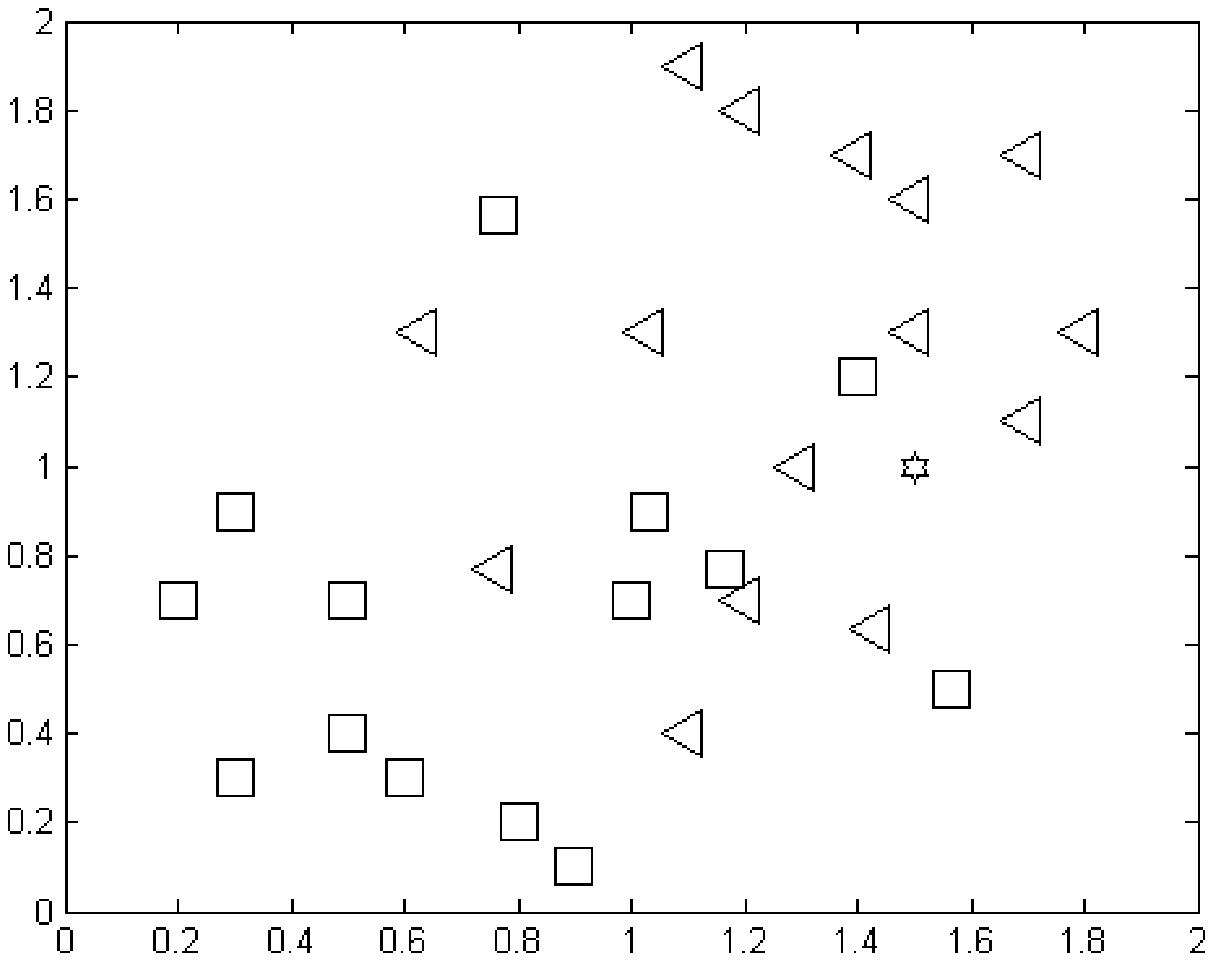}}
  \quad
  \subfloat[Dataset in $\IR^3$, with samples replicated ($h=1$).]{
    \label{fig:rejectModel2}
    \includegraphics[width=0.45\linewidth]{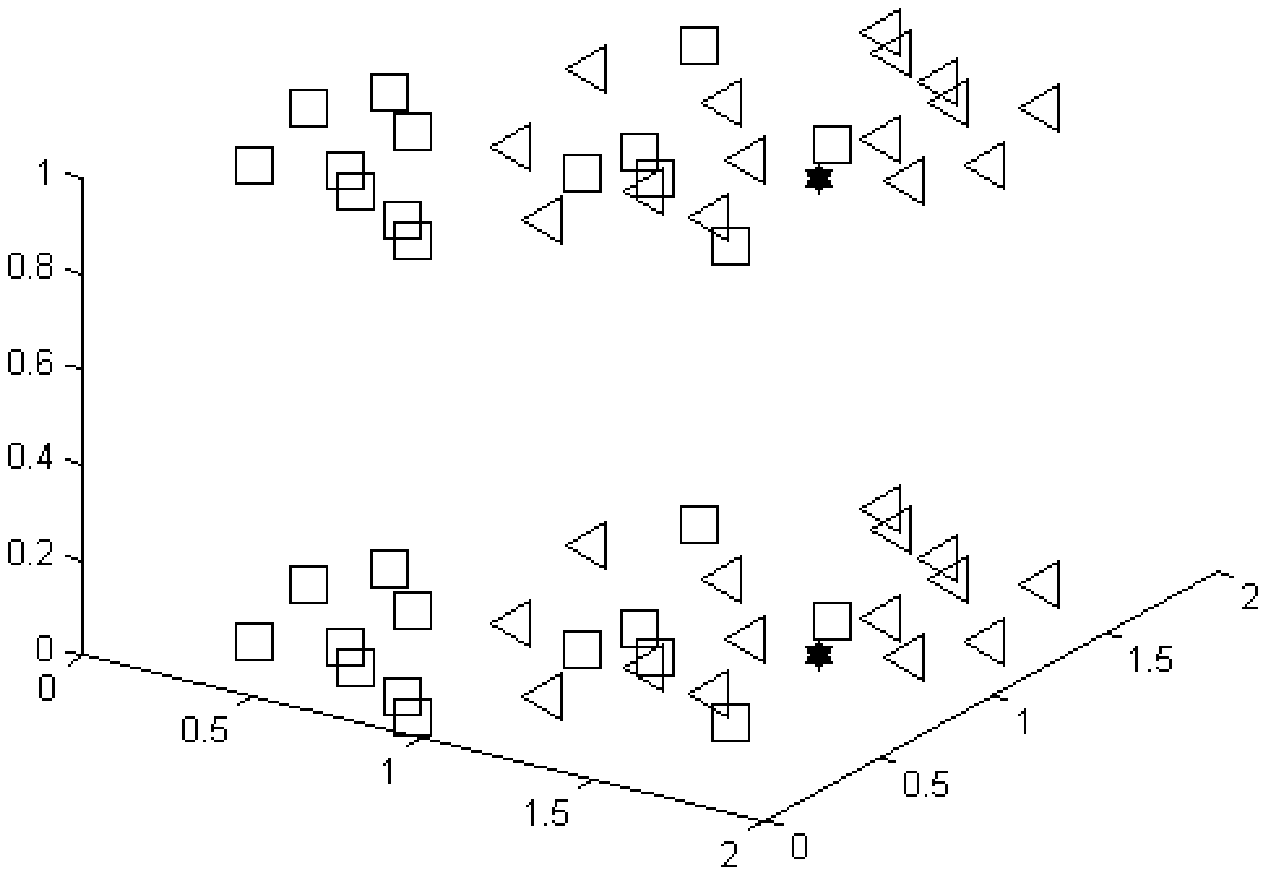}}
  \quad
  \subfloat[Binary problem in $\IR^3$, with filled points representing observations with higher cost of misclassification.]{
    \label{fig:rejectModel3}
    \includegraphics[width=0.45\linewidth]{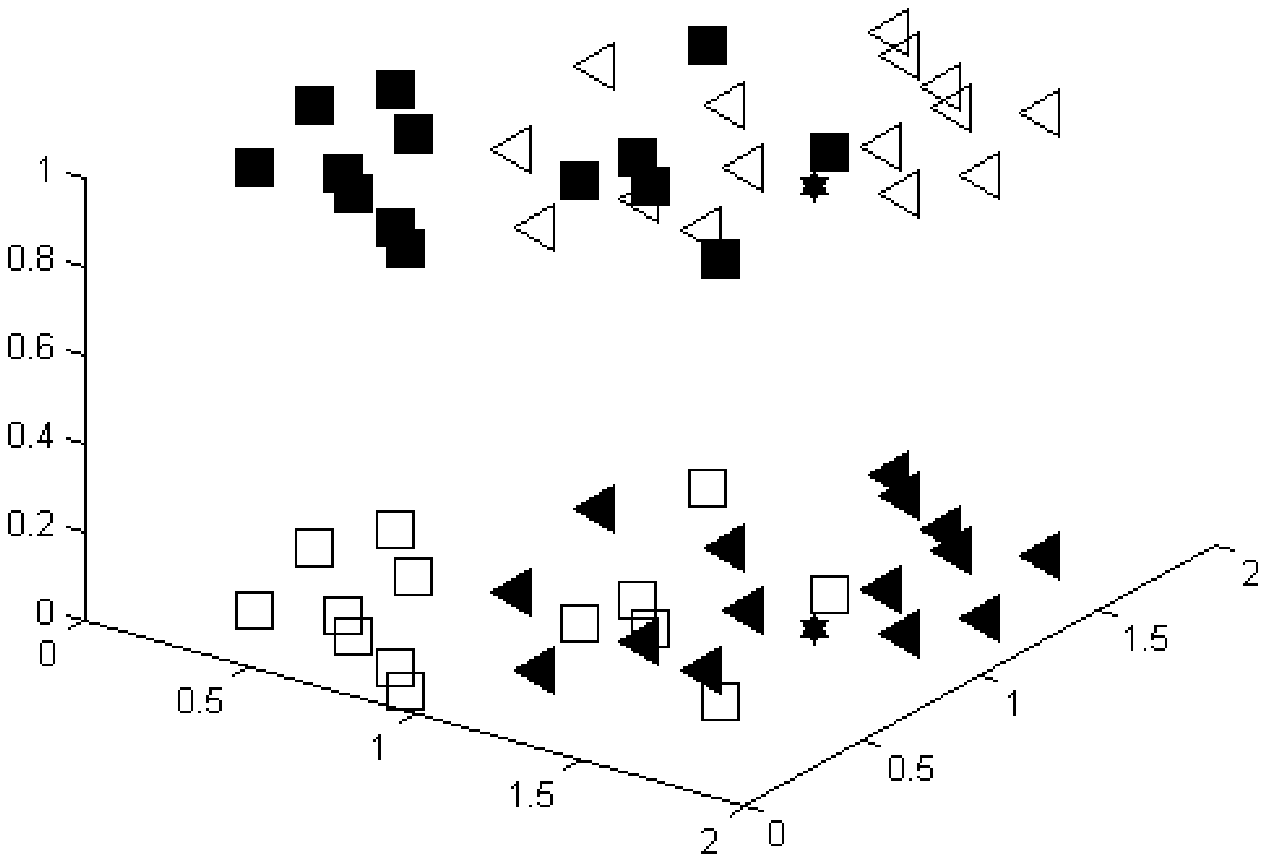}}		
  \quad
  \subfloat[Solution to the binary problem in $\IR^3$.]{
    \label{fig:rejectModel4}
    \includegraphics[width=0.45\linewidth]{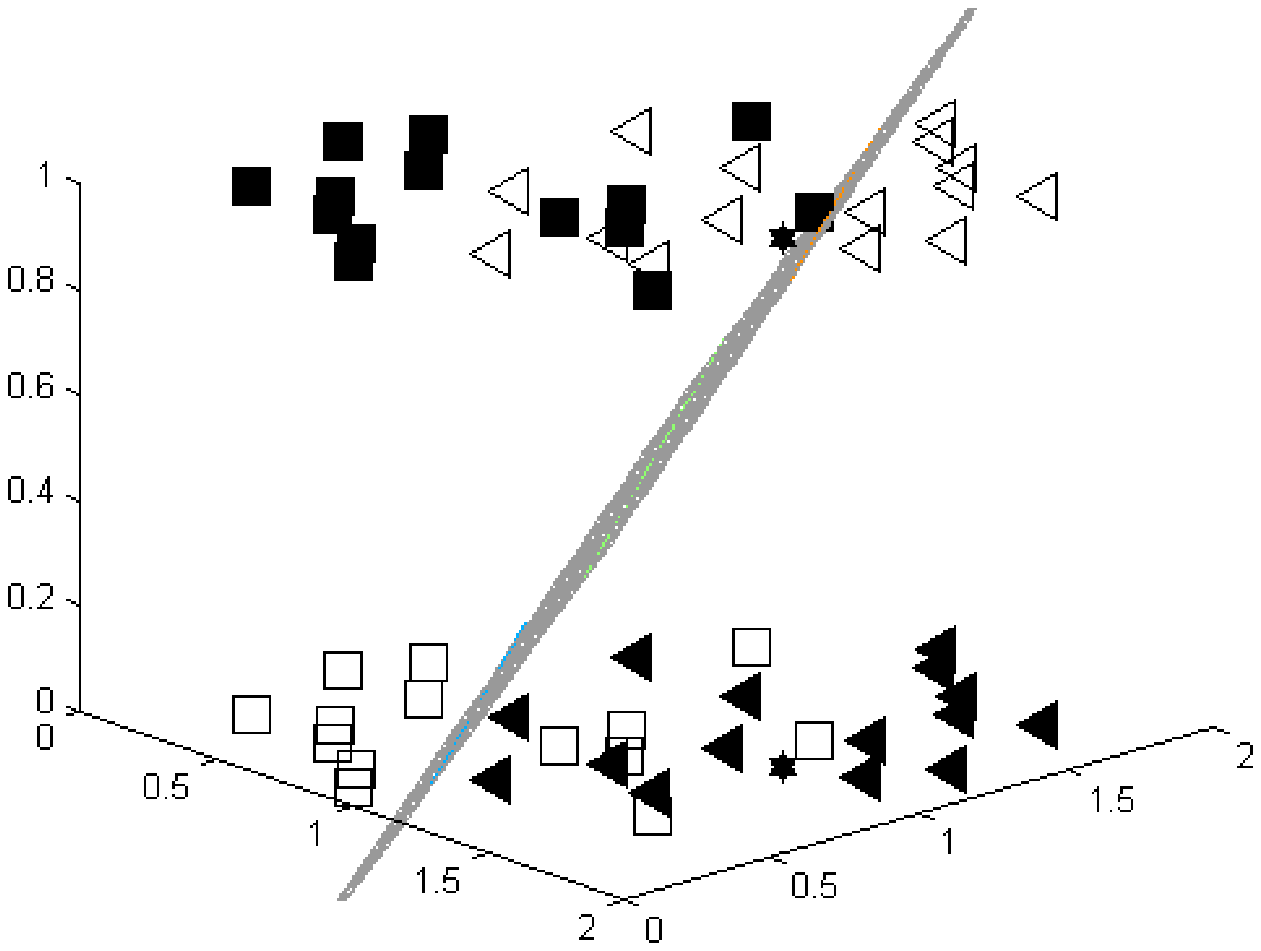}}
  \quad
  \subfloat[Solution with reject region in the original dataset.]{
    \label{fig:rejectModel5}
    \includegraphics[width=0.45\linewidth]{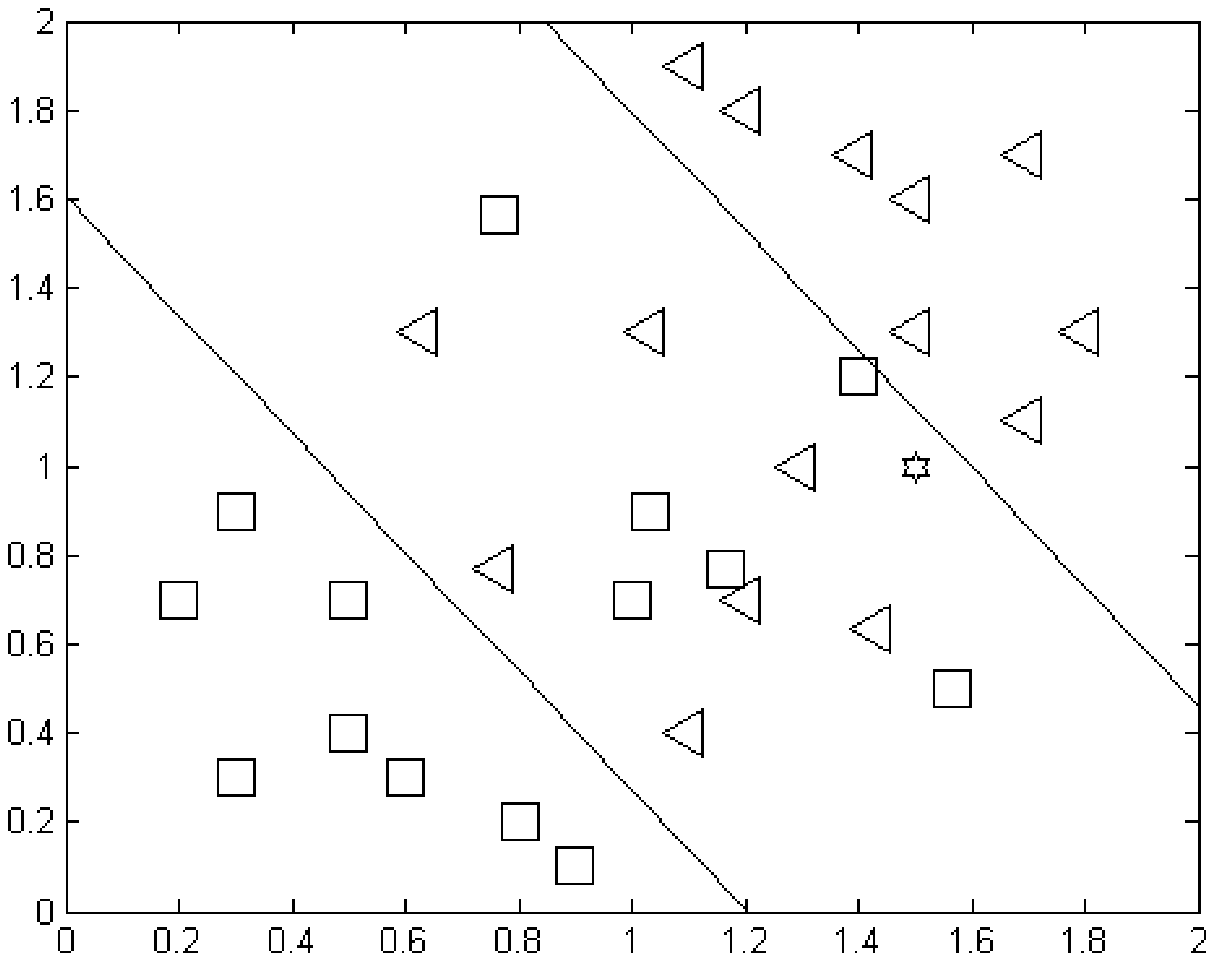}}
  \caption{Proposed reject option model in a toy example.}
  \label{fig:proposedRejectModel}
\end{figure}

We proceed exactly as in the data replication method for ordinal data. We start by transforming the data from the initial space to an extended space, replicating the data, according to the rule (see Fig.~\ref{fig:rejectModel2}):
\[
\textbf{x} \in \IR^d 
\begin{matrix}
\nearrow \\
\searrow 
\end{matrix}
\begin{matrix}
\left[\begin{smallmatrix}\textbf{x}\\ h\end{smallmatrix}\right] \in \IR^{d+1}\\
\\
\left[\begin{smallmatrix}\textbf{x}\\ 0\end{smallmatrix}\right] \in \IR^{d+1}
\end{matrix}, \text{ where $h=$ const} \in \IR^+
\]
If we design a binary classifier on the extended training data, without further considerations, one would obtain the same classification boundary in both data replicas. 
Therefore, we modify the misclassification cost of the observations according to the data replica they belong to. 
In the first replica (the extension feature assumes the value zero), we will discriminate ${\cal C}_1$ from $\{{\cal C}_{reject},{\cal C}_2\}$; therefore we give higher costs to observations belonging to class ${\cal C}_2$ than to observations belonging to class ${\cal C}_1$. This will bias the boundary towards the minimisation of errors in ${\cal C}_2$. 
In the second replica (the extension feature assumes the value $h$), we will discriminate $\{{\cal C}_1, {\cal C}_{reject}\}$ from ${\cal C}_2$; therefore we give higher costs to observations belonging to class ${\cal C}_1$ than to observations belonging to class ${\cal C}_2$. This will bias the boundary towards the minimisation of errors in ${\cal C}_1$. In Fig.~\ref{fig:rejectModel3} this procedure is illustrated by filling the marks of the observations with higher costs. 
TABLE~\ref{tab:costs} summarises this procedure.
\begin{table}
  \centering
  \small
  \begin{tabular}{c|c|c}
    \hline\hline
    Replica \# & points from ${\cal C}_1$ & points from ${\cal C}_2$\\
    \hline
    1 & $-1;C_\ell$ & $+1;C_h$\\
    2 & $-1;C_h$ & $+1;C_\ell$\\
    \hline\hline	
  \end{tabular}
\caption{\label{tab:costs}Labels and costs ($C_\ell$ and $C_h$ represent a low and a high cost value, respectively) for points in different replicas in the extended dataset.}
\end{table}

A two-class classifier can now be applied on the extended dataset, yielding a boundary separating the two classes, see Fig.~\ref{fig:rejectModel4}. The intersection of this boundary with each of the subspace replicas can be used to derive the boundaries in the original dataset, as illustrated in Fig.~\ref{fig:rejectModel5}.

Summing up, with a proper choice of costs, the data replication method can be used to learn a reject region, defined by two non-intersecting boundaries. Note that the reject region is optimised during training and not heuristically defined afterwards. Nonlinear (and non-intersecting) boundaries are treated exactly as the ordinal data scenario. Likewise, prediction follows the same rationale.
\subsubsection{Selecting the Misclassification Costs}
In the reject option scheme, one aims to obtain a minimum error while minimising the number of rejected cases. 
However, when the number of rejected cases decreases the classification error increases, and to decrease the classification error one typically has to increase the reject region. The right balance between these two conflicting goals depends on the relation of the associated costs.

Let $C_{i,q}^{(k)}$ represent the cost of erring a point ${\mathbf x}_i$ from class $k$ in data replica $q$ (or, equivalently, by hyperplane $q$).
Points from class ${\cal C}_1$ misclassified by the hyperplane 1 ($\textbf{w}^t\textbf{x}+b_1 = 0$) but correctly classified by the second hyperplane ($\textbf{w}^t\textbf{x}+b_2 = 0$) incur in a loss $C_{i,1}^{(1)}$; points from class ${\cal C}_1$ misclassified by both hyperplanes incur in a loss $C_{i,1}^{(1)} + C_{i,2}^{(1)}$. 
Likewise, points from class ${\cal C}_2$ misclassified by the hyperplane 2 ($\textbf{w}^t\textbf{x}+b_2 = 0$) but correctly classified by the first hyperplane ($\textbf{w}^t\textbf{x}+b_1 = 0$) incur in a loss $C_{i,2}^{(2)}$; points from class ${\cal C}_2$ misclassified by both hyperplanes incur in a loss $C_{i,1}^{(2)} + C_{i,2}^{(2)}$. The resulting loss matrix is given by
  \begin{center}
    \small
    \begin{tabular}{|c|c|c|c|c|}
      \cline{3-5}
      \multicolumn{2}{c|}{} & \multicolumn{3}{c|}{predicted}\\
      \cline{3-5}
      \multicolumn{2}{c|}{} & ${\cal C}_1$ & ${\cal C}_{reject}$ & ${\cal C}_2$\\
      \hline		
      \multirow{2}{*}{true} & ${\cal C}_1$  &  0 & $C_{i,1}^{(1)}$ & $C_{i,1}^{(1)} + C_{i,2}^{(1)}$\\
      & ${\cal C}_2$  & $C_{i,1}^{(2)} + C_{i,2}^{(2)}$ & $C_{i,2}^{(2)}$ & 0\\
      \hline	
    \end{tabular}
  \end{center}
The typical adoption of the same cost for erring and rejecting on the two classes leads to the following simplified loss matrix:
  \begin{center}
    \small
    \begin{tabular}{|c|c|c|c|c|}
      \cline{3-5}
      \multicolumn{2}{c|}{} & \multicolumn{3}{c|}{predicted}\\
      \cline{3-5}
      \multicolumn{2}{c|}{} & ${\cal C}_1$ & ${\cal C}_{reject}$ & ${\cal C}_2$\\
      \hline			
      \multirow{2}{*}{true}	& ${\cal C}_1$  &  0 & $C_{low}$ & $C_{high}$\\
      & ${\cal C}_2$  & $C_{high}$ & $C_{low}$ & 0\\
      \hline	
    \end{tabular}
  \end{center}
Therefore, $C_{reject} = \frac{ C_{low}}{C_{high}} = w_r$ is the cost of rejecting (normalised by the cost of erring).
The data replication method with reject option tries to minimizes the empirical risk $w_rR+E$, where $R$ accounts for the rejection rate and $E$ for the misclassification rate.
\subsubsection{Prediction}
To predict the class of an unseen example, classify both replicas of the example in the extended dataset with the binary classifier. 
From the sequence of binary labels one can infer the predicted label on the original ordinal classes 
\begin{equation*}
\overline{{\cal C}}_1, \overline{{\cal C}}_1 \Longrightarrow {\cal C}_1\quad 
\overline{{\cal C}}_2, \overline{{\cal C}}_1 \Longrightarrow {\cal C}_{reject}\quad 
\overline{{\cal C}}_2, \overline{{\cal C}}_2 \Longrightarrow {\cal C}_2
\end{equation*}

Henceforth, the target class can be obtained by counting the number of $\overline{{\cal C}}_2$ labels in the sequence, $N_{\overline{{\cal C}}_2}$: 
if $N_{\overline{{\cal C}}_2}/2+1$ is integer, it yields the target class; otherwise the option is to reject.

\subsection{Mapping the Data Replication Method to Learning Algorithms}
\label{Mapping}
In this section the method just introduced is instantiated in two important machine learning algorithms: support vector machines and neural networks.

\subsubsection{Mapping the Data Replication Method with Reject Option to SVMs}
The learning task in a classification problem is to select a prediction function $f(\textbf{x}, \alpha)$ from a family of possible functions that minimizes the expected {\em loss}, where $\alpha$ is a parameter denoting a particular function in the set.

The SVM classification technique has been originally derived by applying the SRM (structural risk minimization) principle to a two-class problem using the 0/1 (indicator) loss function: 

$L(\textbf{x}, \alpha, y) = \begin{cases} 0, \ \mbox{ if } f(\textbf{x},\alpha) = y\\ 1, \ \mbox{ if } f(\textbf{x},\alpha) \not= y \end{cases}$

The simplest generalization of the indicator loss function to classification with reject option is the following loss function
\[
L(\textbf{x}, \alpha, y) = \begin{cases} 0, \ \mbox{ if } f(\textbf{x},\alpha) = y\\ 
w_r, \ \mbox{ if } f(\textbf{x},\alpha) = reject\\
1, \ \mbox{ if } f(\textbf{x},\alpha) \not= y \text{ and } f(\textbf{x},\alpha) \neq reject \end{cases}
\]
where $w_r$ denotes the cost of rejection (with the cost of erring normalized to 1). Obviously $0\leq w_r \leq 1$.
The corresponding expected risk is $w_rP(reject)+P(error)$.
The expression of the empirical risk is $w_rR+E$.

Let us formulate the problem of classifying with reject option in the spirit of SVMs.
Starting from the generalization of the two-class separating hyperplane presented in the beginning of previous section, let us look for $2$ parallel hyperplanes represented by vector $\textbf{w}\in \IR^p$ and scalars $b_1, b_2$,
such that the feature space is divided into $3$ regions by the decision boundaries
$\textbf{w}^t\textbf{x}+b_r = 0, \ r = 1,2$.

A pair of parallel hyperplanes which minimizes the empirical risk can be obtained by minimizing the following functional (where $\mbox{ sgn } (x)$ returns $+1$ if $x$ is greater than zero; $0$ if $x$ equals zero; $-1$ if $x$ is less than zero)
\begin{equation}
\label{eq:eqno2}
\min_{\textbf{w}, b_i, \xi_i}	\quad \frac{1}{2}\textbf{w}^t\textbf{w} + 
C \sum_{q=1}^{2} \sum_{k=1}^{2} \sum_{i=1}^{\ell_k} C_{i,q}^{(k)}\mbox{ sgn }(\xi_{i,q}^{(k)})
\end{equation}
under the constraints
$$ 
\begin{array}{ll}
								-(\textbf{w}^t\textbf{x}_i^{(1)}+ b_{1})& \geq +1-\xi_{i,1}^{(1)}\\
								+(\textbf{w}^t\textbf{x}_i^{(2)}+ b_{1})& \geq +1-\xi_{i,1}^{(2)}\\
								-(\textbf{w}^t\textbf{x}_i^{(1)}+ b_{2})& \geq +1-\xi_{i,2}^{(1)}\\
								+(\textbf{w}^t\textbf{x}_i^{(2)}+ b_{2})& \geq +1-\xi_{i,2}^{(2)}\\
								 \xi_{i,q}^{(k)} \geq 0 &
 \end{array}
 $$
In practice the regularization term $\mbox{ sgn }(\xi_{i,q}^{(k)})$ is usually replaced by $\xi_{i,q}^{(k)}$ mainly for computational efficiency.

It is important to note that, although the formulation was constructed from the two-class SVM, it is no longer solvable with the same algorithms. 
Let us now examine the mapping of the data replication method with reject option on SVMs, which is solvable with a single standard binary SVM classifier. 

\paragraph*{\bf The rejoSVM algorithm}
The insight gained from studying the toy example paves the way for the formal presentation of the instantiation of the data replication method with reject region in SVMs, rejoSVM.

Following the same procedure delineated in~\cite{JaimeJMLR2007}, it is straightforward to conclude that the formulation corresponding to the mapping of the data replication method with reject option in SVMs results in 
\begin{equation}
\label{eq:eqno3}
\min_{\textbf{w}, b_i, \xi_i}	\quad \frac{1}{2}\textbf{w}^t\textbf{w} + \frac{1}{2}\frac{1}{h^2} (b_2-b_1)^2+ 
C\sum_{q=1}^{2} \sum_{k=1}^{2} \sum_{i=1}^{\ell_k} C_{i,q}^{(k)}\mbox{ sgn }(\xi_{i,q}^{(k)})
\end{equation}
with $b_2 = b_1 + w_{p+1}h$ and with the same set of constraints as in~\eqref{eq:eqno2}.

This formulation for the high-dimensional data set matches the previous formulation~\eqref{eq:eqno2} up to an additional regularization member in the objective function. 
This additional member is responsible for the unique determination of the thresholds~\cite{JaimeJMLR2007}.
We see that the rejoSVM captures the essence of the SRM of SVMs, while being solvable with existing binary SVM classifiers.

\subsubsection{Mapping the Data Replication Method with Reject Option to Neural Networks}
\begin{figure*}[!htp]
  \centering
  \begin{tikzpicture}
  [rounded corners=2pt,scale=.8,
  arrow/.style={line width=1pt,->,>=stealth}]

  \filldraw[lightgray] (0,0) ellipse (200pt and 80pt);
  \draw (0,2.5) node {\small Generic neural network};
  
  \draw (-5,1.3) node {$\mathbf{x}_1$};        \filldraw[black] (-4.7,1.3) circle (2pt and 2pt);
  \draw (-4.7,0) node {$\vdots$};
  \draw (-5,-1.5) node {$\mathbf{x}_p$};       \filldraw[black] (-4.7,-1.5) circle (2pt and 2pt);
  \draw (-5.3,-2.5) node {$\mathbf{x}_{p+1}$}; \filldraw[black] (-4.7,-2.5) circle (2pt and 2pt);
  \draw (-4.7,-2.8) node {$\vdots$};
  \draw (-5.6,-3.3) node {$\mathbf{x}_{p+2K-3}$}; \filldraw[black] (-4.7,-3.3) circle (2pt and 2pt);
  \draw (1.5,-2.8) node {$\vdots$};

  \filldraw[thick,fill=gray]  (-1,.8) rectangle (-4,1.8);
  \filldraw[thick,fill=white] (-1.2,.9) rectangle (-2.6,1.7);
  \filldraw[thick,fill=white] (-3.3,1.3) circle (11pt and 11pt);
  \draw (-3.3,1.3) node {\Large +};
  \draw (-3.3,2.3) node {\scriptsize bias};
  \draw (-1.9,1.3) node {\tiny \begin{tabular}{c}activaction\\function\\$f_1$\end{tabular}};
  \draw (-2.6,0) node {$\vdots$};

  \filldraw[thick,fill=gray]  (-1,-1) rectangle (-4,-2);
  \filldraw[thick,fill=white] (-1.2,-1.1) rectangle (-2.6,-1.9);
  \filldraw[thick,fill=white] (-3.3,-1.5) circle (11pt and 11pt);
  \draw (-3.3,-1.5) node {\Large +};
  \draw (-3.3,-0.5) node {\scriptsize bias};
  \draw (-1.9,-1.5) node {\tiny \begin{tabular}{c}activaction\\function\\$f_1$\end{tabular}};

  \draw (-.6,-.1) node {$\ldots$};

  \filldraw[thick,fill=gray]  (3,0.3) rectangle (0,1.3);
  \filldraw[thick,fill=white] (2.8, 0.4) rectangle (1.4,1.2);
  \filldraw[thick,fill=white] (0.7, 0.8) circle (11pt and 11pt);
  \draw (0.7,0.8) node {\Large +};
  \draw (0.7,1.7) node {\scriptsize bias};
  \draw (2.1,.8) node {\tiny \begin{tabular}{c}activaction\\function\\$f_{N-2}$\end{tabular}};
  \draw (1.5,0) node {$\vdots$};

  \filldraw[thick,fill=gray]  (3,-1.5) rectangle (0,-.5);
  \filldraw[thick,fill=white] (2.8, -1.4) rectangle (1.4,-.6);
  \filldraw[thick,fill=white] (0.7, -1) circle (11pt and 11pt);
  \draw (0.7,-1) node {\Large +};
  \draw (0.7,-0.1) node {\scriptsize bias};
  \draw (2.1,-1) node {\tiny \begin{tabular}{c}activaction\\function\\$f_{N-2}$\end{tabular}};

  \filldraw[thick,fill=gray]  (6.5,-0.5) rectangle (3.5,0.5);
  \filldraw[thick,fill=white] (6.3,-0.4) rectangle (4.9,0.4);
  \filldraw[thick,fill=white] (4.2, 0) circle (11pt and 11pt);
  \draw (4.2, 0) node {\Large +};
  \draw (4.2,1.1) node {\scriptsize bias};
  \draw (5.6, 0) node {\tiny \begin{tabular}{c}activaction\\function\\$f_{N-1}$\end{tabular}};

  \filldraw[thick,fill=gray]  (10.0,-1.5) rectangle (7.0,-0.5);
  \filldraw[thick,fill=white] ( 9.8,-1.4) rectangle (8.4,-0.6);
  \filldraw[thick,fill=white] ( 7.7,-1.0) circle (11pt and 11pt);
  \draw (7.7, -1.0) node {\Large +};
  \draw (7.7,  0.1) node {\scriptsize bias};
  \draw (9.1, -1.0) node {\tiny \begin{tabular}{c}activaction\\function\\$f_{N}$\end{tabular}};

  \draw (11.1, -1.0) node {\tiny \begin{tabular}{c}binary\\classifier\end{tabular}};

  \draw[arrow] (-3.3, 2.2) -- (-3.3, 1.7);
  \draw[arrow] (-3.3,-0.6) -- (-3.3,-1.1);
  \draw[arrow] ( 0.7, 1.6) -- ( 0.7, 1.1);
  \draw[arrow] ( 0.7,-0.2) -- ( 0.7,-0.7);
  \draw[arrow] ( 4.2, 0.9) -- ( 4.2, 0.3);
  \draw[arrow] ( 7.7,-0.1) -- ( 7.7,-0.7);

  \draw[arrow] (9.8,-1.0) -- (10.5,-1.0);
  
  \draw[arrow] (-4.7, 1.3) -- (-3.7,1.3);
  \draw[arrow] (-4.7, 1.3) -- (-3.7,-1.5);
  \draw[arrow] (-4.7,-1.5) -- (-3.7,1.3);
  \draw[arrow] (-4.7,-1.5) -- (-3.7,-1.5);

  \draw[arrow] (2.8, 0.8) .. controls (3.3, 0.8) .. (4, 0.3);
  \draw[arrow] (2.8,-1.0) .. controls (3.3,-1.0) .. (4,-0.3);

  \draw (6.9,0.2) node {\scriptsize $G(\mathbf{x})$};
  \draw[arrow] (6.5, 0.0) .. controls (7,0.0) .. (7.5,-.7);

  \draw[arrow] (-4.7,-2.5) .. controls (5,-2.5) .. (7.3,-1.);
  \draw[arrow] (-4.7,-3.3) .. controls (5,-3.3) .. (7.7,-1.4);

\end{tikzpicture}
  \caption{\label{fig:oNNmodel}Data replication method for neural networks with reject option (adapted from~\cite{JaimeJMLR2007}).}
\end{figure*}
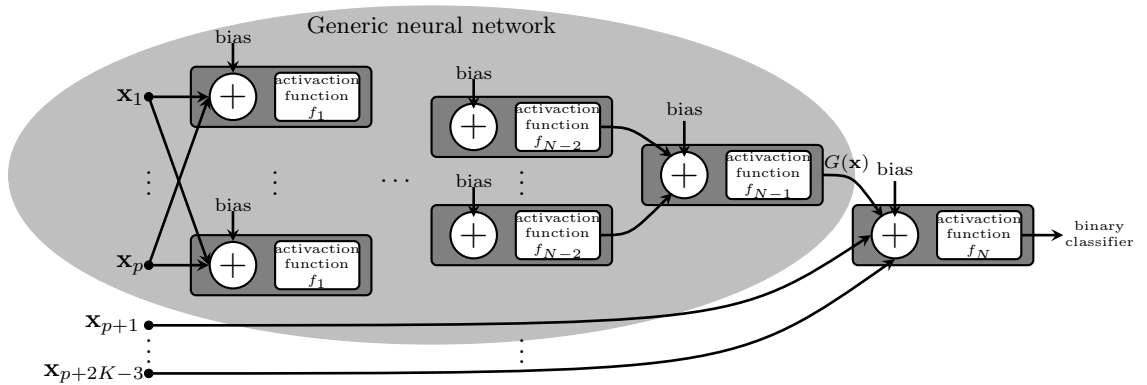
The mapping of the data replication method with reject option to NNs, rejoNN, is easily accomplished with the architecture proposed for ordinal data in~\cite{JaimeJMLR2007}.
Nonintersecting boundaries were enforced by making use of a partially linear function $\overline{G}(\overline{\textbf{x}}) = G(\textbf{x}) + \underline{\textbf{w}}^t\textbf{e}_i$ defined in the extended space.  
Setting $G (\textbf{x})$ as the output of a neural network, a flexible architecture for classification with reject option can be devised, as represented diagrammatically in Fig.~\ref{fig:oNNmodel}.

For the mapping of the data replication method with reject option in SVMs and NNs, rejoSVM and rejoNN, if we allow the samples in all the classes to contribute to each threshold, the order inequalities on the thresholds are satisfied automatically, in spite of the fact that such constraints on the thresholds are not explicitly included in the formulation. The proof follows closely the derivation presented in~\cite{JaimeJMLR2007} for the oNN algorithm.

\subsection{Classifying ordinal data with reject option---a general framework}
Although the reject option is usually only considered on binary data, it makes sense to extend it to multiclass data. In particular, the proposed approach extends nicely to ordinal data. In settings where we have $K$ ordered classes, it may be interesting to define $K-1$ reject regions, between class $k$ and class $k+1$, $k=1,\ldots, K-1$.

In the standard data replication method for ordinal data, one would have a data replica for each boundary to be defined ($K-1$ data replicas), requiring $K-2$ extension features. Now, as we need to have two boundaries between consecutive classes, we will use $2(K-1)$ data replicas, requiring $2(K-1)-1$ extension features. The goal is to find $2(K-1)$ boundaries $\mathbf{w}^t\mathbf{x}+b_i$, $i=1,\ldots, 2(K-1)$, with reject regions defined between boundaries $2j-1$ and $2j$, $j=1,\ldots, K-1$.

Replicas $q$ and $q+1$, $q=1, 3, 5, \ldots$ will have exactly the same binary labels, but different costs. Replicas $q$ and $q+1$, $q=2, 4, 6, \ldots$ will have exactly the same costs, but different binary labels. The boundaries obtained from replicas $2q-1$ and $2q$ will both discriminate ${\cal C}_1, \ldots, {\cal C}_i$ against ${\cal C}_{i+1}, \ldots, {\cal C}_K$. TABLE~\ref{tab:ordinalcosts} summarizes this setting.

\begin{table*}[!ht]
  \begin{center}
    \begin{tabular}{|c|c|c|c|c|c|}
      \hline
      \hline
      Replica \# & points from ${\cal C}_1$  & points from  ${\cal C}_2$  & \ldots & points from  ${\cal C}_{K-1}$ & ${\cal C}_K$\\

      1 & $-1;C_\ell$ & $+1;C_h$ & $+1;C_h$ & $+1;C_h$ & $+1;C_h$\\
      2 & $-1;C_h$ & $+1;C_\ell$ & $+1;C_h$ & $+1;C_h$ & $+1;C_h$\\
      $\cdots$ & & & & & \\
      2(K-1)-1 & $-1;C_h$ & $-1;C_h$ & $-1;C_h$ & $-1;C_\ell$ & $+1;C_h$ \\
      2(K-1)   & $-1;C_h$ & $-1;C_h$ & $-1;C_h$ & $-1;C_h$ & $+1;C_\ell$\\
      \hline	
      \hline	
    \end{tabular}
  \end{center}
  \caption{\label{tab:ordinalcosts}Labels and costs ($C_\ell$ and $C_h$ represent a low and a high cost value, respectively) for points in different replicas in the extended dataset.}
\end{table*}
Similarly to the binary case, the prediction of the target class for an unseen examples uses the 
sequence of $2(K-1)$ labels $\in \{\overline{{\cal C}}_1, \overline{{\cal C}}_2\}^{2(K-1)}$ by classifying each of the $2(K-1)$ replicas in the extended dataset with the binary classifier.
The target class can be obtained by counting the number of $\overline{{\cal C}}_2$ labels in the sequence, $N_{\overline{{\cal C}}_2}$: 
if $N_{\overline{{\cal C}}_2}/2+1$ is integer, is yields the target class; otherwise the option is to reject.

\section{Two classifier approach for ordinal data with reject option}
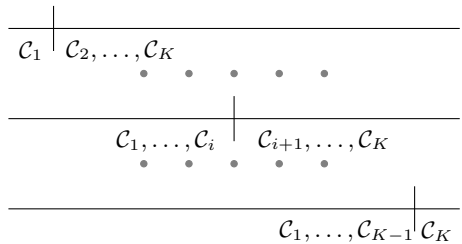
\begin{figure}[!h]
    \centering
    \begin{tikzpicture}[scale=0.6,font=\small]
      \draw (-5,0) -- (5,0);
      \draw (-4,-0.5) -- (-4,0.5);
      \draw (-4.5,-0.5) node{$\mathcal{C}_1$};
      \draw (-2.5,-0.5) node{$\mathcal{C}_2,\ldots,\mathcal{C}_K$};

      \filldraw [gray] (-2,-1) circle (2pt)
      (-1,-1) circle (2pt)
      (0,-1) circle (2pt)
      (1,-1) circle (2pt)
      (2,-1) circle (2pt);

      \draw (-5,-2) -- (5,-2);
      \draw (0,-2.5) -- (0,-1.5);
      \draw (-1.5,-2.5) node{$\mathcal{C}_1,\ldots,\mathcal{C}_i$};
      \draw (2,-2.5) node{$\mathcal{C}_{i+1},\ldots,\mathcal{C}_K$};

      \filldraw [gray] (-2,-3) circle (2pt)
      (-1,-3) circle (2pt)
      (0,-3) circle (2pt)
      (1,-3) circle (2pt)
      (2,-3) circle (2pt);

      \draw (-5,-4) -- (5,-4);
      \draw (4,-4.5) -- (4,-3.5);
      \draw (2.5,-4.5) node{$\mathcal{C}_1,\ldots,\mathcal{C}_{K-1}$};
      \draw (4.5,-4.5) node{$\mathcal{C}_K$};
      
    \end{tikzpicture}
  \caption{Transformation of an ordinal data classification problem in (K-1) binary problems.\label{fig:ordinalclassification}}
\end{figure}

In this section, and for experimental comparison purposes, we introduce an extension to ordinal data of the two-classifier approach for binary data with reject option. The extension involves a simple adaptation of the Frank and Hall~\cite{FH01} method for ordinal data.
Frank and Hall~\cite{FH01} proposed to use $(K-1)$ standard binary classifiers to address the $K$-class ordinal data problem. Toward that end, the training of the $i^{th}$ classifier is performed by converting the ordinal dataset with classes ${\cal C}_1, \ldots, {\cal C}_K$ into a binary dataset, discriminating ${\cal C}_1, \ldots, {\cal C}_i$ against ${\cal C}_{i+1}, \ldots, {\cal C}_K$ (see Fig.~\ref{fig:ordinalclassification}).
The $i^{th}$ classifier represents the test $\mathcal{C}_x > \mathcal{C}_i$.
To predict the class value of an unseen instance, the $K-1$ binary outputs are combined to produce a single estimation. 
The extension of the two classifier approach for reject option to ordinal data involves replacing the $i^{th}$ classifier in Frank and Hall method by two classifiers, both discriminating ${\cal C}_1, \ldots, {\cal C}_i$ against ${\cal C}_{i+1}, \ldots, {\cal C}_K$ but trained with different costs, exactly as given in TABLE~\ref{tab:ordinalcosts} for our proposal.
Observe that, under our approach, the $(2i-1)^{th}$ and $(2i)^{th}$ boundaries are also discriminating ${\cal C}_1, \ldots, {\cal C}_i$ against ${\cal C}_{i+1}, \ldots, {\cal C}_K$; the major difference lies in the independence of the boundaries found with Frank and Hall's method. This independence is likely to lead to intersecting boundaries.

\section{Experimental Results}
\label{sec:four}
In the following subsections, experimental results are provided for several models based on SVMs and NNs, when applied to diverse data sets, ranging from synthetic to real data, for binary and ordinal data. The set of models under comparison include the proposed rejoSVM and rejoNN methods, the standard one and two classifier approaches, and the Fumera~\cite{Fumera2002a} state of the art method.
The work was performed in a reproducible research manner, and the MATLAB code needed to reproduce all reported results is available at \url{http://www.inescporto.pt/~jsc/ReproducibleResearch.html}\footnote{Page under construction.}. The proposed rejoSVM is based on the binary SVM from the Bioinformatics Toolbox and the rejoNN is based on the Neural Network Toolbox. We thank G. Fumera for providing the source code (in C/C++) of his method.
Please note that this method is for SVMs only and the provided implementation works only with linear kernels. 

\subsection{Experiments with binary data}
The performance of the classification methods were assessed over three binary datasets. The first two were synthetically generated; the third dataset includes real data from a medical application.

\begin{figure}[!ht] 
  \centering
  \includegraphics[width=0.6\linewidth]{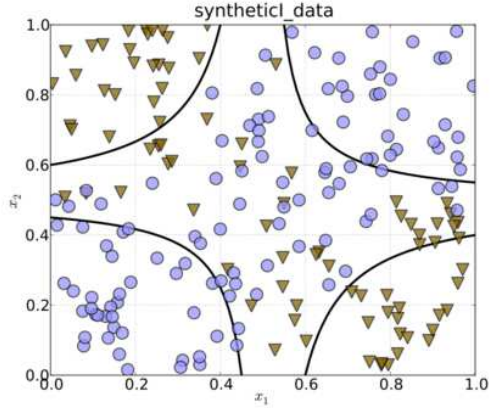}
  \caption{Sample of $200$ examples from SyntheticI dataset.}
  \label{fig:syntheticI}
\end{figure}

For the first synthetic dataset---henceforth called SyntheticI---, we began by generating $400$ example points $\textbf{x}=[x_1 \ x_2]^t$ in the unit square $[0, 1]\times [0, 1] \subset \IR^2$ according to a uniform distribution. Then, we assigned to each example $\textbf{x}$ a class $y\in \{-1, +1\}$ corresponding to
\begin{equation*}
\centering
\begin{split}
(b_{-2},b_{-1},b_0,b_1) = (-\infty;-0.5;0.25;+\infty)\\
\varepsilon_1 \sim N(0, 0.125^2)\\
\alpha = 10 (x_1 - 0.5)(x_2 - 0.5)
\end{split}
\end{equation*}
\begin{equation*}
\centering
\begin{split}
t  = \min_{r \in \{-1,0,+1\} } \left \{ r : b_{r-1} < \alpha + \varepsilon_1 < b_r \right \}\\
\varepsilon_2 \sim \mbox{Uniform}(b_{-1}, b_0)\\
y = \begin{cases} t \quad t \not=0 \\ +1 \quad t=0 \wedge  \varepsilon_2 < \alpha \\ -1 \quad t=0 \wedge \varepsilon_2 > \alpha \end{cases}
\end{split}
\end{equation*}
This distribution creates two plateau uniformly distributed and a transition zone of linearly decreasing probability, delimited by hyperbolic boundaries. 
Fig.~\ref{fig:syntheticI} depicts a sample of $100$ examples drawn according to this distribution. The two boundaries correspond to $\alpha = b_{-1}$ and $\alpha = b_{0}$.

A second synthetic dataset of $400$ points---SyntheticII---was generated from two Gaussian in $\IR^2$:
$\mathbf{y}_{-1}\sim N(\begin{bmatrix}-2\\-2\end{bmatrix},\begin{bmatrix}9 & 0\\0 & 9\end{bmatrix})+\varepsilon$ and $\mathbf{y}_{+1}\sim N(\begin{bmatrix}+2\\+2\end{bmatrix},\begin{bmatrix}25 & 0\\0 & 25\end{bmatrix})+\varepsilon$ corresponding to classes $\{-1,+1\}$ respectively, where $\varepsilon$ follows a uniform distribution in $[0.025, 0.25]$.

Finally, the third dataset, encompassing $960$ observations, expresses the aesthetic evaluation of Breast Cancer Conservative Treatment~\cite{2007JaimeAIIM}. For each patient submitted to BCCT, $30$ measurements were recorded, capturing visible skin alterations or changes in breast volume or shape. The aesthetic outcome of the treatment for each and every patient was classified in one of the four categories: Excellent, Good, Fair and Poor. 
For the experimental work with binary models, the multiclass problem was transformed into a binary one, by aggregating Excellent and Good in one class, and the Fair and Poor cases in another class. 

\subsection{Experiments with multiclass data}

\begin{figure}[!ht] 
  \centering
  \subfloat[Sample of $100$ examples from SyntheticIIIb dataset (K=5).]{\label{fig:syntheticIIIb}\includegraphics[width=0.6\linewidth]{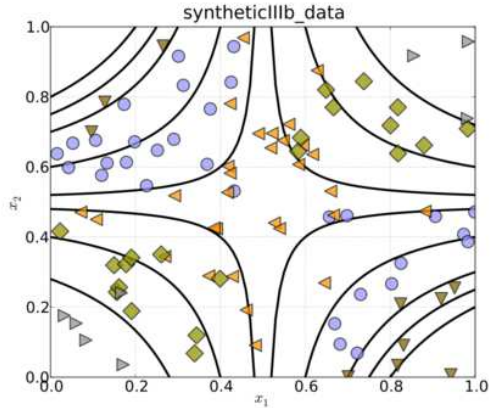}}
  \caption{SyntheticIII example points with theoretical decision boundaries.}
  \label{fig:syntheticIII}
\end{figure}

To evaluate the generalization of our approach, we extended the SyntheticI dataset into two different datasets. 
SyntheticIII dataset was generated similarly as SyntheticI but now with five classes.
\begin{multline*}
(b_{0.5},b_{1},b_{1.5},b_{2},b_{2.5},b_3,b_{3.5},b_4,b_{4.5},b_5) \\ = (-\infty;-1.5;-1.25;-1;-0.5;-0.1;0.1,0.5;1.1;+\infty)
\end{multline*}
Fig.~\ref{fig:syntheticIII} depicts a sample of $100$ examples drawn according to this distribution.

Another dataset named syntheticIV was used in our experiments.
This dataset is an extension of the syntheticII with one additional class generated accordingly to the Gaussian distribution with mean $[7 \ 7]^t$ and covariance $\Sigma=4\mathbf{I}$, where $\mathbf{I}$ is the identity matrix.

Regarding to the BCCT dataset, we used the original multiclass problem: Excellent, Good, Fair and Poor. 

We randomly split each dataset into training and test sets; in order to study the effect of varying the size of the training set, we considered three possibilities: 5\%, 25\% and 40\% of all the data available.
The splitting of the data into training and test sets was repeated 100 times in order to obtain more stable results for accuracy by averaging and also to assess the variability of this measure. The best parameterization of each model was found by `grid-search', based on a 5-fold cross validation scheme conducted on the training set. Finally, the error of the model was estimated on the test set.

The performance of a classifier with reject option can be represented by the classification accuracy achieved for any value of the reject rate (the so-called Accuracy-Reject curve). The trade-off between errors and rejections depends on the cost of a rejection $w_r$. This implies that different points of the A-R curve correspond to different values of $w_r$. 
We considered values of $w_r$ less than $0.5$, as above this value it is preferable to just try to guess randomly.

\subsection{Results}
\begin{figure*}[!htp]
  \centering
  \subfloat[5\% of training data.\label{fig:SyntheticI0ROC}]{\includegraphics[width=0.3\linewidth]{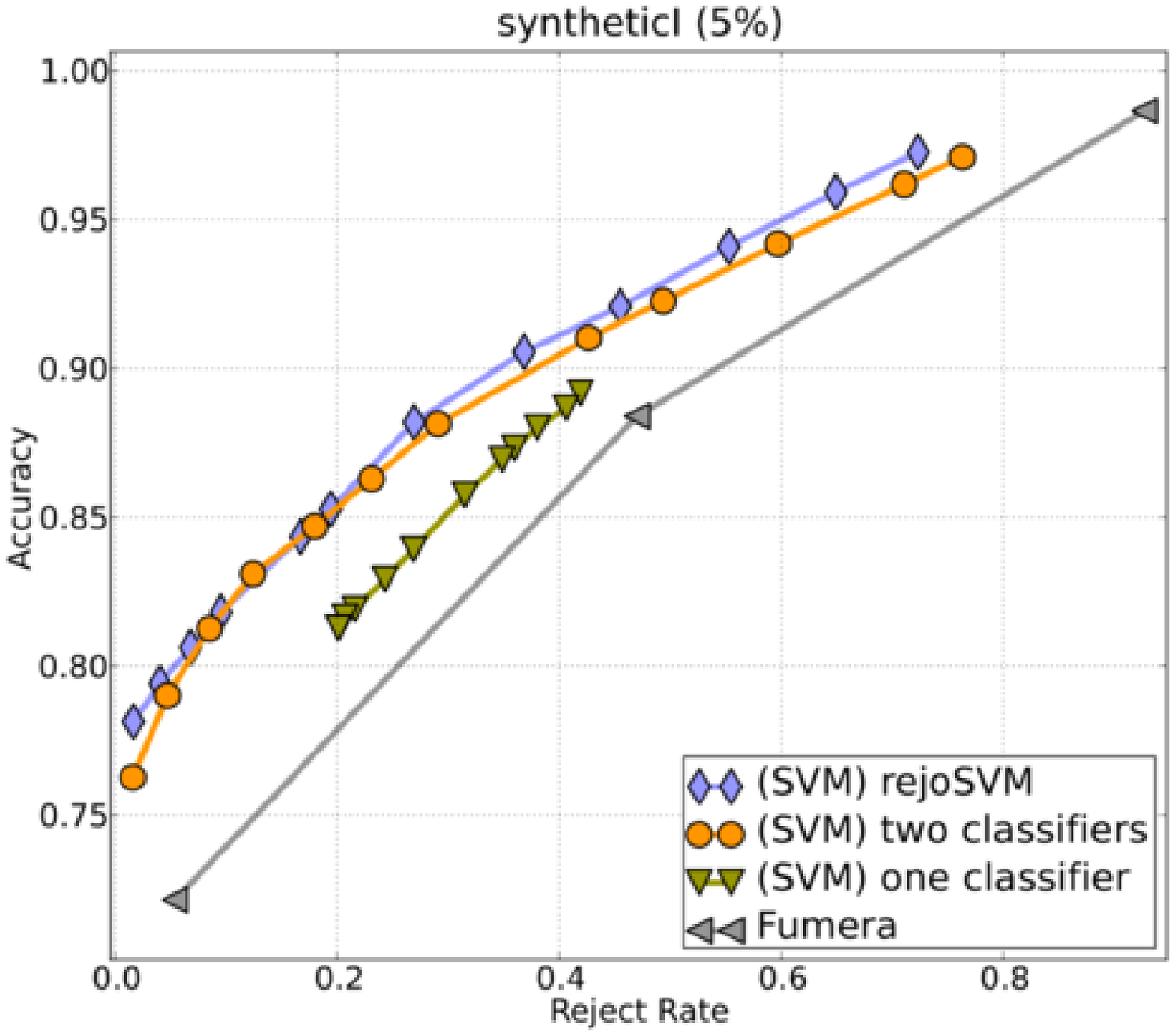}}
  \quad
  \subfloat[25\% of training data.\label{fig:SyntheticI4ROC}]{\includegraphics[width=0.3\linewidth]{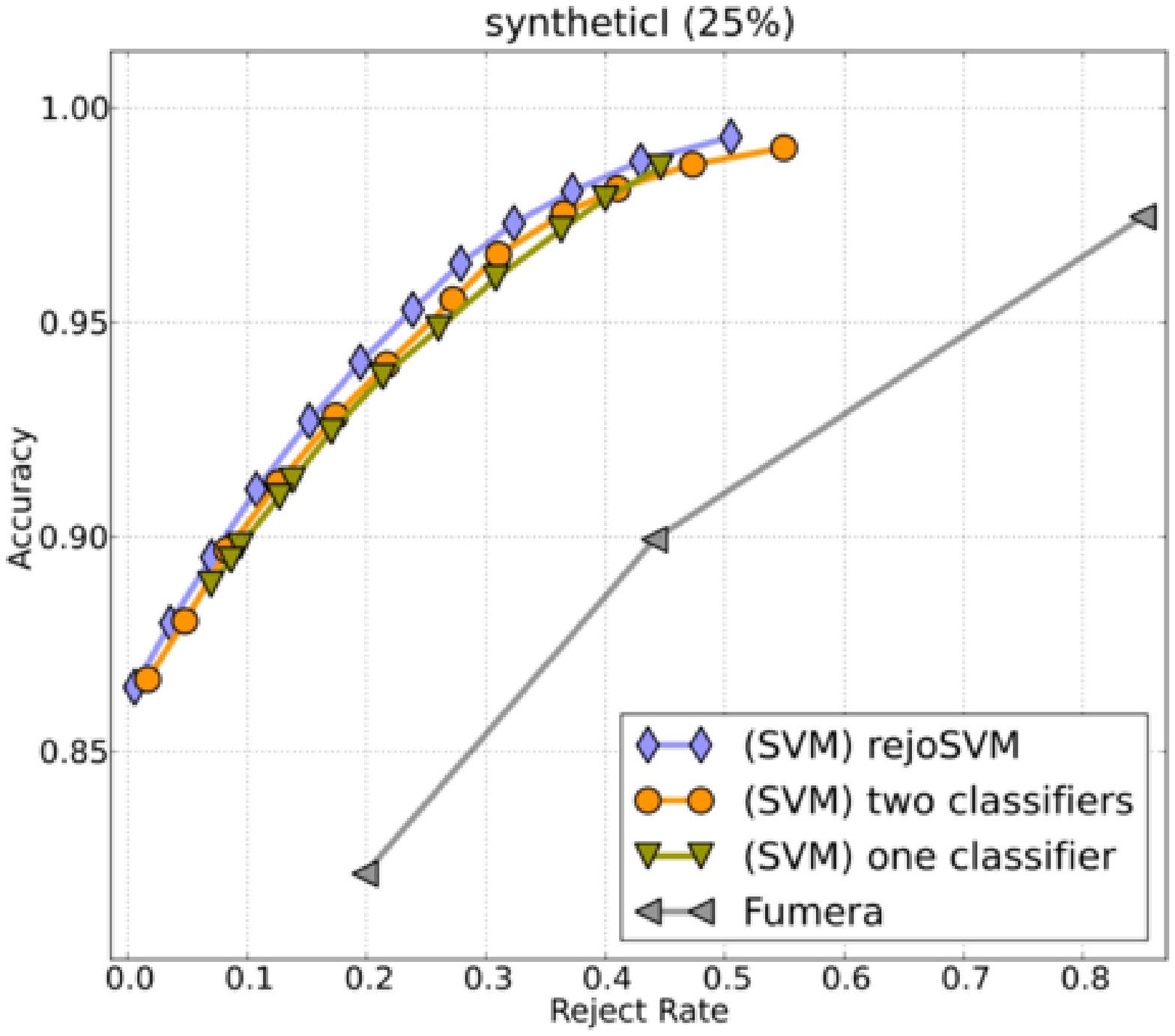}}
  \quad
  \subfloat[40\% of training data.\label{fig:SyntheticI7ROC}]{\includegraphics[width=0.3\linewidth]{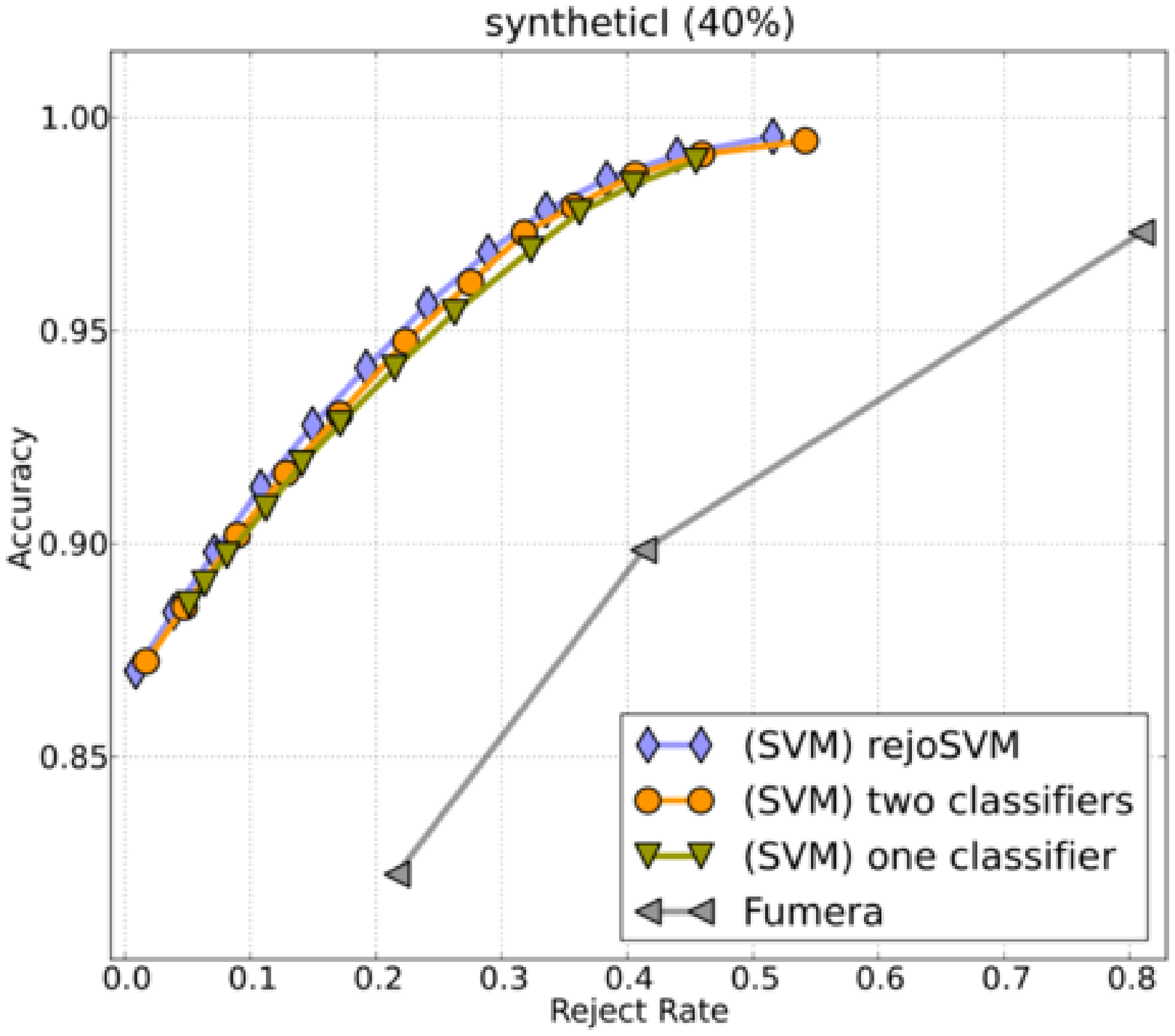}}
  \caption{The A-R curves for the syntheticI dataset (SVM methods only).\label{fig:resSyntI}}
\end{figure*}

Figures Fig.~\ref{fig:resSyntI} to Fig.~\ref{fig:resBCCT} summarise the results obtained for all datasets.
For the multiclass datasets we only include the results for SVMs since the NNs portrayed the same relative performance for the methods under comparison.

\begin{figure*}[!htp]
  \centering
  \subfloat[5\% of training data.\label{fig:SyntheticI0ROC5nn}]{\includegraphics[width=0.3\linewidth]{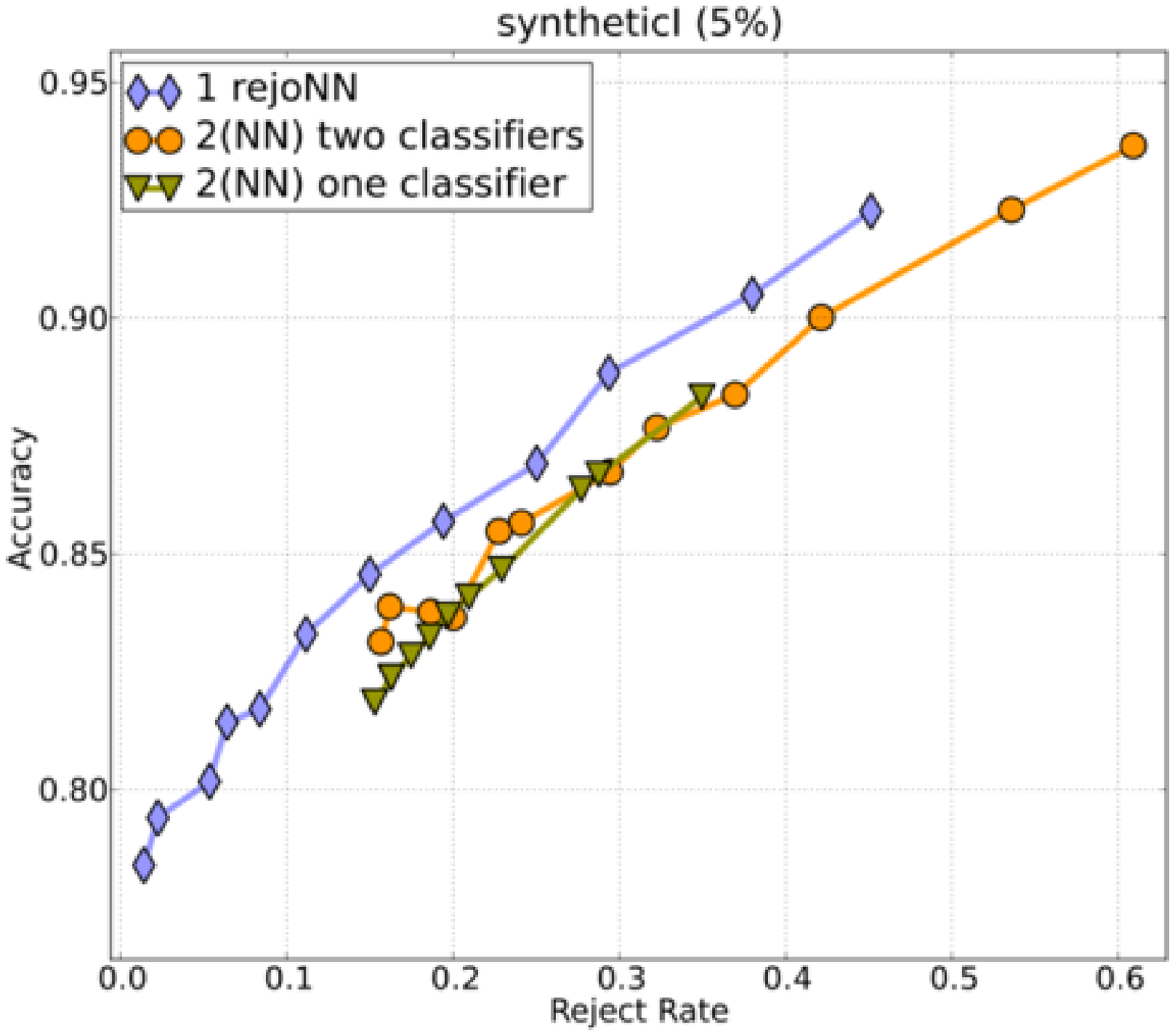}}
  \quad
  \subfloat[25\% of training data.\label{fig:SyntheticI2ROC25nn}]{\includegraphics[width=0.3\linewidth]{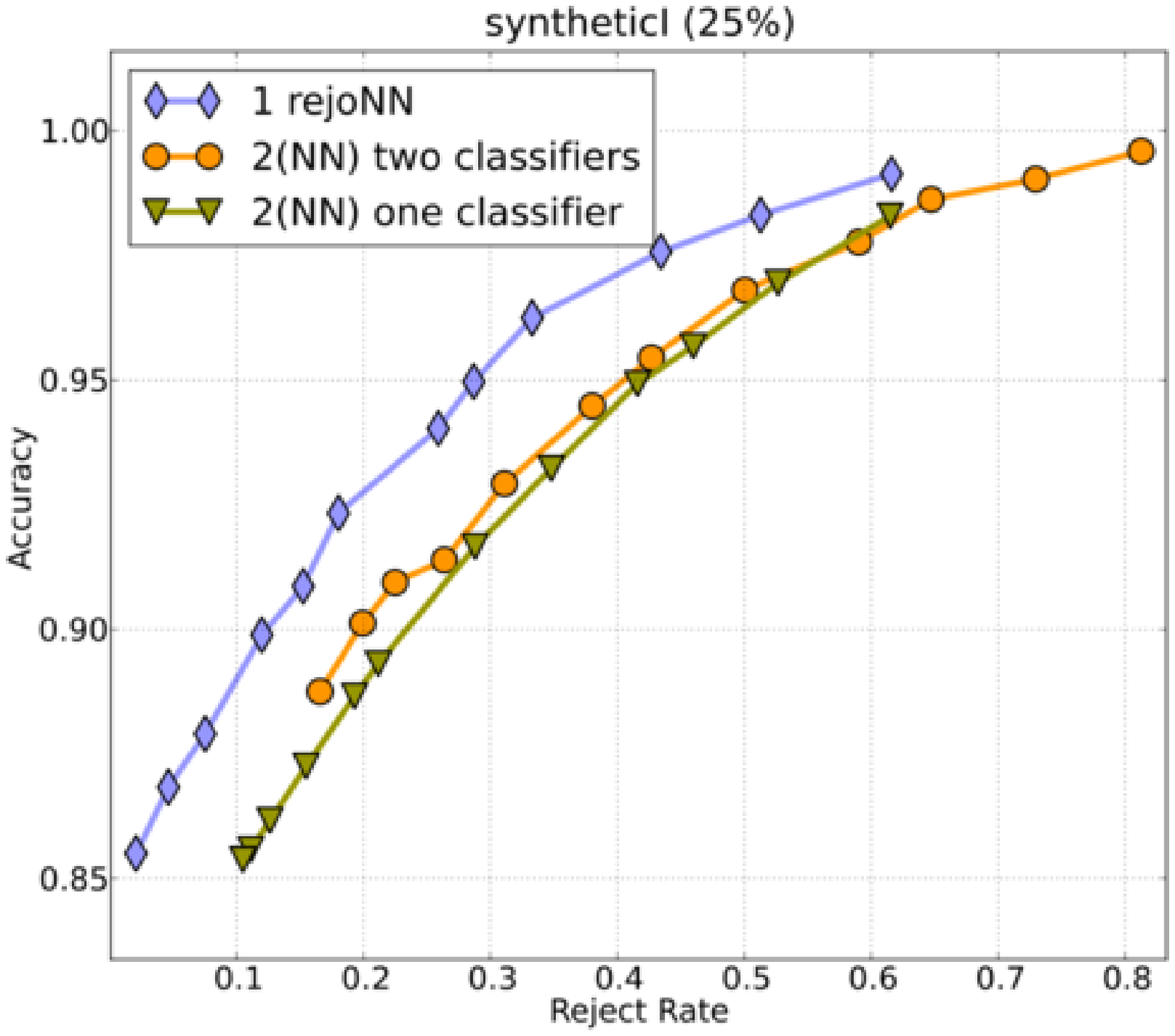}}
  \quad
  \subfloat[40\% of training data.\label{fig:SyntheticI2ROC40nn}]{\includegraphics[width=0.3\linewidth]{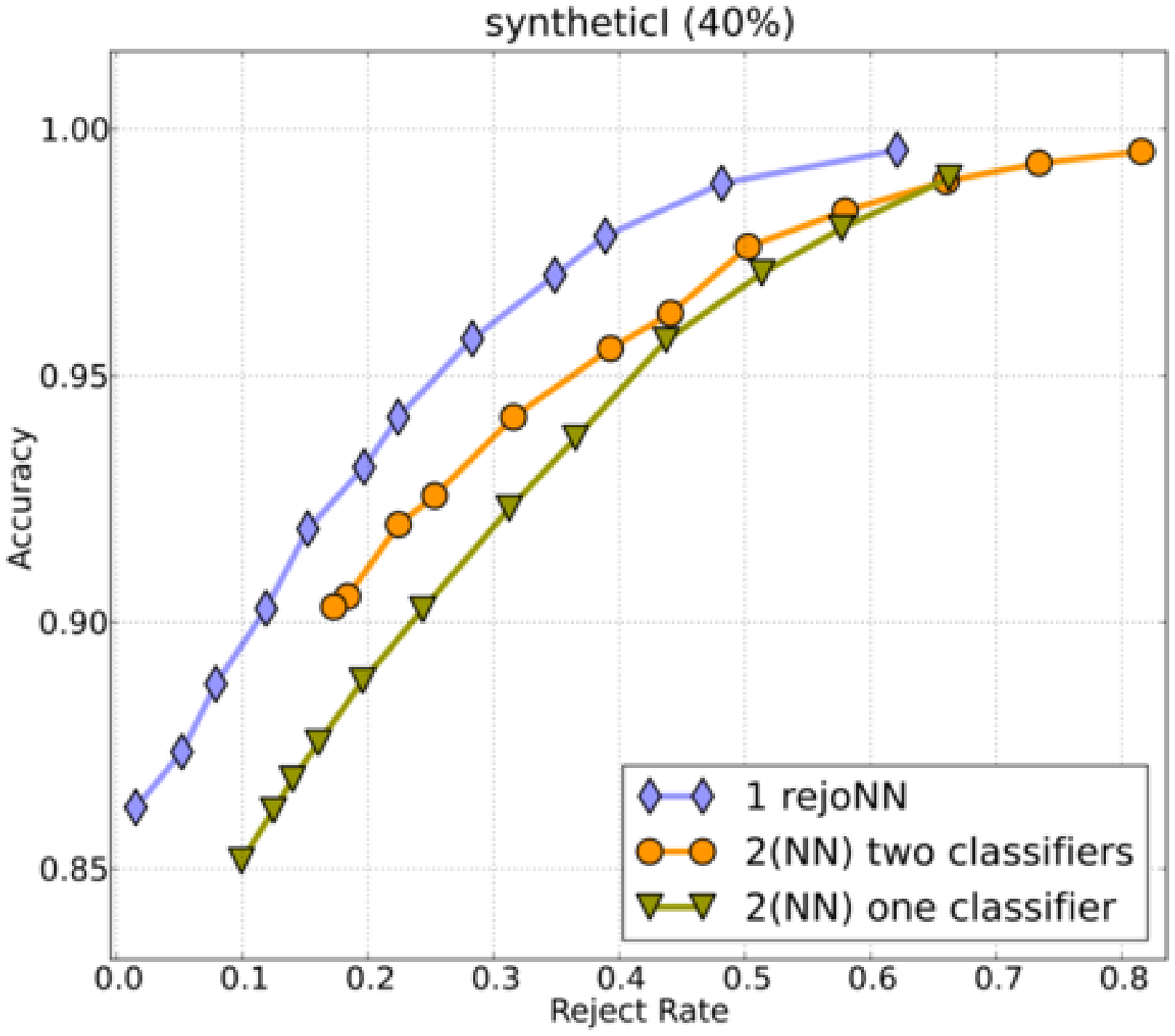}}
  \caption{The A-R curves for the syntheticI dataset (NN methods only).\label{fig:resSyntInn}}
\end{figure*}
\begin{figure*}[!htp]
  \centering
  \subfloat[5\% of training data.\label{fig:SyntheticIIaROC5}]{\includegraphics[width=0.3\linewidth]{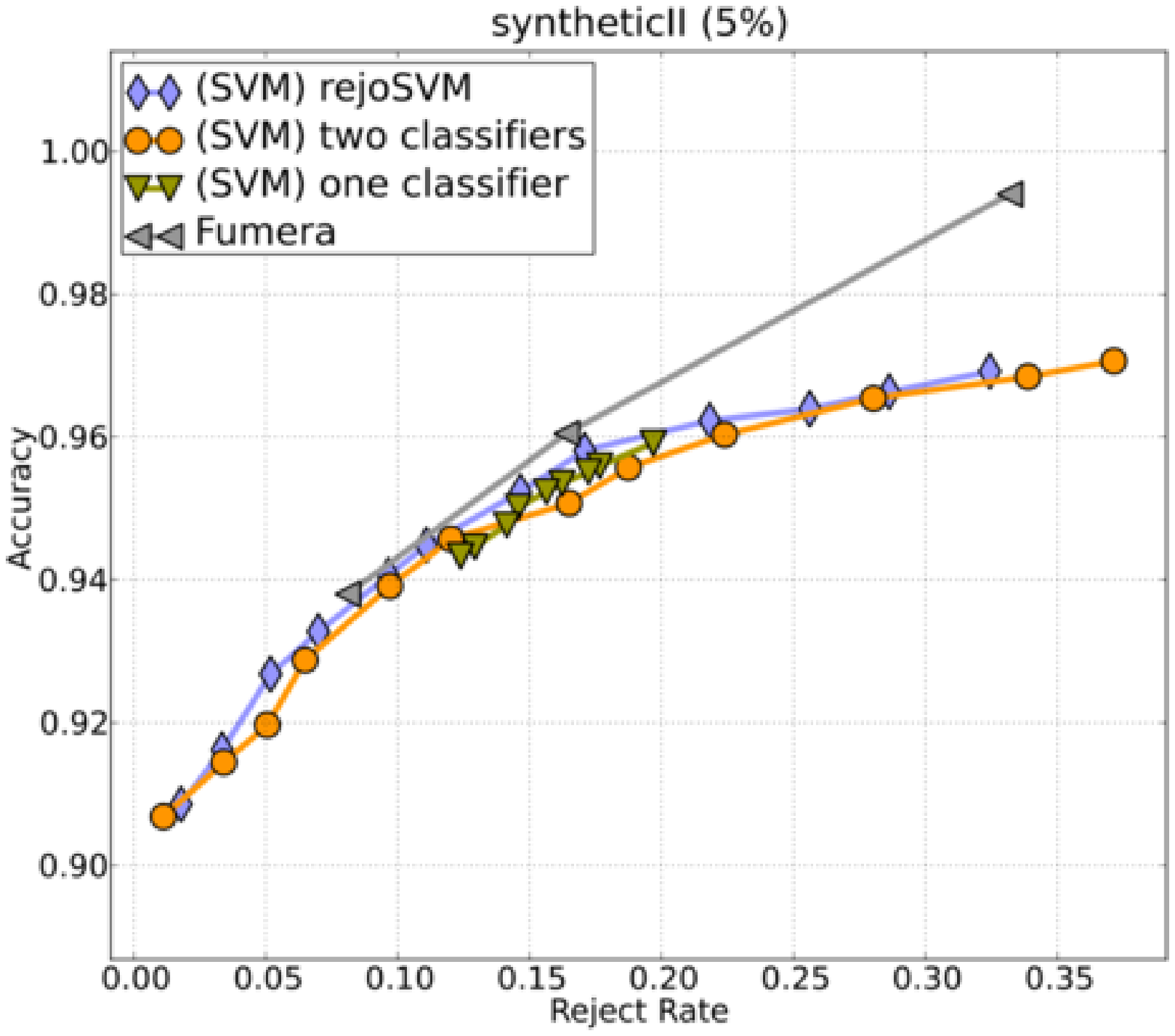}}
  \quad
  \subfloat[25\% of training data.\label{fig:SyntheticIIcROC25}]{\includegraphics[width=0.3\linewidth]{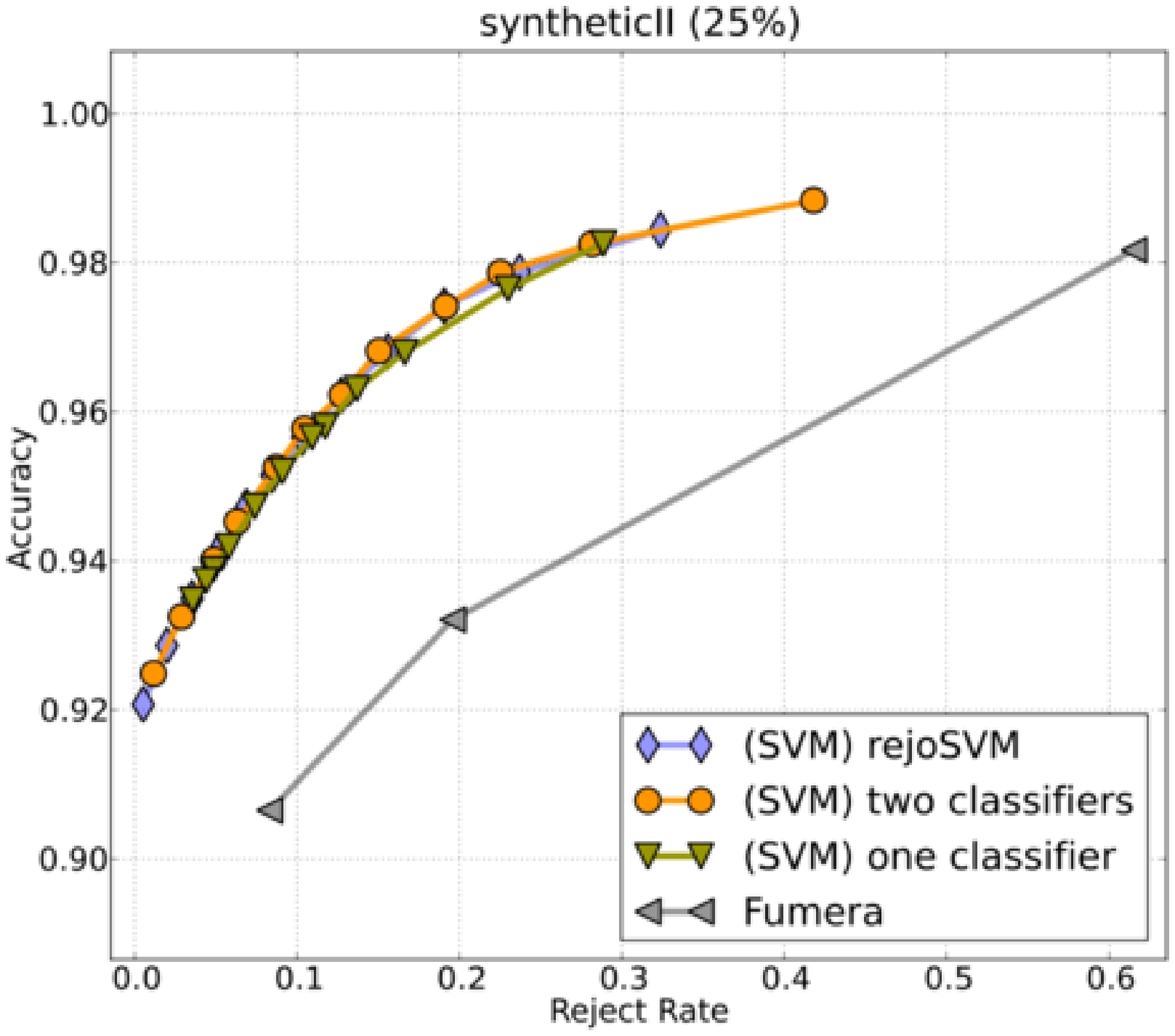}}
  \quad
  \subfloat[40\% of training data.\label{fig:SyntheticIIcROC40}]{\includegraphics[width=0.3\linewidth]{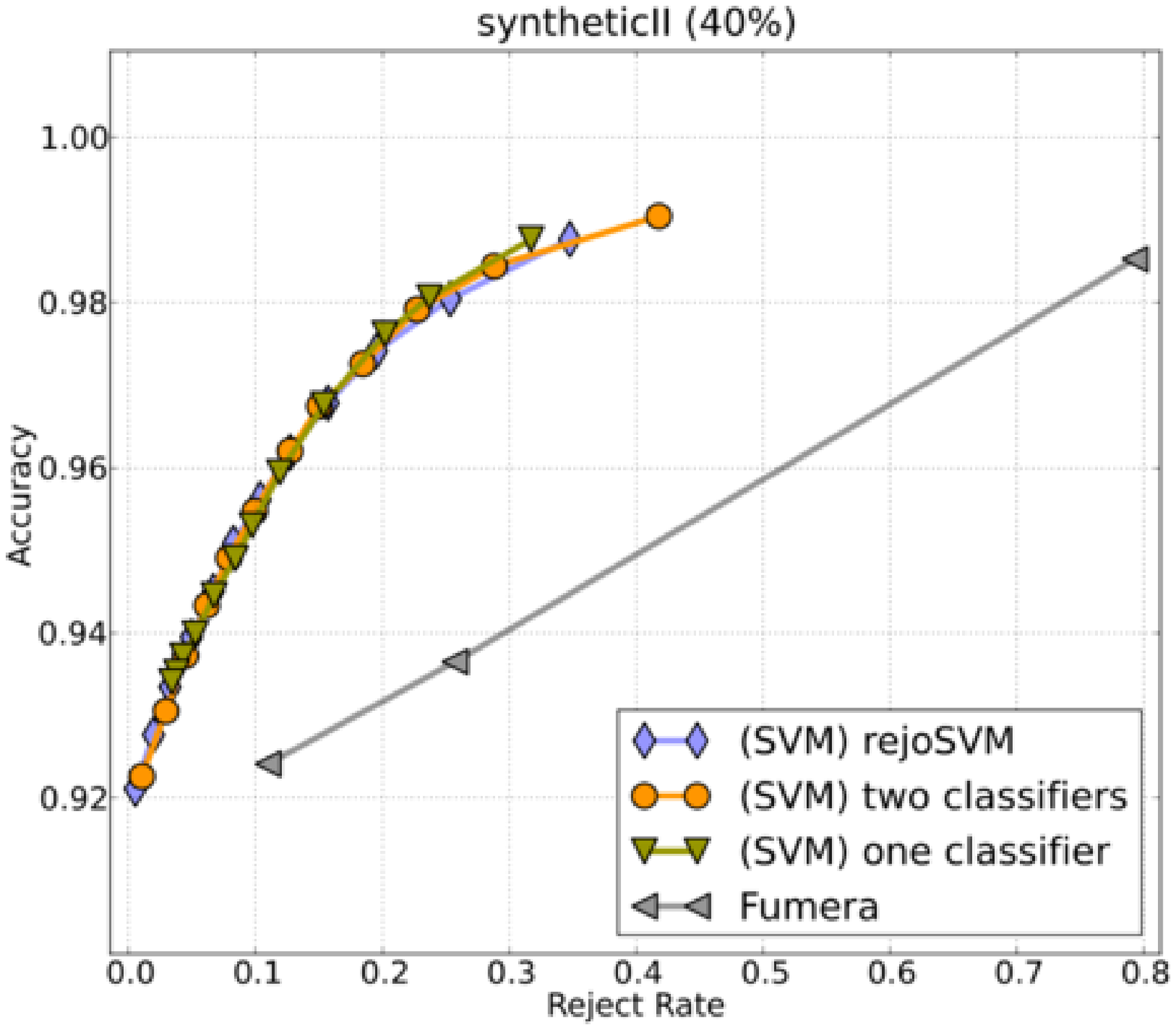}}
  \caption{The A-R curves for the syntheticII dataset (SVM methods only).\label{fig:resSyntII}}
\end{figure*}
\begin{figure*}[!htp]
  \centering
  \subfloat[5\% of training data.\label{fig:SyntheticIIaROC5nn}]{\includegraphics[width=0.3\linewidth]{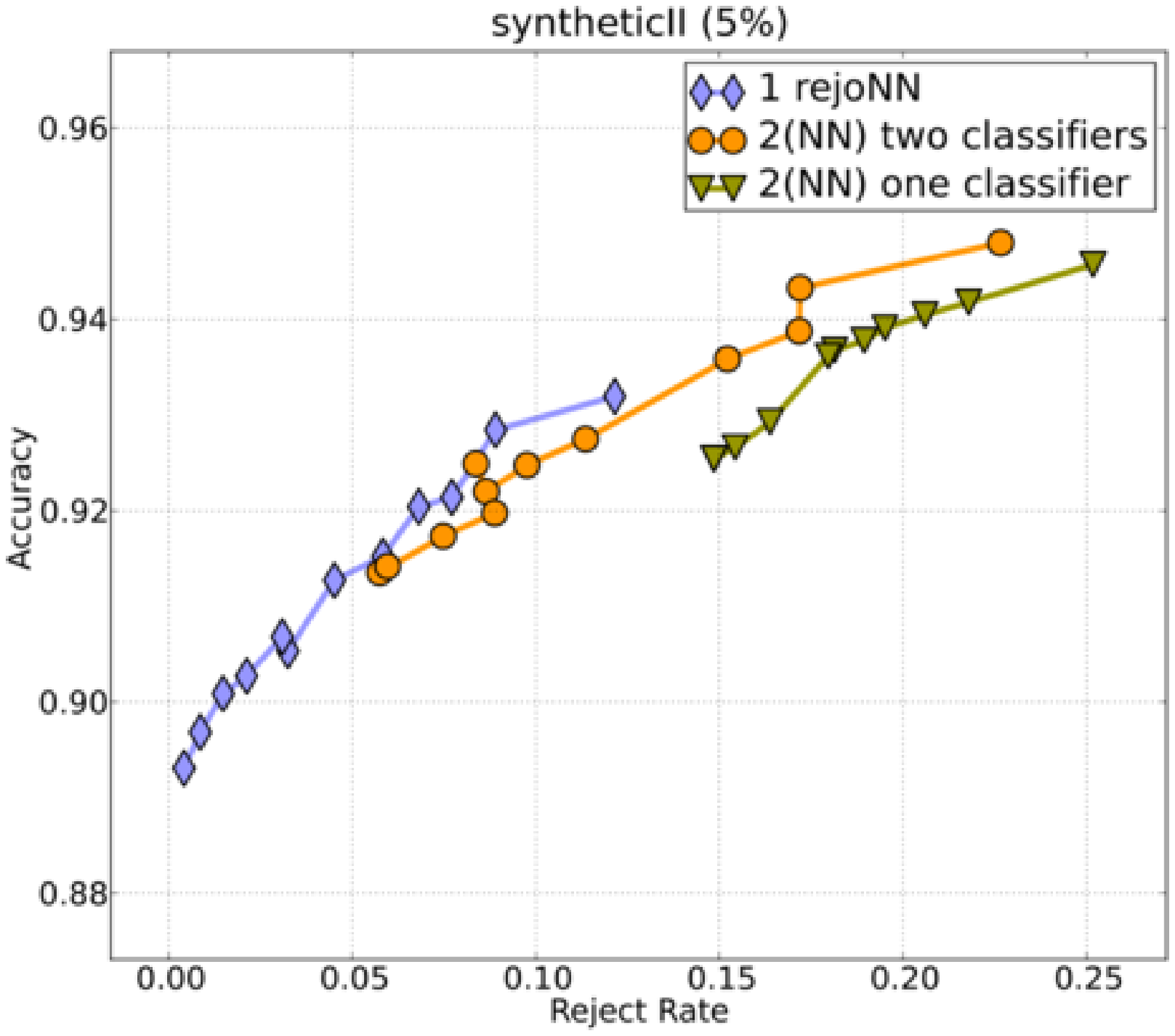}}
  \quad
  \subfloat[25\% of training data.\label{fig:SyntheticIIcROC25nn}]{\includegraphics[width=0.3\linewidth]{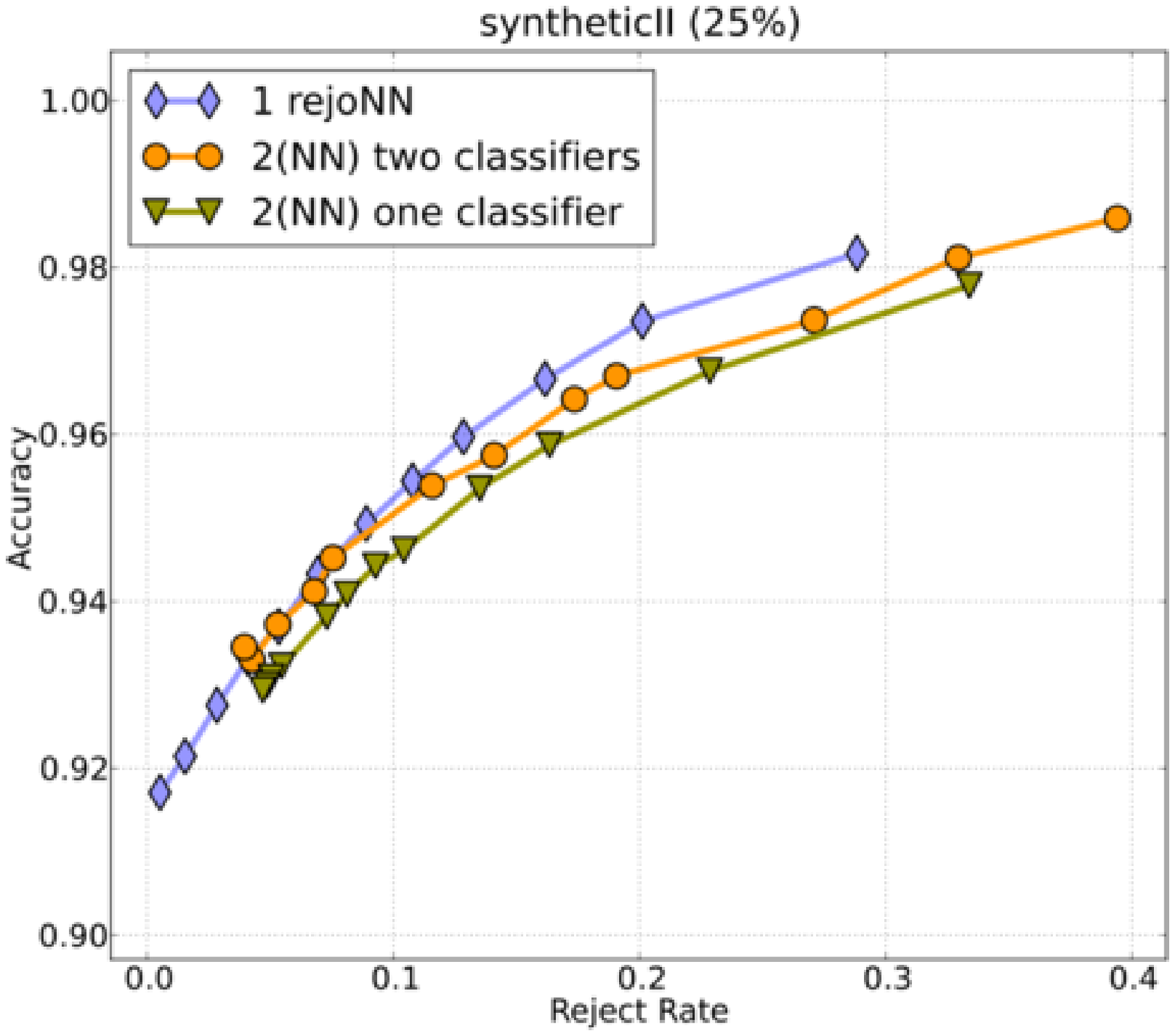}}
  \quad
  \subfloat[40\% of training data.\label{fig:SyntheticIIcROC40nn}]{\includegraphics[width=0.3\linewidth]{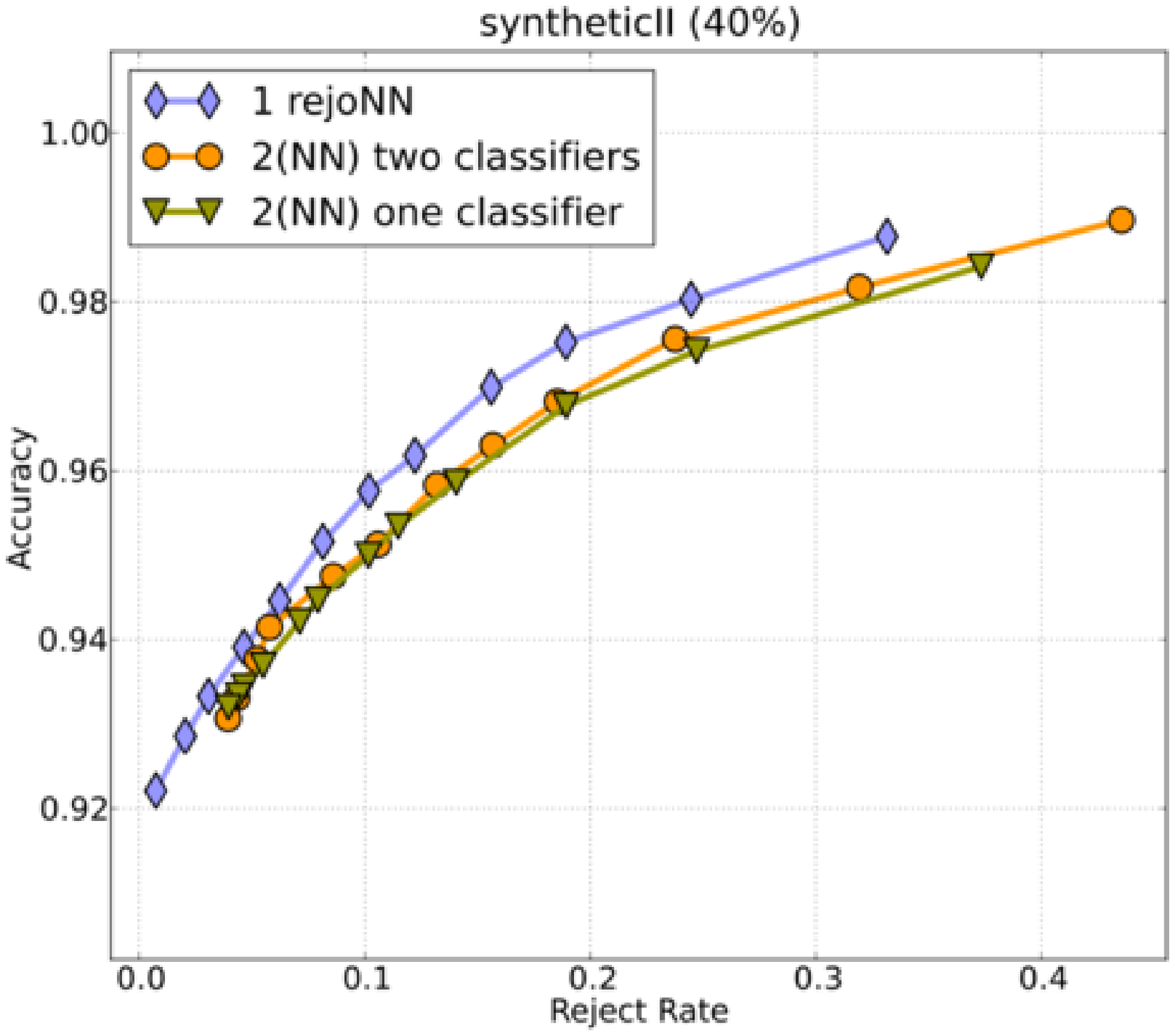}}
  \caption{The A-R curves for the syntheticII dataset (NN methods only).\label{fig:resSyntIInn}}
\end{figure*}
\begin{figure*}[!htp]
  \centering
  \subfloat[5\% of training data.\label{fig:BCCTaROC5}]{\includegraphics[width=0.3\linewidth]{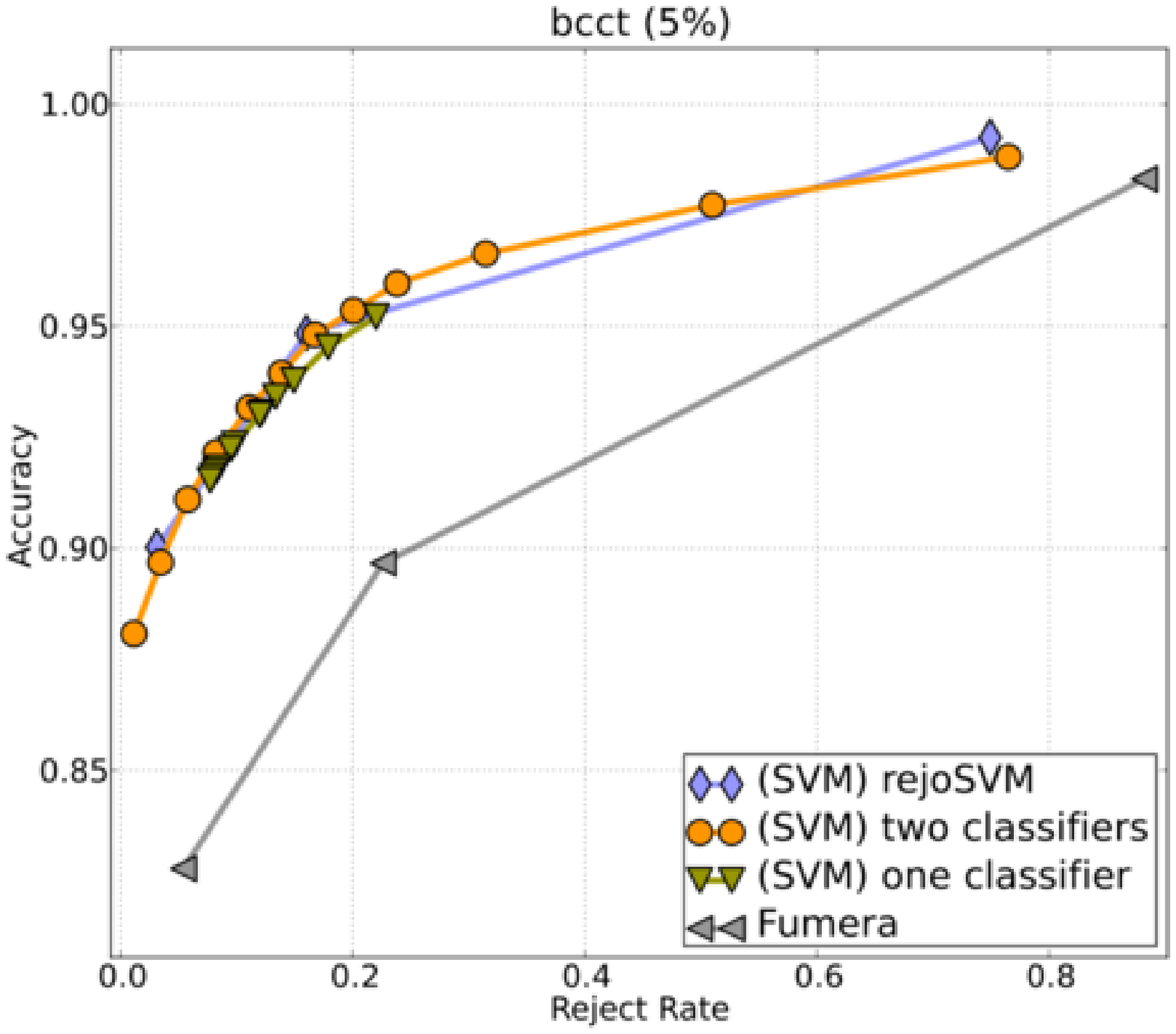}}
  \quad
  \subfloat[25\% of training data.\label{fig:BCCTcROC25}]{\includegraphics[width=0.3\linewidth]{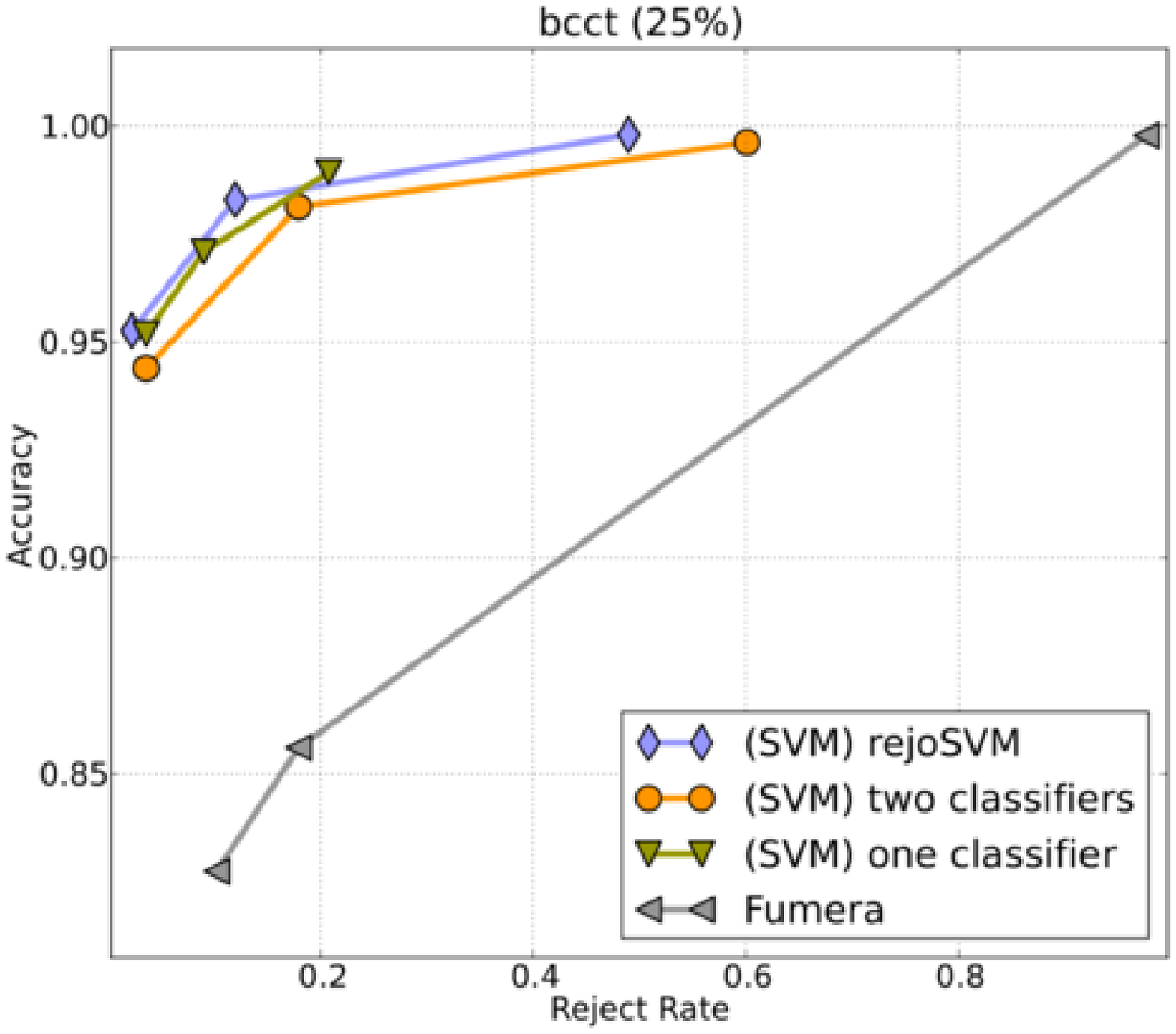}}
  \quad
  \subfloat[40\% of training data.\label{fig:BCCTcROC40}]{\includegraphics[width=0.3\linewidth]{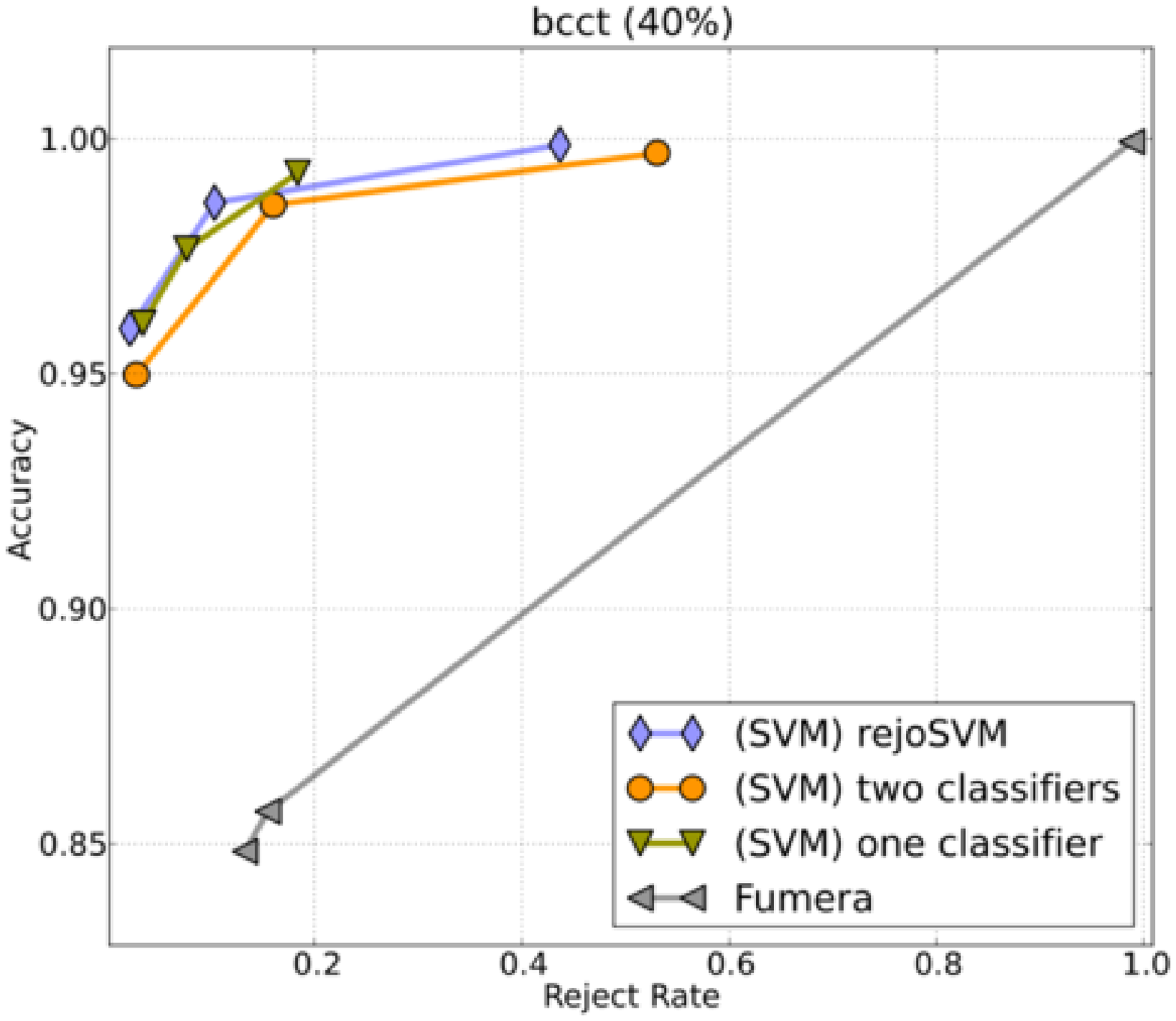}}
  \caption{The A-R curves for the binary BCCT dataset (SVM methods only).\label{fig:resBCCT}}
\end{figure*}
\begin{figure*}[!htp]
  \centering
  \subfloat[5\% of training data.\label{fig:BCCTaROC5}]{\includegraphics[width=0.3\linewidth]{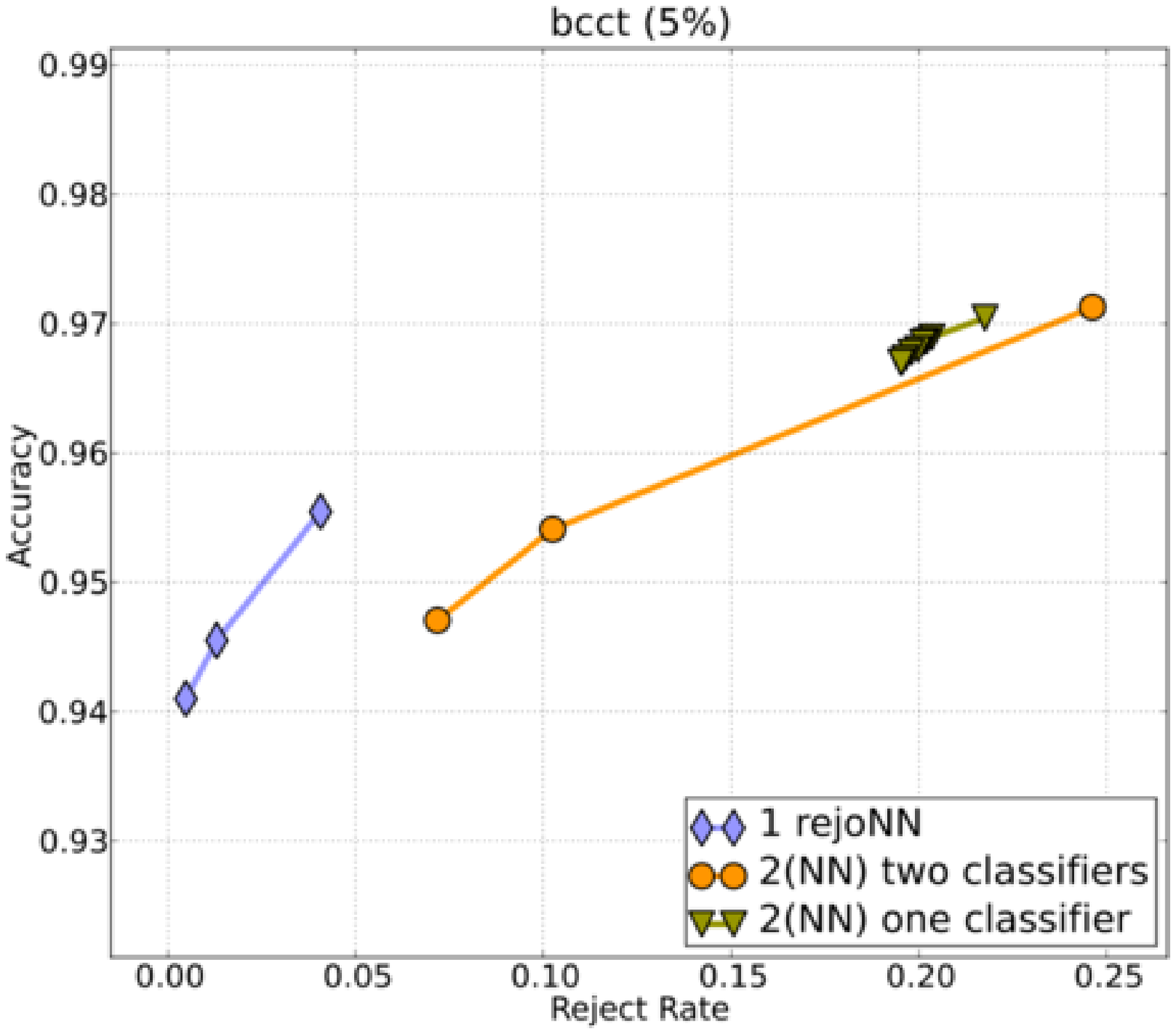}}
  \quad
  \subfloat[25\% of training data.\label{fig:BCCTeROC25}]{\includegraphics[width=0.3\linewidth]{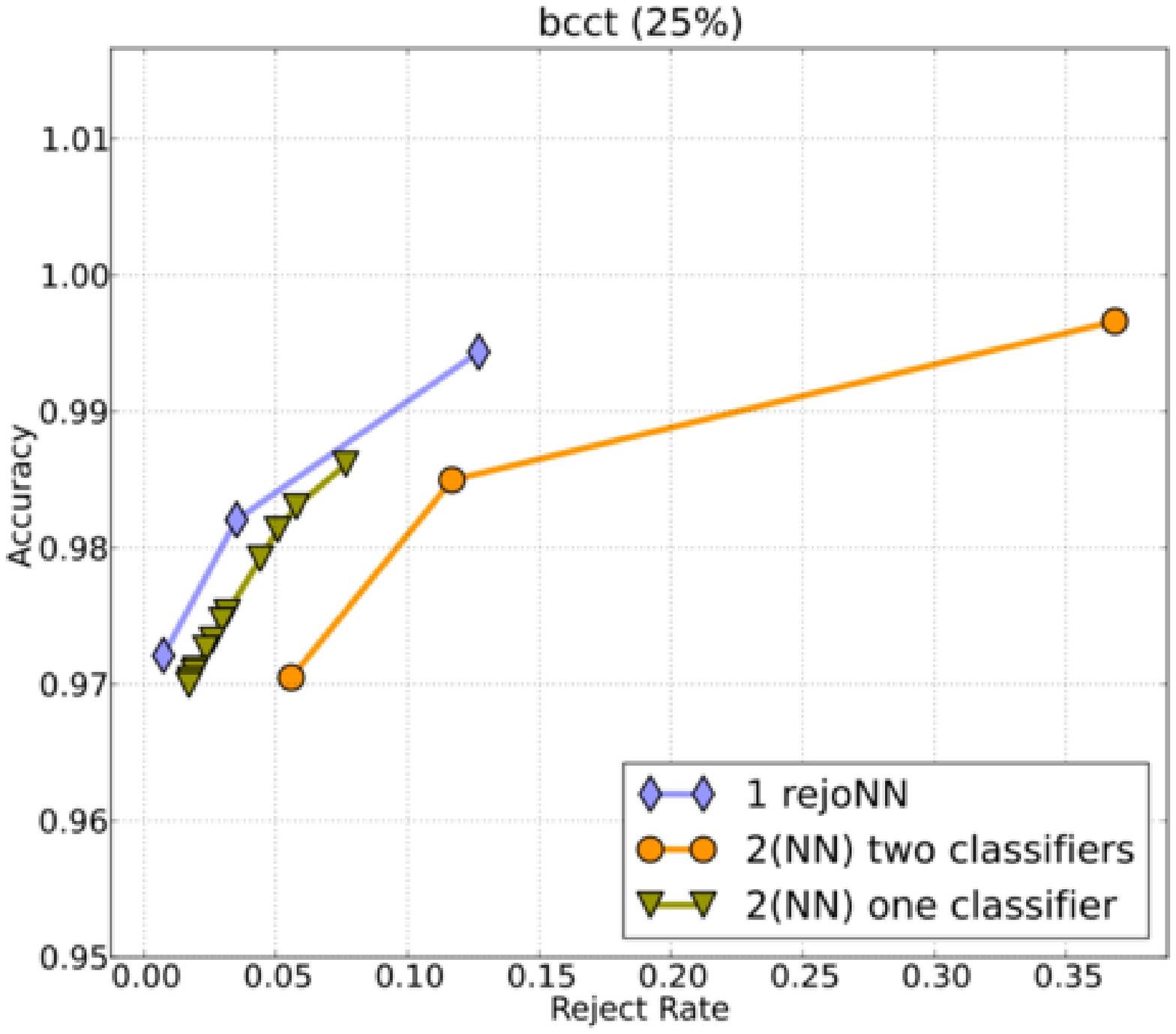}}
  \quad
  \subfloat[40\% of training data.\label{fig:BCCTcROC40}]{\includegraphics[width=0.3\linewidth]{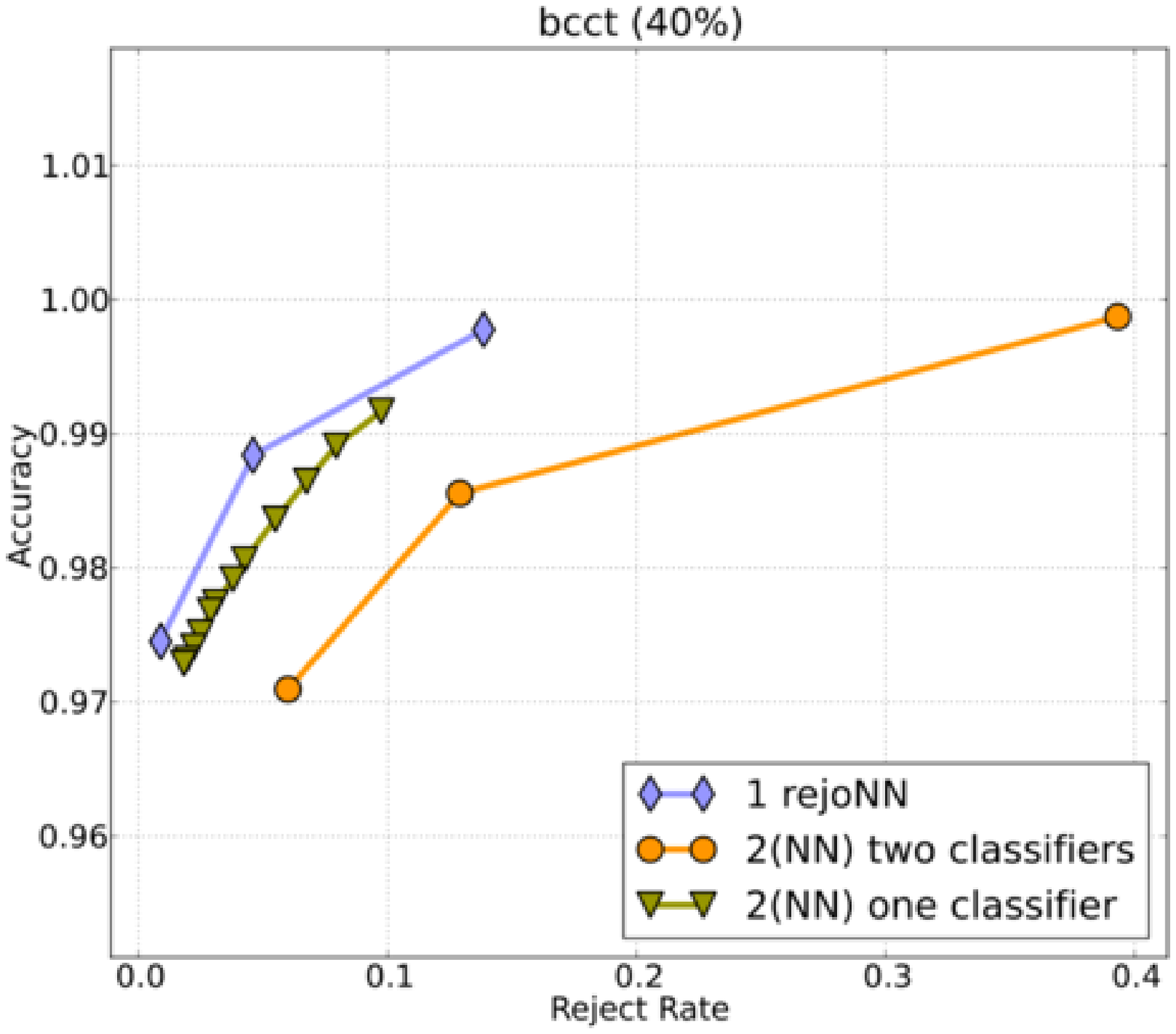}}
  \caption{The A-R curves for the binary BCCT dataset (NN methods only).\label{fig:resBCCT}}
\end{figure*}
\begin{figure*}[!htp]
  \centering
  \subfloat[5\% of training data.\label{fig:BCCTaROC5}]{\includegraphics[width=0.3\linewidth]{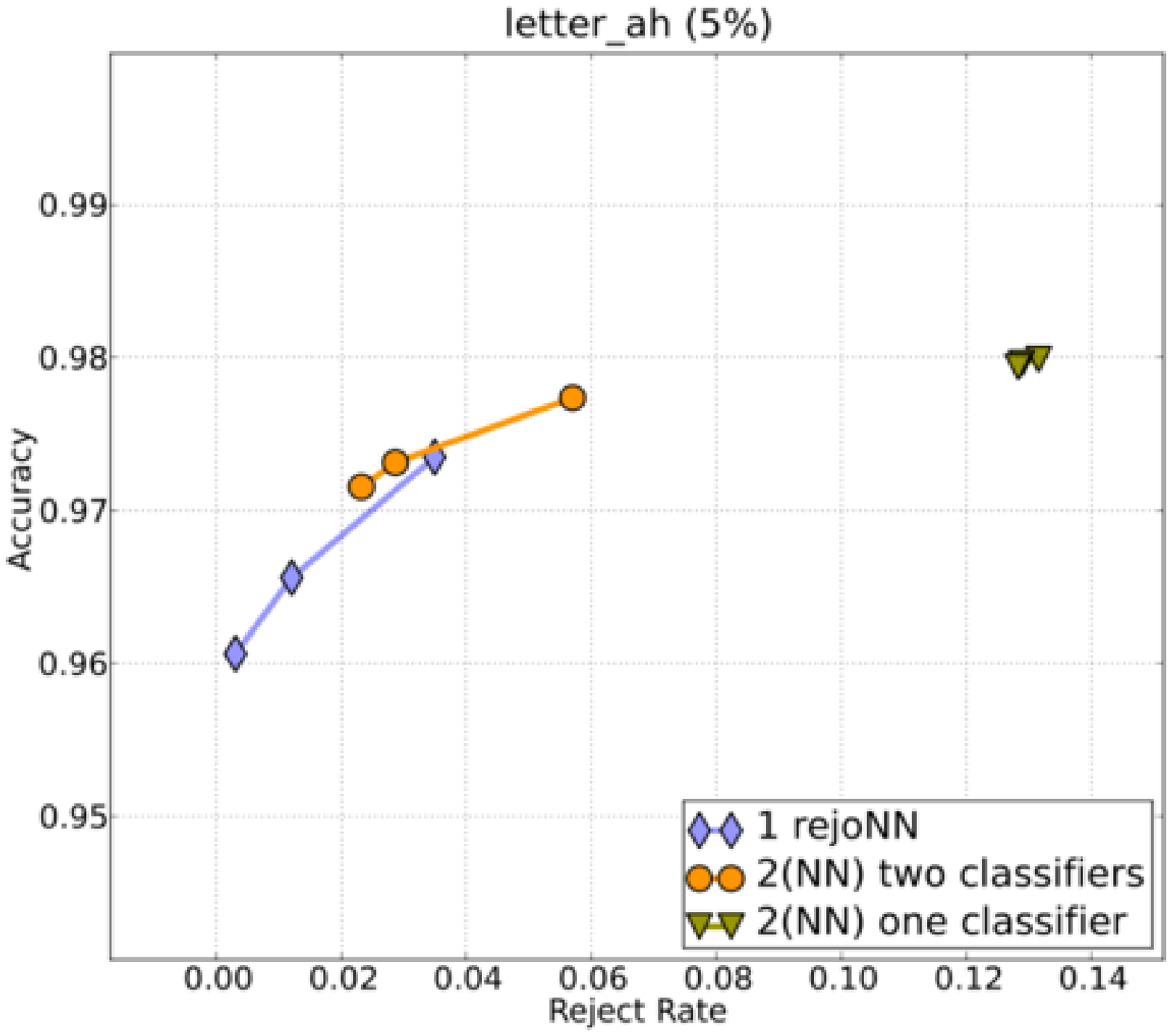}}
  \quad
  \subfloat[25\% of training data.\label{fig:BCCTeROC25}]{\includegraphics[width=0.3\linewidth]{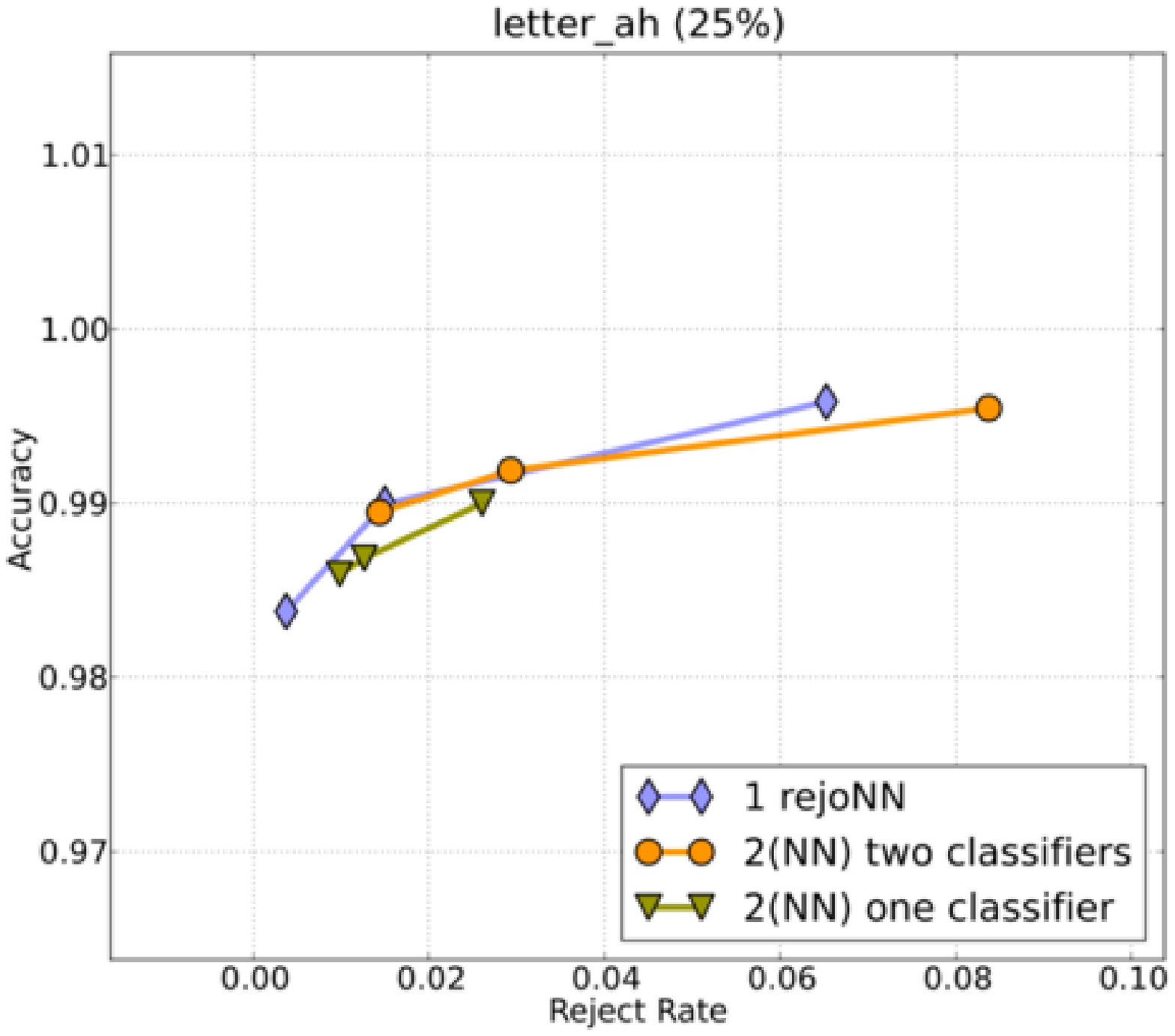}}
  \quad
  \subfloat[40\% of training data.\label{fig:BCCTcROC40}]{\includegraphics[width=0.3\linewidth]{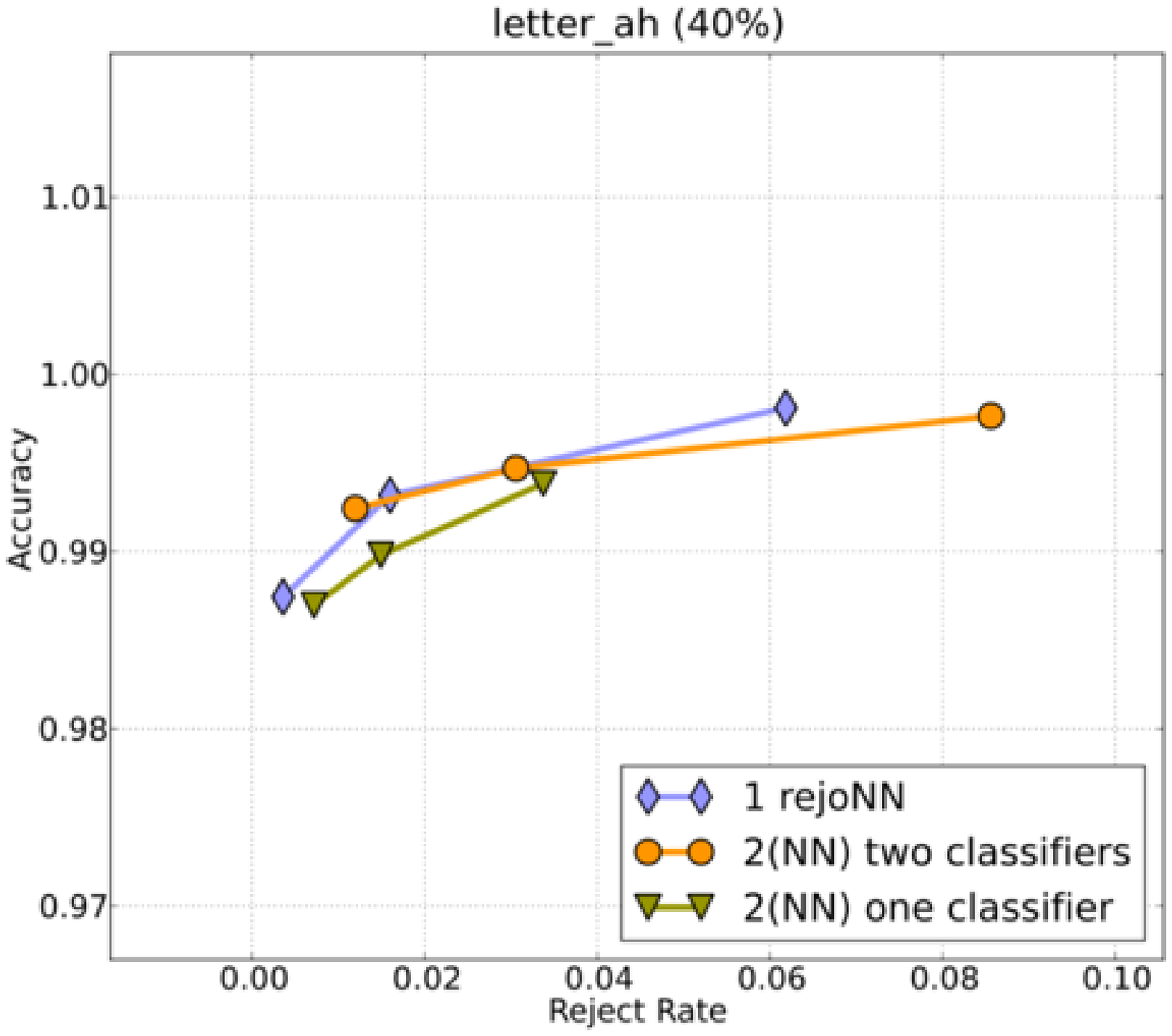}}
  \caption{The A-R curves for the binary Letter AH dataset (NN methods only).\label{fig:resLetter}}
\end{figure*}
\begin{figure*}[!htp]
  \centering
  \subfloat[5\% of training data.\label{fig:SyntheticIIIabROC5}]{\includegraphics[width=0.3\linewidth]{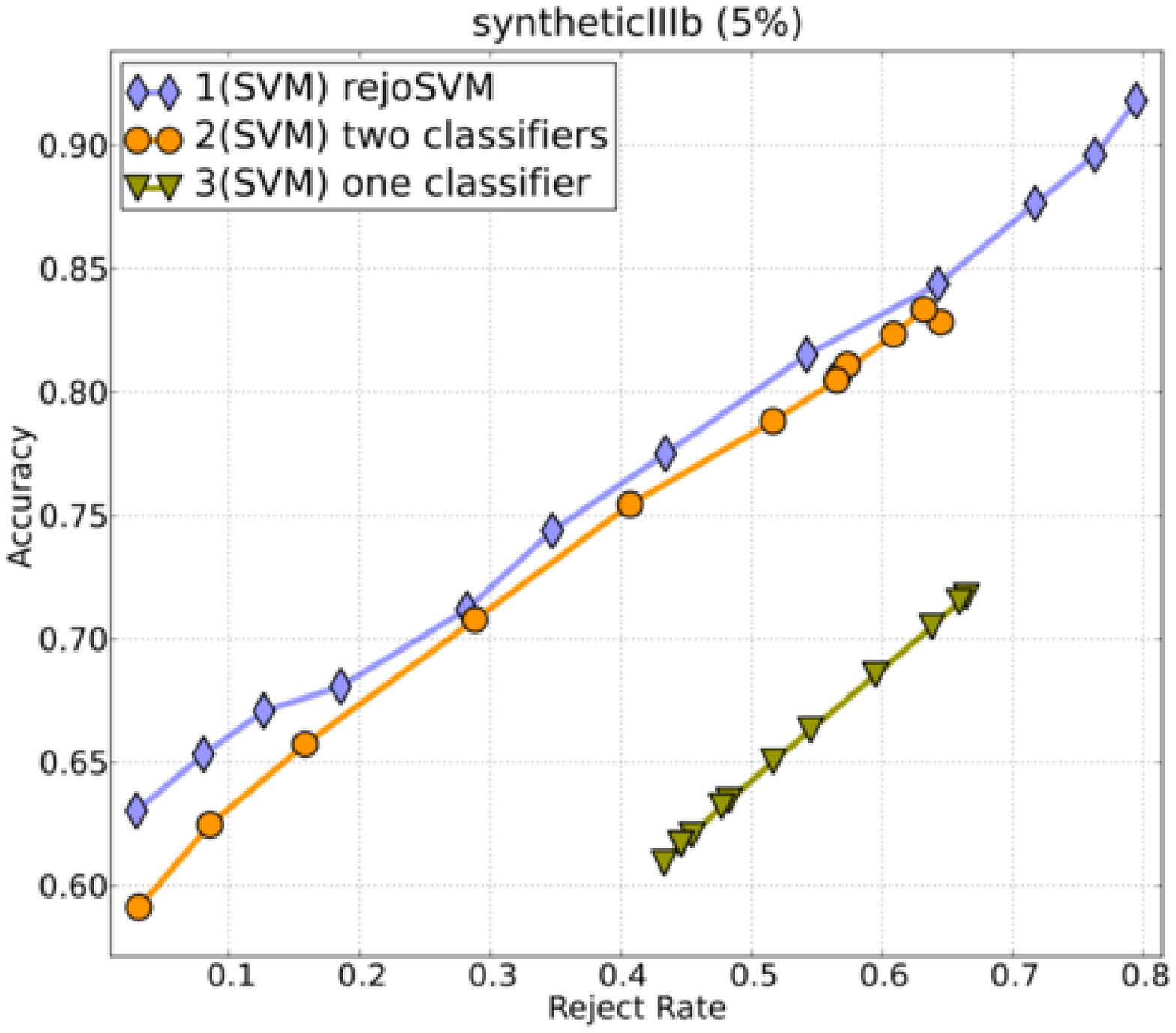}}
  \quad
  \subfloat[25\% of training data.\label{fig:SyntheticIIIcbROC25}]{\includegraphics[width=0.3\linewidth]{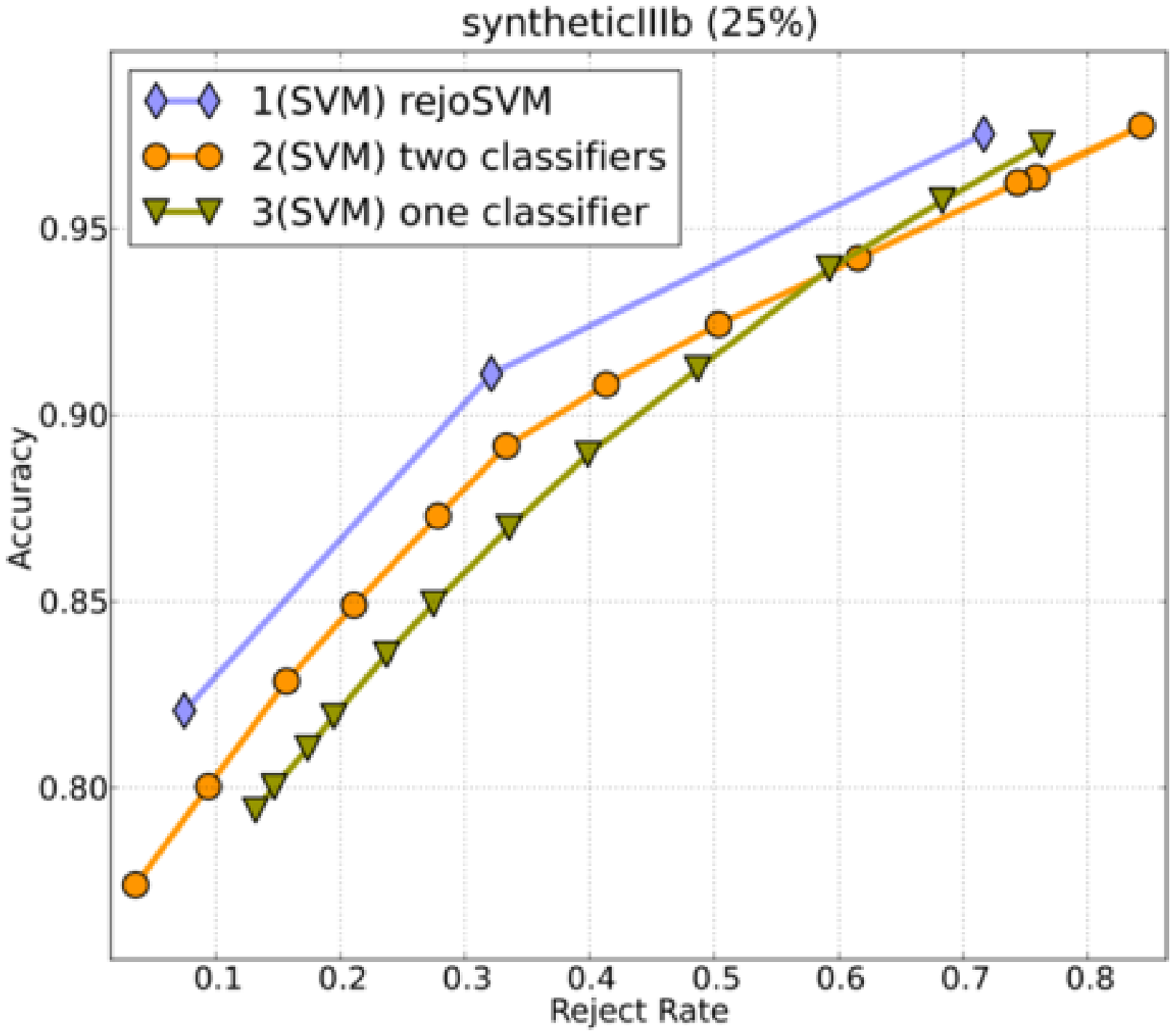}}
  \quad
  \subfloat[40\% of training data.\label{fig:SyntheticIIIcbROC40}]{\includegraphics[width=0.3\linewidth]{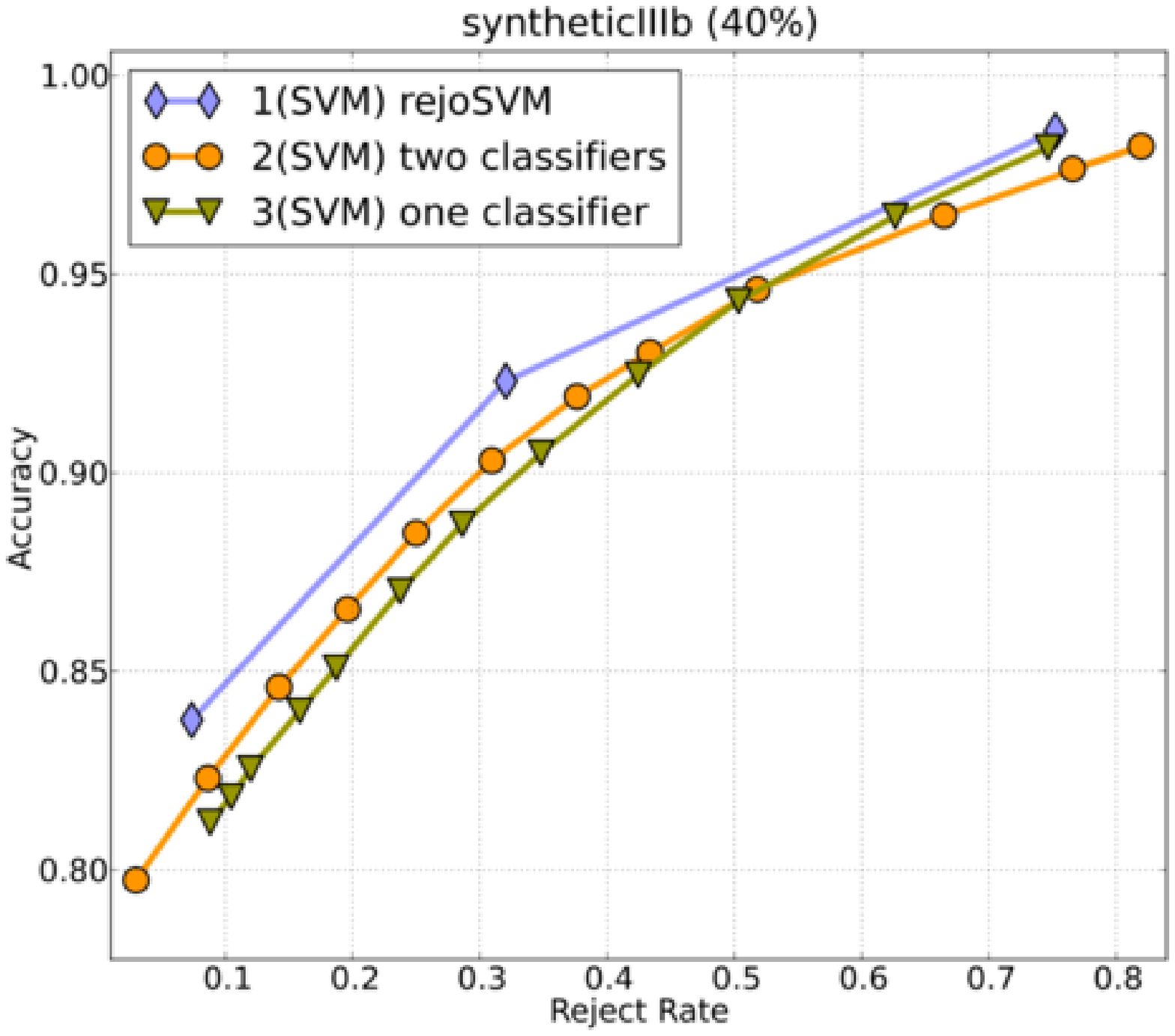}}
  \caption{The A-R curves for the syntheticIII dataset (SVM methods only).\label{fig:resSyntIIIb}}
\end{figure*}
\begin{figure*}[!htp]
  \centering
  \subfloat[5\% of training data.\label{fig:SyntheticIVaROC}]{\includegraphics[width=0.3\linewidth]{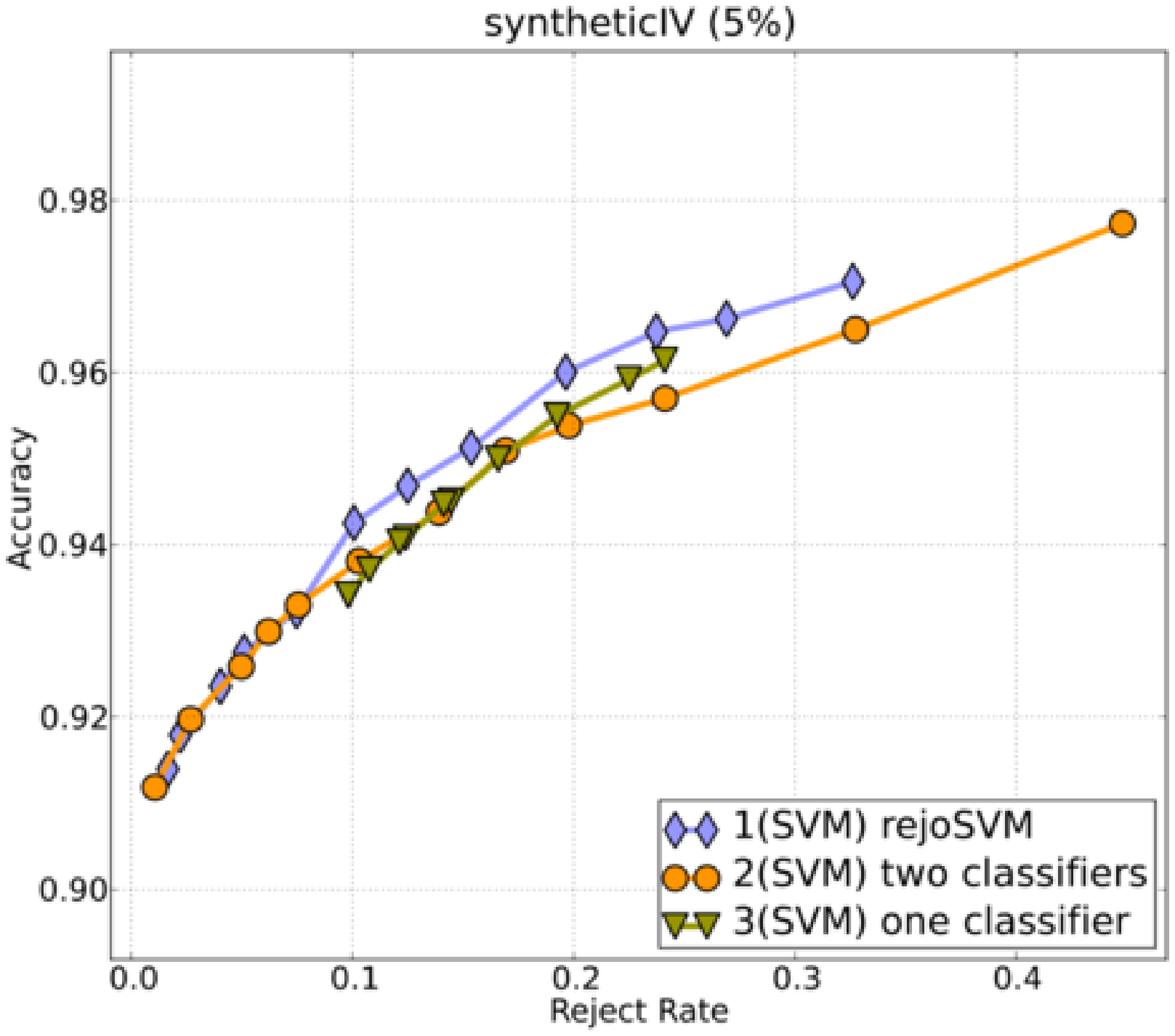}}
  \quad
  \subfloat[25\% of training data.\label{fig:SyntheticIVcROC}]{\includegraphics[width=0.3\linewidth]{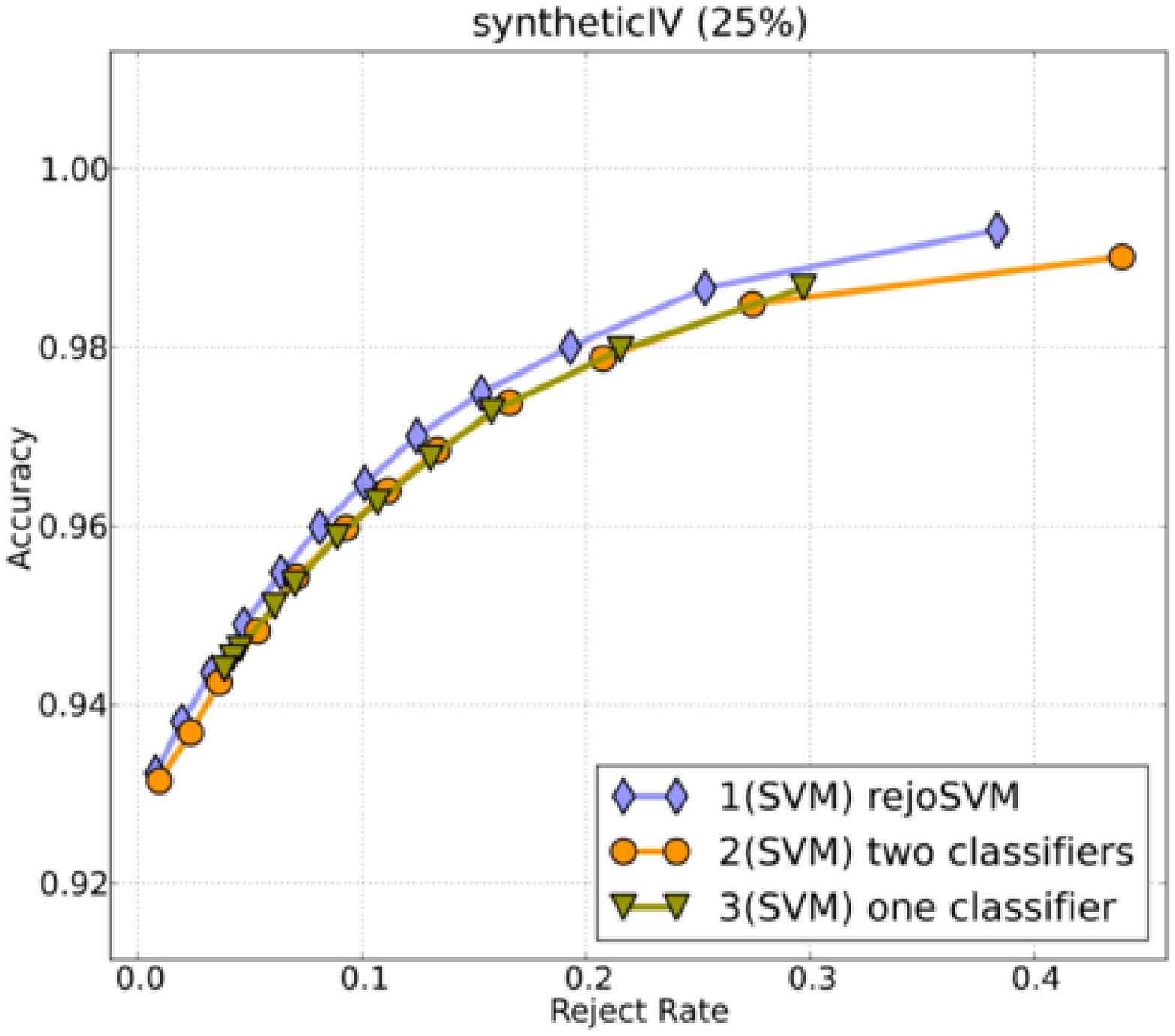}}
  \quad
  \subfloat[40\% of training data.\label{fig:SyntheticIVcROC}]{\includegraphics[width=0.3\linewidth]{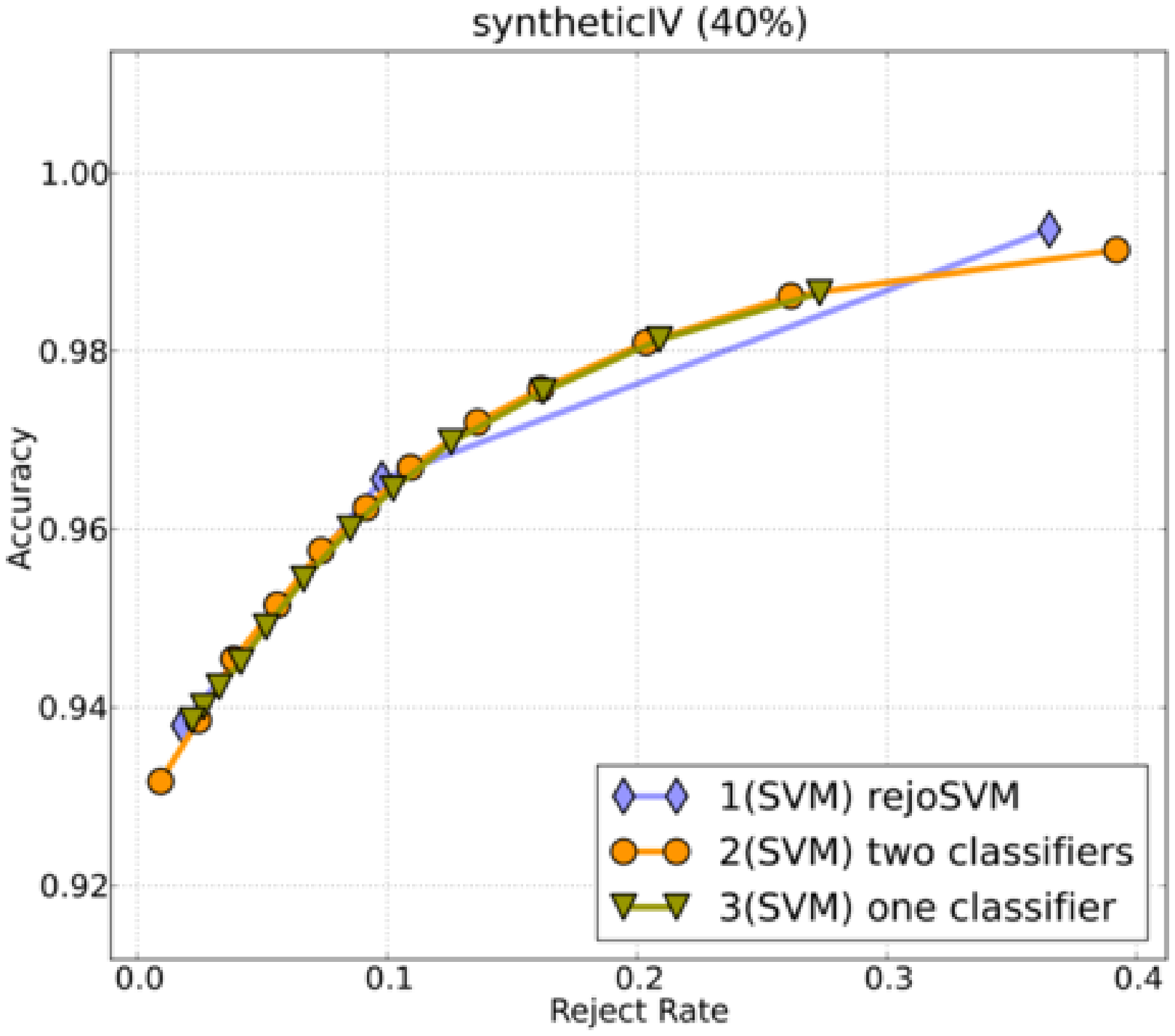}}
  \caption{The A-R curves for the syntheticIV dataset (SVM methods only).\label{fig:resSyntIV}}
\end{figure*}
\begin{figure*}[!htp]
  \centering
  \subfloat[5\% of training data.\label{fig:BCCTaROC5}]{\includegraphics[width=0.3\linewidth]{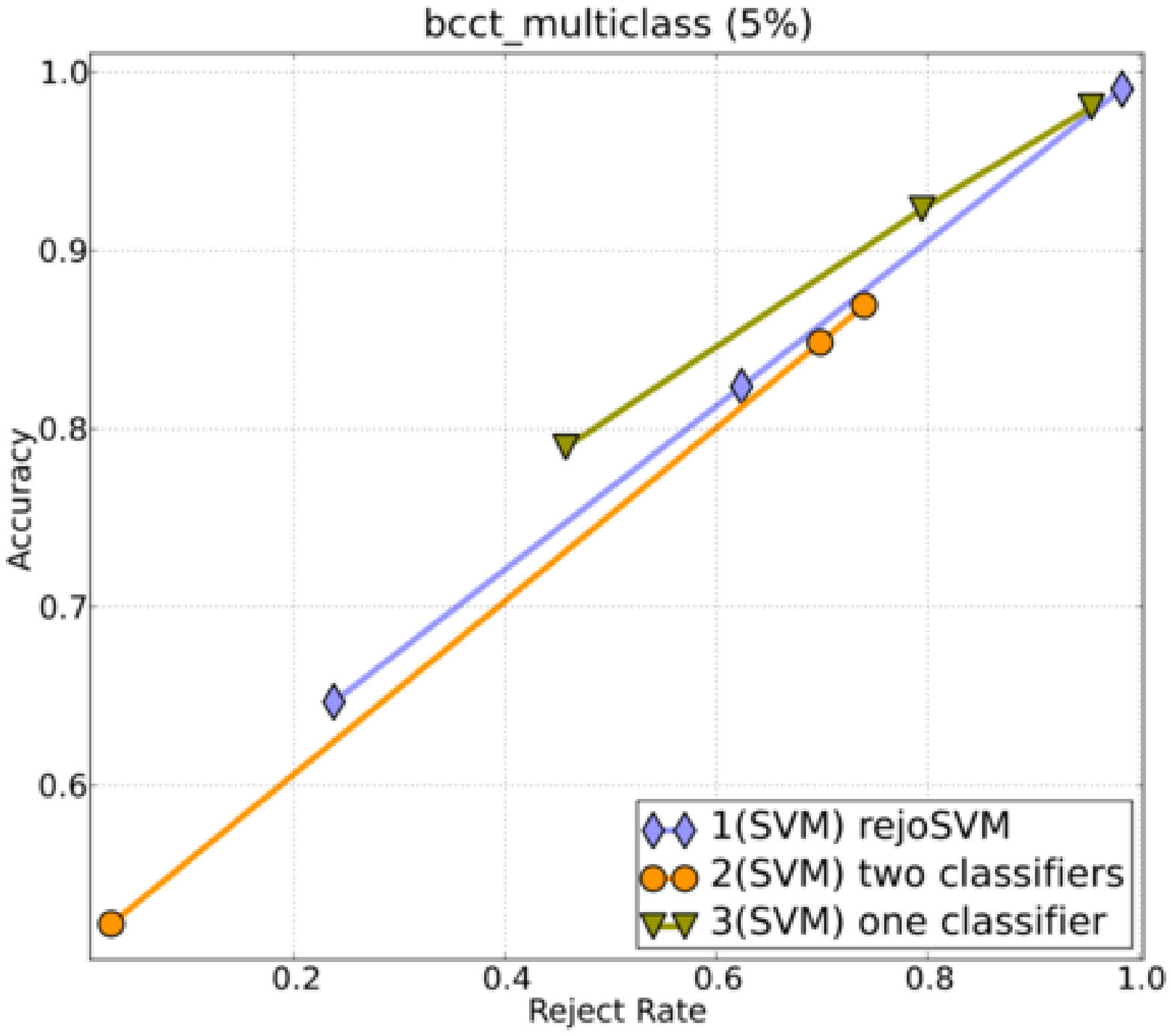}}
  \quad
  \subfloat[25\% of training data.\label{fig:BCCTeROC25}]{\includegraphics[width=0.3\linewidth]{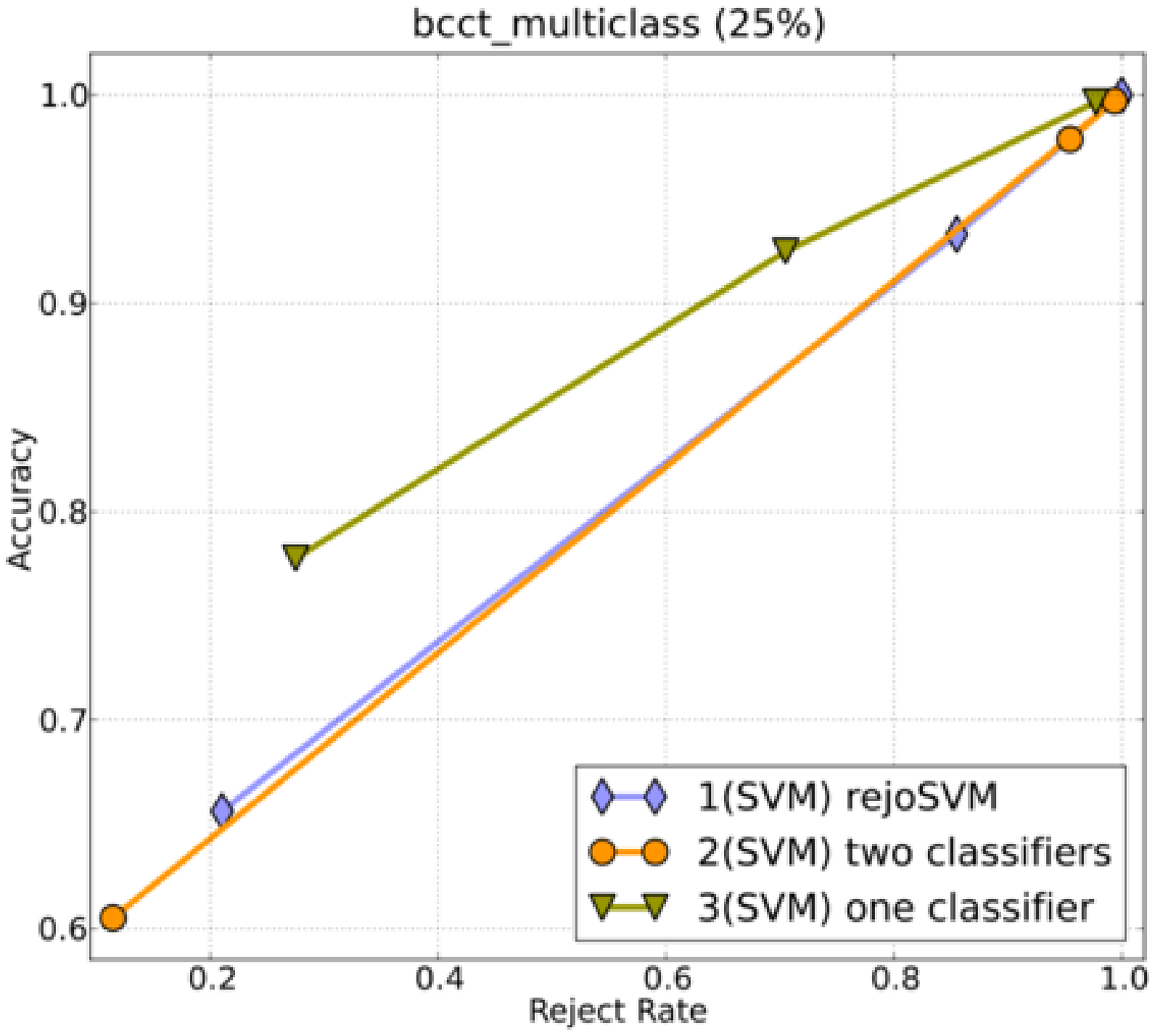}}
  \quad
  \subfloat[40\% of training data.\label{fig:BCCTcROC40}]{\includegraphics[width=0.3\linewidth]{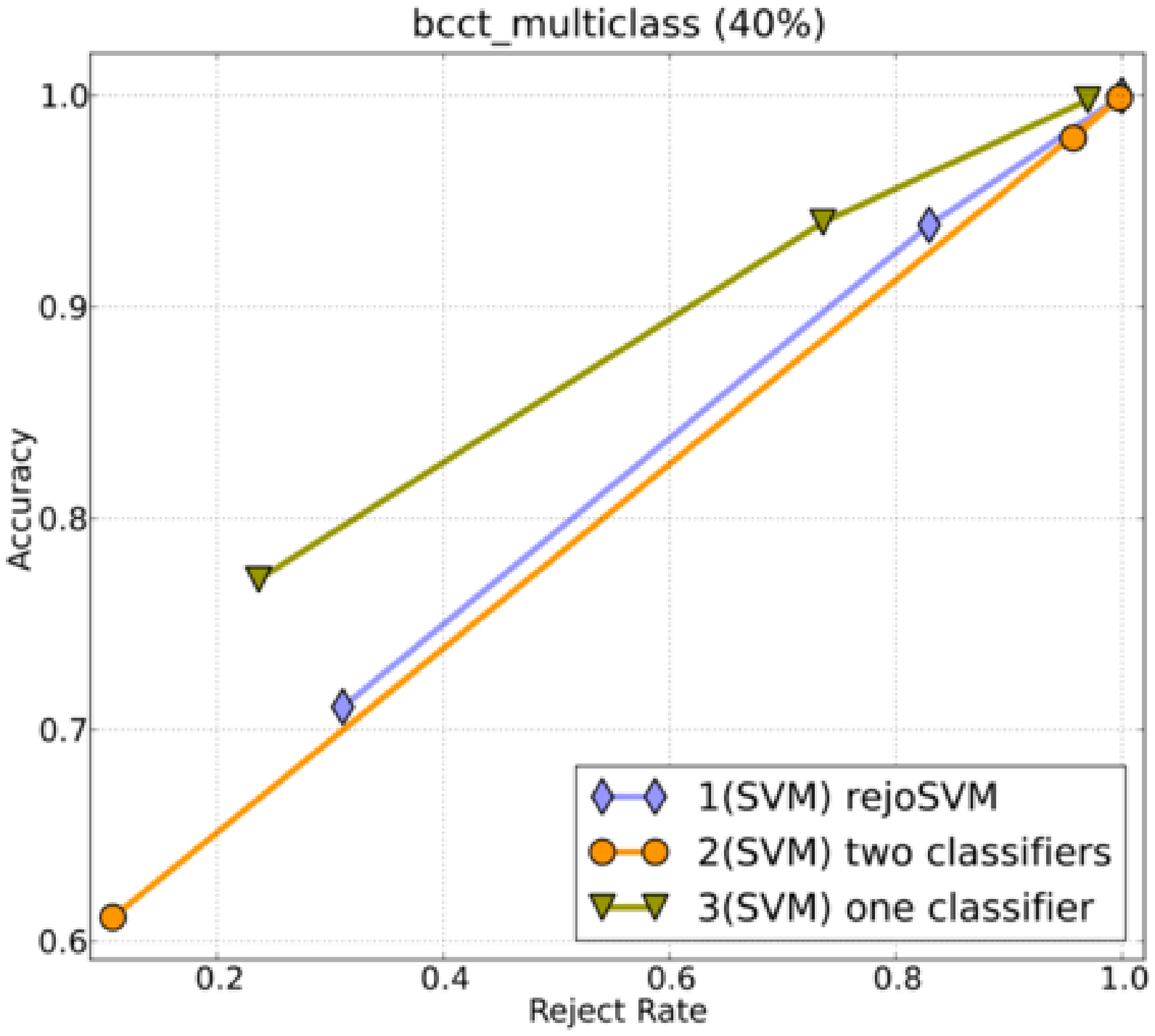}}
  \caption{The A-R curves for the BCCT dataset (SVM methods only).\label{fig:resBCCT}}
\end{figure*}

A first main assertion is that in overall rejoSVM performed better than any of the other methods under comparison, over the full range of values for $w_r$. 
The less positive results obtained by Fumera's method may be due to the incorrect use of the implementation at our disposal. Nevertheless, since only linear kernels were implemented, we extended the datasets with second order terms $x_i x_j$ when evaluating this method. In this extended space, the optimal solutions for the synthetic datasets are indeed linear.

It is also observable that, in general, SVM based methods outperform the neural network counterparts, in line with the current view in the research community.
When restricting the attention to neural network methods, the proposed rejoNN exhibits often the best performance. Moreover, it is important to emphasise that rejoSVM and rejoNN approaches have the advantage of simplicity, using a single direction for all boundaries, and interpretability. The insight of looking to the reject option problem as an ordinal class setting can promote new lines of research.

Finally, we highlight that the proposed framework:
1) has the capability to detect reject regions with a single standard binary classifier;
2) does not required the addition of any confidence level, or thresholds, to define the trust regions; and
3) does not generate ambiguity regions as the two classifier approach, as it was presented in Fig.~\ref{fig:IntersectingSetting}.

\section{Conclusion}
\label{sec:five}
In this paper, we proposed an extension of the data replication method~\cite{JaimeJMLR2007} that directly embeds reject option. 
This extension was derived by taken a new perspective of the classification with reject option problem, viewing the three output classes as naturally ordered. 
A pair of non-intersecting boundaries delimits the rejection region provided by our model. Our proposal has the advantages of using a standard binary classifier and embedding the design of the reject region during the training process. Moreover, the method allows a flexible definition of the position and orientation of the boundaries, which can change for different values of the cost of rejections $w_r$. 
This method was mapped into neural networks and support vector machines with very positive results.
This work can be a useful contribution in the area and the availability of the code under the reproducible research guidelines can encourage others to make use of and to build on it.

\bibliography{rejectOptionArticle}
\bibliographystyle{plain}

\end{document}